\documentclass[a4paper,12pt]{ThesisStyle}
\usepackage[T1]{fontenc}
\usepackage{bm}
\usepackage{bbm}
\usepackage[utf8]{inputenc}
\usepackage{latexsym}
\usepackage[english]{babel}
\usepackage{indentfirst}
\usepackage{graphicx}
\usepackage{lmodern}
\usepackage[sc]{mathpazo}
\usepackage{geometry}
\input Kramer.fd
\usepackage{yfonts,color}
\usepackage{minitoc}
\usepackage{titlesec}
\usepackage{subfigure}
\usepackage{amsmath}
\usepackage{amssymb}
\usepackage{stmaryrd}
\usepackage{url}
\usepackage{pgfplots}
\pgfplotsset{compat=1.5}
\usepackage[colorlinks,linkcolor=red!80!black,
citecolor=red!80!black,pdfpagelabels,hyperindex=true]{hyperref}
\hypersetup{
    colorlinks=true,       % false: boxed links; true: colored links
    linkcolor=brown!80!black,  % color of internal links (change box color with linkbordercolor)
    citecolor=green!50!black,        % color of links to bibliography
    filecolor=magenta,      % color of file links
    urlcolor=red!80!black           % color of external links
}
\usepackage{csquotes}
\usepackage[sorting=none,backend=bibtex,style=numeric-comp]{biblatex}
\usepackage{cancel}
\usepackage{pifont}
\usepackage{tikz}
\usetikzlibrary{matrix,arrows,decorations,backgrounds,shapes,calc,fit}
\usetikzlibrary{fadings}
\usetikzlibrary{decorations.pathmorphing}
\usetikzlibrary{decorations.markings}
\usepackage{xcolor}
\usepackage{amsfonts}
\usepackage{slashed}
\usepackage{fancybox}
\usepackage{fancyhdr}
{\cleardoublepage \null \vfill \begin{center}%
\bfseries \abstractname\end{center}}%
{\vfill \null}

\newcommand{\slv}{\raise.15ex\hbox{$/$}\kern-.53em\hbox{$v$}}
\newcommand{\slF}{\raise.15ex\hbox{$/$}\kern-.53em\hbox{$F$}}
\newcommand{\slL}{\raise.15ex\hbox{$/$}\kern-.53em\hbox{$L$}}
\newcommand{\slP}{\raise.15ex\hbox{$/$}\kern-.53em\hbox{$P$}}
\newcommand{\slp}{\raise.15ex\hbox{$/$}\kern-.53em\hbox{$p$}}
\newcommand{\slq}{\raise.15ex\hbox{$/$}\kern-.53em\hbox{$q$}}
\newcommand{\slR}{\raise.15ex\hbox{$/$}\kern-.53em\hbox{$R$}}
\newcommand{\slQ}{\raise.15ex\hbox{$/$}\kern-.53em\hbox{$Q$}}
\newcommand{\slK}{\raise.15ex\hbox{$/$}\kern-.53em\hbox{$K$}}
\newcommand{\slk}{\raise.15ex\hbox{$/$}\kern-.53em\hbox{$k$}}
\newcommand{\slD}{\raise.15ex\hbox{$/$}\kern-.73em\hbox{$D$}}
\newcommand{\slC}{\raise.15ex\hbox{$/$}\kern-.53em\hbox{$C$}}
\newcommand{\slA}{\raise.15ex\hbox{$/$}\kern-.53em\hbox{$A$}}
\newcommand{\slSigma}{\raise.15ex\hbox{$/$}\kern-.53em\hbox{$\Sigma$}}
\newcommand{\slpartial}{\raise.15ex\hbox{$/$}\kern-.53em\hbox{$\partial$}}
\newcommand{\slcalP}{\raise.15ex\hbox{$/$}\kern-.63em\hbox{$\cal P$}}
\definecolor{purp}{RGB}{0,0,0}
\def\p{{\boldsymbol p}}

\def\m{{\boldsymbol m}}
\def\x{{\boldsymbol x}}

\def\y{{\boldsymbol y}}

\tikzstyle{blueballa} = [circle,shading=ball, ball color=blue!30,inner sep =1.2mm]
\tikzstyle{redballa} = [circle,shading=ball, ball color=red,inner sep =1.2mm]
\tikzstyle{greenballa} = [circle,shading=ball, ball color=green!70!black,inner sep =1.2mm]
\tikzstyle{blueball} = [circle,shading=ball, ball color=blue!30,inner sep =0.3mm]
\tikzstyle{redball} = [circle,shading=ball, ball color=red,inner sep =0.3mm]
\tikzstyle{greenball} = [circle,shading=ball, ball color=green!70!black,inner sep =0.3mm]
\tikzstyle{gluon} = [thick, style={decorate,decoration={coil,amplitude=4pt, segment length=4pt}}]

\tikzstyle{photon} = [very thin, style={decorate, decoration={snake,amplitude=0.4pt, segment length=2pt}}]
\tikzfading [name=radialfade, inner color=transparent!40, outer color=transparent!100]

\newlength{\longueurAdHoc}
\usepackage{setspace}
\usepackage{pagedecouv}

\usepackage[boxruled,vlined]{algorithm2e}

\usepackage{float}
\floatstyle{plain}
\newfloat{myalgo}{tbhp}{mya}
{\begin{myalgo}[#1]
\centering
\begin{minipage}{#2}
\begin{algorithm}[H]}%
{\end{algorithm}
\end{minipage}
\end{myalgo}}

\makeatletter
\newcounter {subsubsubsection}[subsubsection]
\renewcommand\thesubsubsubsection{\thesubsubsection .\@alph\c@subsubsubsection}
\newcommand\subsubsubsection{\@startsection{subsubsubsection}{4}{\z@}%
                                     {-3.25ex\@plus -1ex \@minus -.2ex}%
                                     {1.5ex \@plus .2ex}%
                                     {\normalfont\normalsize\bfseries}}
\renewcommand\paragraph{\@startsection{paragraph}{5}{\z@}%
                                    {3.25ex \@plus1ex \@minus.2ex}%
                                    {-1em}%
                                    {\normalfont\normalsize\bfseries}}
\renewcommand\subparagraph{\@startsection{subparagraph}{6}{\parindent}%
                                       {3.25ex \@plus1ex \@minus .2ex}%
                                       {-1em}%
                                      {\normalfont\normalsize\bfseries}}
\newcommand*\l@subsubsubsection{\@dottedtocline{4}{10.0em}{4.1em}}
\renewcommand*\l@paragraph{\@dottedtocline{5}{10em}{5em}}
\renewcommand*\l@subparagraph{\@dottedtocline{6}{12em}{6em}}
\newcommand*{\subsubsubsectionmark}[1]{}
\makeatother

\makeatletter
\def\toclevel@subsubsubsection{4}
\def\toclevel@paragraph{5}
\def\toclevel@subparagraph{6}
\makeatother
\tikzset{
    hyperlink node/.style={
        alias=sourcenode,
        append after command={
            let     \p1 = (sourcenode.north west),
                \p2=(sourcenode.south east),
                \n1={\x2-\x1},
                \n2={\y1-\y2} in
            node [inner sep=0pt, outer sep=0pt,anchor=north west,at=(\p1)] {\hyperlink{#1}{\phantom{\rule{\n1}{\n2}}}}
        }
    }
}
\usepackage{appendix}
\usepackage{etoolbox}
\usepackage{lipsum}
\AtBeginEnvironment{subappendices}{%
\section*{Appendix}
\addcontentsline{toc}{section}{Appendices}
}

\renewbibmacro*{author}{%
\mkbibbold{%
  \ifboolexpr{
    test \ifuseauthor
    and
    not test {\ifnameundef{author}}
  }
    {\printnames{author}%
     \iffieldundef{authortype}
       {}
       {\setunit{\addcomma\space}%
        \usebibmacro{authorstrg}}}
    {}}}
\begin{document}
\def\layersep{2.5cm}
\pgfmathsetseed{12}
\newcommand{\midarrow}{\tikz \draw[-stealth] (0,0) -- +(.1,0);}
\newcommand{\midarroww}{\tikz \draw[-stealth] (0,0.1) -- +(.1,0);}
\newcommand{\orient}{\tikz \draw[-stealth] (0,0) -- +(0,.001);}
\newcommand{\orientl}{\tikz \draw[-stealth] (0,0.15) -- +(.005,.01);}
\newcommand{\orientr}{\tikz \draw[-stealth] (0,0.15) -- +(-.005,.01);}
 
\thispagestyle{empty}
\newgeometry{hmargin=0cm,vmargin=1cm}
\begin{center}
\begin{tikzpicture}
\node[] at (0,0) {\includegraphics[scale=1]{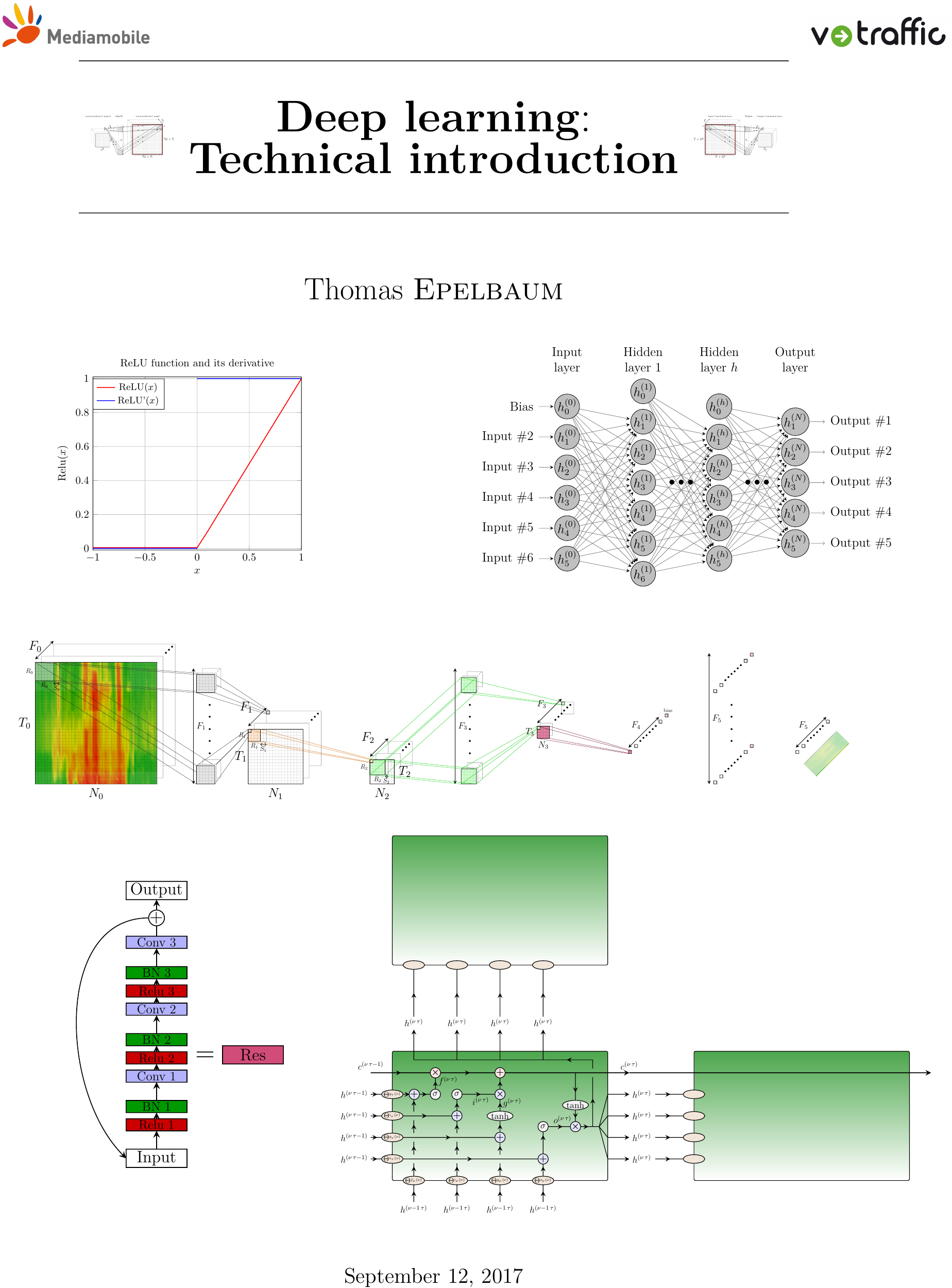}};
\end{tikzpicture}
\end{center}

\restoregeometry
\dominitoc \tableofcontents

\chapter{Preface}

\yinipar{\fontsize{60pt}{72pt}\usefont{U}{Kramer}{xl}{n}I} started learning about deep learning fundamentals in February 2017. At this time, I knew nothing about backpropagation, and was completely ignorant about the differences between a Feedforward, Convolutional and a Recurrent Neural Network.

\vspace{0.2cm}

As I navigated through the humongous amount of data available on deep learning online, I found myself quite frustrated when it came to really understand what deep learning is, and not just applying it with some available library.

\vspace{0.2cm}

In particular, the backpropagation update rules are seldom derived, and never in index form. Unfortunately for me, I have an "index" mind: seeing a 4 Dimensional convolution formula in matrix form does not do it for me. Since I am also stupid enough to like recoding the wheel in low level programming languages, the matrix form cannot be directly converted into working code either.

\vspace{0.2cm}

I therefore started some notes for my personal use, where I tried to rederive everything from scratch in index form.

\vspace{0.2cm}

I did so for the vanilla Feedforward network, then learned about L1 and L2 regularization , dropout\cite{Srivastava:2014:DSW:2627435.2670313}, batch normalization\cite{Ioffe2015}, several gradient descent optimization techniques... Then turned to convolutional networks, from conventional single digit number of layer conv-pool architectures\cite{Lecun98gradient-basedlearning} to recent VGG\cite{DBLP:journals/corr/SimonyanZ14a} ResNet\cite{He2015} ones, from local contrast normalization and rectification to bacthnorm... And finally I studied Recurrent Neural Network structures\cite{GravesA2016}, from the standard formulation to the most recent LSTM one\cite{Gers:2000:LFC:1121912.1121915}.

\vspace{0.2cm}

As my work progressed, my notes got bigger and bigger, until a point when I realized I might have enough material to help others starting their own deep learning journey.

\vspace{0.2cm}

This work is bottom-up at its core. If you are searching a working Neural Network in 10 lines of code and 5 minutes of your time, you have come to the wrong place. If you can mentally multiply and convolve 4D tensors, then I have nothing to convey to you either.

\vspace{0.2cm}

If on the other hand you like(d) to rederive every tiny calculation of every theorem of every class that you stepped into, then you might be interested by what follow!

\chapter{Acknowledgements}

\yinipar{\fontsize{60pt}{72pt}\usefont{U}{Kramer}{xl}{n}T}his work has no benefit nor added value to the deep learning topic on its own. It is just the reformulation of ideas of brighter researchers to fit a peculiar mindset: the one of preferring formulas with ten indices but where one knows precisely what one is manipulating rather than (in my opinion sometimes opaque) matrix formulations where the dimension of the objects are rarely if ever specified.

\vspace{0.2cm}

Among the brighter people from whom I learned online are Andrew Ng. His Coursera class (\href{https://www.coursera.org/learn/machine-learning}{here}) was the first contact I got with Neural Network, and this pedagogical introduction allowed me to build on solid ground.

\vspace{0.2cm}

I also wish to particularly thanks Hugo Larochelle, who not only built a wonderful deep learning class (\href{http://info.usherbrooke.ca/hlarochelle/neural_networks/content.html}{here}), but was also kind enough to answer emails from a complete beginner and stranger!

\vspace{0.2cm}

The Stanford class on convolutional networks (\href{http://cs231n.github.io/convolutional-networks/}{here}) proved extremely valuable to me, so did the one on Natural Language processing (\href{http://web.stanford.edu/class/cs224n/}{here}).

\vspace{0.2cm}

I also benefited greatly from Sebastian Ruder's blog (\href{http://ruder.io/#open}{here}), both from the blog pages on gradient descent optimization techniques and from the author himself.

\vspace{0.2cm}

I learned more about LSTM on colah's blog (\href{http://colah.github.io/posts/2015-08-Understanding-LSTMs/}{here}), and some of my drawings are inspired from there.

\vspace{0.2cm}

I also thank Jonathan Del Hoyo for the great articles that he regularly shares on LinkedIn.

\vspace{0.2cm}

Many thanks go to my collaborators at Mediamobile, who let me dig as deep as I wanted on Neural Networks. I am especially indebted to Clément, Nicolas, Jessica, Christine and Céline.

\vspace{0.2cm}

Thanks to Jean-Michel Loubes and Fabrice Gamboa, from whom I learned a great deal on probability theory and statistics.

\vspace{0.2cm}

I end this list with my employer, Mediamobile, which has been kind enough to let me work on this topic with complete freedom. A special thanks to Philippe, who supervised me with the perfect balance of feedback and freedom!

\chapter{Introduction}

\yinipar{\fontsize{60pt}{72pt}\usefont{U}{Kramer}{xl}{n}T}his note aims at presenting the three most common forms of neural network architectures. It does so in a technical though hopefully pedagogical way, buiding up in complexity as one progresses through the chapters.

\vspace{0.2cm}

Chapter \ref{sec:chapterFNN} starts with the first type of network introduced historically: a regular feedforward neural network, itself an evolution of the original perceptron \cite{Rosenblatt58theperceptron:} algorithm. One should see the latter as a non-linear regression, and feedforward networks schematically stack perceptron layers on top of one another.

\vspace{0.2cm}

We will thus introduce in chapter \ref{sec:chapterFNN} the fundamental building blocks of the simplest neural network layers: weight averaging and activation functions. We will also introduce gradient descent as a way to train the network when joint with the backpropagation algorithm, as a way to minimize a loss function adapted to the task at hand (classification or regression). The more technical details of the backpropagation algorithm are found in the appendix of this chapter, alongside with an introduction to the state of the art feedforward neural network, the ResNet. One can finally find a short matrix description of the feedforward network.

\vspace{0.2cm}

In chapter \ref{sec:chapterCNN}, we present the second type of neural network studied: the convolutional networks, particularly suited to treat images and label them. This implies presenting the mathematical tools related to this network: convolution, pooling, stride... As well as seeing the modification of the building block introduced in chapter \ref{sec:chapterFNN}. Several convolutional architectures are then presented, and the appendices once again detail the difficult steps of the main text.

\vspace{0.2cm}

Chapter \ref{sec:chapterRNN} finally presents the network architecture suited for data with a temporal structure -- as time series for instance, the recurrent neural network. There again, the novelties and the modifications of the material introduced in the two previous chapters are detailed in the main text, while the appendices give all what one needs to understand the most cumbersome formula of this kind of network architecture.

%\part{Theoretical background} \label{Part1}

 \chapter{Feedforward Neural Networks} \label{sec:chapterFNN}

\minitoc

\section{Introduction}

\yinipar{\fontsize{60pt}{72pt}\usefont{U}{Kramer}{xl}{n}I}n this section we review the first type of neural network that has been developed historically: a regular Feedforward Neural Network (FNN). This network does not take into account any particular structure that the input data might have. Nevertheless, it is already a very powerful machine learning tool, especially when used with the state of the art regularization techniques. These techniques -- that we are going to present as well -- allowed to circumvent the training issues that people experienced when dealing with "deep" architectures: namely the fact that neural networks with an important number of hidden states and hidden layers have proven historically to be very hard to train (vanishing gradient and overfitting issues).

\section{FNN architecture}

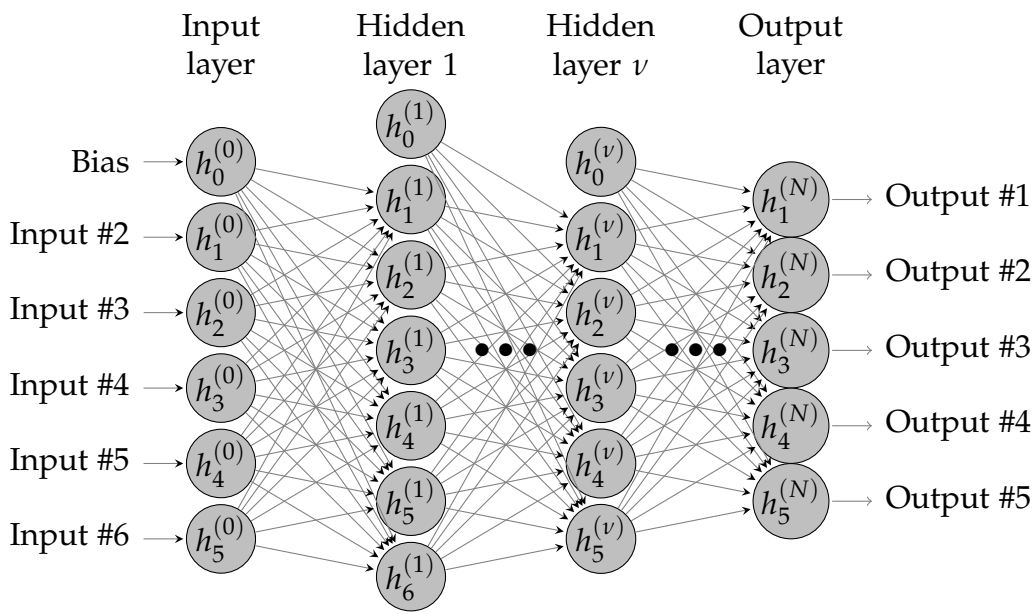
\begin{figure}[H]
\begin{center}
\begin{tikzpicture}[shorten >=1pt,-stealth,draw=black!50, node distance=\layersep]
    \tikzstyle{every pin edge}=[stealth-,shorten <=1pt]
    \tikzstyle{neuron}=[circle,draw=black,fill=black!25,minimum size=17pt,inner sep=0pt]
    \tikzstyle{input neuron}=[neuron, fill=gray!50];
    \tikzstyle{output neuron}=[neuron, fill=gray!50];
    \tikzstyle{hidden neuron}=[neuron, fill=gray!50];
    \tikzstyle{annot} = [text width=4em, text centered]

    % Draw the input layer nodes
    \foreach \name / \y in {1}
       	\pgfmathtruncatemacro{\m}{int(\y-1)}
    % This is the same as writing \foreach \name / \y in {1/1,2/2,3/3,4/4}
        \node[input neuron, pin=left:Bias] (I-\name) at (0,-\y) {$h_{\m}^{(0)}$};

    \foreach \name / \y in {2,...,6}
       	\pgfmathtruncatemacro{\m}{int(\y-1)}
    % This is the same as writing \foreach \name / \y in {1/1,2/2,3/3,4/4}
        \node[input neuron, pin=left:Input \#\y] (I-\name) at (0,-\y) {$h_{\m}^{(0)}$};

    % Draw the hidden layer 1 nodes
    \foreach \name / \y in {1,...,7}
    	\pgfmathtruncatemacro{\m}{int(\y-1)}
        \path[yshift=0.5cm]
            node[hidden neuron] (H1-\name) at (\layersep,-\y cm) {$h_{\m}^{(1)}$};

    % Draw the hidden layer 1 node
    \foreach \name / \y in {1,...,6}
        \pgfmathtruncatemacro{\m}{int(\y-1)}
        \path[yshift=0.0cm]
            node[hidden neuron] (H2-\name) at (2*\layersep,-\y cm) {$h_{\m}^{(\nu)}$};

    % Draw the output layer node
    \foreach \name / \y in {1,...,5}
        \path[yshift=-0.5cm]
    node[output neuron,pin={[pin edge={->}]right:Output \#\y}] (O-\name) at (3*\layersep,-\y cm) {$h_{\y}^{(N)}$};

    % Connect every node in the input layer with every node in the
    % hidden layer.
    \foreach \source in {1,...,6}
        \foreach \dest in {2,...,7}
            \path (I-\source) edge (H1-\dest);

     \foreach \source in {1,...,7}
       \foreach \dest in {2,...,6}
           \path (H1-\source) edge (H2-\dest);

    % Connect every node in the hidden layer with the output layer
    \foreach \source in {1,...,6}
       \foreach \dest in {1,...,5}
          \path (H2-\source) edge (O-\dest);

    % Annotate the layers
    \node[annot,above of=H1-1, node distance=1cm] (hl) {Hidden layer 1};
    \node[annot,left of=hl] {Input layer};
    \node[annot,right of=hl] (hm) {Hidden layer $\nu$};
    \node[annot,right of=hm] {Output layer};

    \node at ((1.5*\layersep,-3.5 cm) {$\bullet\bullet\bullet$};
    \node at ((2.5*\layersep,-3.5 cm) {$\bullet\bullet\bullet$};
\end{tikzpicture}
\caption{\label{fig:1}Neural Network with $N+1$ layers ($N-1$ hidden layers). For simplicity of notations, the index referencing the training set has not been indicated. Shallow architectures use only one hidden layer. Deep learning amounts to take several hidden layers, usually containing the same number of hidden neurons. This number should be on the ballpark of the average of the number of input and output variables.}
\end{center}
\end{figure}

A FNN is formed by one input layer, one (shallow network) or more (deep network, hence the name deep learning) hidden layers and one output layer. Each layer of the network (except the output one) is connected to a following layer. This connectivity is central to the FNN structure and has two main features in its simplest form: a weight averaging feature and an activation feature. We will review these features extensively in the following

\section{Some notations}

In the following, we will call
\begin{itemize}
\item[$\bullet$] $N$ the number of layers (not counting the input) in the Neural Network.
\item[$\bullet$] $T_{{\rm train}}$ the number of training examples in the training set.
\item[$\bullet$] $T_{{\rm mb}}$ the number of training examples in a mini-batch (see section \ref{sec:FNNlossfunction}).
\item[$\bullet$] $t \in \llbracket0,T_{{\rm mb}}-1\rrbracket$ the mini-batch training instance index.
\item[$\bullet$] $\nu\in\llbracket0,N\rrbracket$ the number of layers in the FNN.
\item[$\bullet$] $F_\nu$ the number of neurons in the $\nu$'th layer.
\item[$\bullet$] $X^{(t)}_f=h_{f}^{(0)(t)}$ where $f\in\llbracket0,F_0-1\rrbracket$ the input variables.
\item[$\bullet$] $y^{(t)}_f$ where $f\in[0,F_N-1]$ the output variables (to be predicted).
\item[$\bullet$] $\hat{y}^{(t)}_f$ where $f\in[0,F_N-1]$ the output of the network.
\item[$\bullet$] $\Theta_{f}^{(\nu)f'}$ for $f\in [0,F_{\nu}-1]$, $f'\in [0,F_{\nu+1}-1]$ and $\nu\in[0,N-1]$ the weights matrices
\item[$\bullet$] A bias term can be included. In practice, we will see when talking about the batch-normalization procedure that we can omit it.
\end{itemize}

\section{Weight averaging}

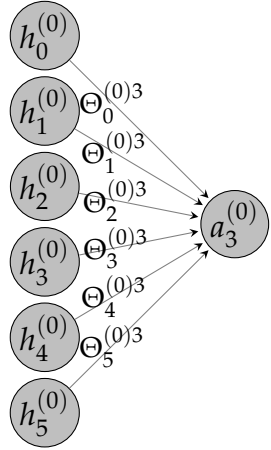
\begin{figure}[H]
\begin{center}
\begin{tikzpicture}[shorten >=1pt,-stealth,draw=black!50, node distance=\layersep]
    \tikzstyle{every pin edge}=[stealth-,shorten <=1pt]
    \tikzstyle{neuron}=[circle,draw=black,fill=black!25,minimum size=17pt,inner sep=0pt]
    \tikzstyle{input neuron}=[neuron, fill=gray!50];
    \tikzstyle{output neuron}=[neuron, fill=gray!50];
    \tikzstyle{hidden neuron}=[neuron, fill=gray!50];
    \tikzstyle{annot} = [text width=4em, text centered]

    % Draw the input layer nodes
    \foreach \name / \y in {1}
       	\pgfmathtruncatemacro{\m}{int(\y-1)}
    % This is the same as writing \foreach \name / \y in {1/1,2/2,3/3,4/4}
        \node[input neuron] (I-\name) at (0,-\y) {$h_{\m}^{(0)}$};

    \foreach \name / \y in {2,...,6}
       	\pgfmathtruncatemacro{\m}{int(\y-1)}
    % This is the same as writing \foreach \name / \y in {1/1,2/2,3/3,4/4}
        \node[input neuron] (I-\name) at (0,-\y) {$h_{\m}^{(0)}$};

    % Draw the hidden layer 1 nodes
    \foreach \name / \y in {4}
    	\pgfmathtruncatemacro{\m}{int(\y-1)}
        \path[yshift=0.5cm]
            node[hidden neuron] (H1-\name) at (\layersep,-\y cm) {$a_{\m}^{(0)}$};

 \path (I-1) edge node[pos=0.3,scale=0.9] {$\Theta^{(0)3}_0$} (H1-4);
 \path (I-2) edge node[pos=0.3,scale=0.9] {$\Theta^{(0)3}_1$} (H1-4);
 \path (I-3) edge node[pos=0.3,scale=0.9] {$\Theta^{(0)3}_2$} (H1-4);
 \path (I-4) edge node[pos=0.3,scale=0.9] {$\Theta^{(0)3}_3$} (H1-4);
 \path (I-5) edge node[pos=0.3,scale=0.9] {$\Theta^{(0)3}_4$} (H1-4);
 \path (I-6) edge node[pos=0.3,scale=0.9] {$\Theta^{(0)3}_5$} (H1-4);

\end{tikzpicture}
\caption{\label{fig:3}Weight averaging procedure.}
\end{center}
\end{figure}

One of the two main components of a FNN is a weight averaging procedure, which amounts to average the previous layer with some weight matrix to obtain the next layer. This is illustrated on the figure \ref{fig:3}

Formally, the weight averaging procedure reads:

\begin{align}
a_{f}^{(t)(\nu)}&=\sum^{F_\nu-1+\epsilon}_{f'=0}\Theta^{(\nu)f}_{\,f'}h^{(t)(\nu)}_{f'}\;,
\end{align}
where $\nu\in\llbracket 0,N-1\rrbracket$, $t \in \llbracket0,T_{{\rm mb}}-1\rrbracket$ and $f\in \llbracket 0,F_{\nu+1}-1\rrbracket$. The $\epsilon$ is here to include or exclude a bias term. In practice, as we will be using batch-normalization, we can safely omit it ($\epsilon=0$ in all the following).

\section{Activation function}

The hidden neuron of each layer is defined as
\begin{align}
h_{f}^{(t)(\nu+1)}&=g\left(a_{f}^{(t)(\nu)}\right)\;,
\end{align}
where $\nu\in\llbracket 0,N-2\rrbracket$, $f\in \llbracket 0,F_{\nu+1}-1\rrbracket$ and as usual $t \in \llbracket0,T_{{\rm mb}}-1\rrbracket$. Here $g$ is an activation function -- the second main ingredient of a FNN -- whose non-linearity allow to predict arbitrary output data. In practice, $g$ is usually taken to be one of the functions described in the following subsections.

\subsection{The sigmoid function}

The sigmoid function takes its value in $]0,1[$ and reads
\begin{align}
g(x)&=\sigma(x)=\frac{1}{1+e^{-x}}\;.
\end{align}
Its derivative is
\begin{align}
\sigma'(x)&=\sigma(x)\left(1-\sigma(x)\right)\;.
\end{align}
This activation function is not much used nowadays (except in RNN-LSTM networks that we will present later in chapter \ref{sec:chapterRNN}).

\begin{figure}[H]
\begin{center}
\begin{tikzpicture}
\node at (0,0) {\includegraphics[scale=1]{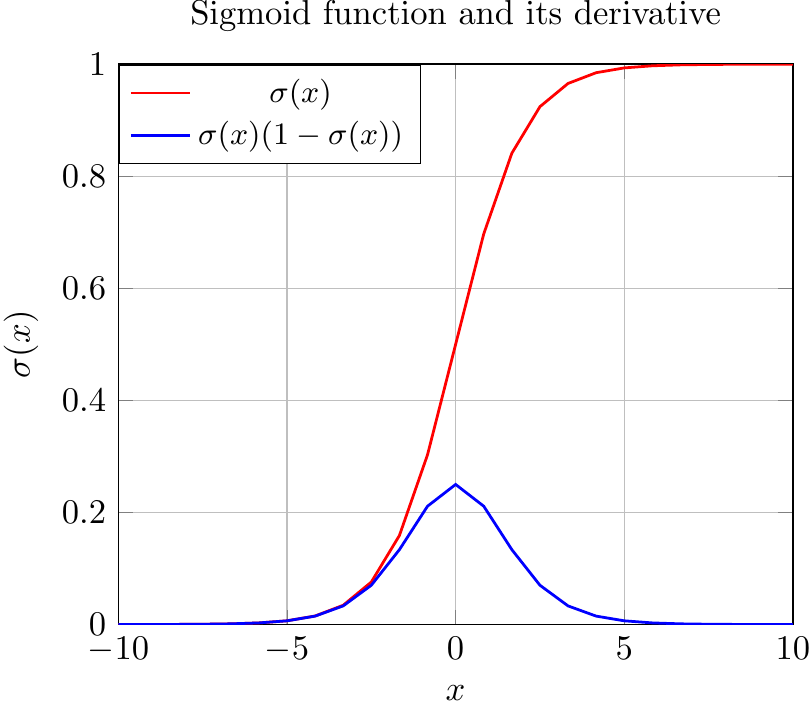}};
\end{tikzpicture}
\end{center}
\caption{\label{fig:sigmoid} the sigmoid function and its derivative.}
\end{figure}

\subsection{The tanh function}

The tanh function takes its value in $]-1,1[$ and reads
\begin{align}
g(x)&=\tanh(x)=\frac{1-e^{-2x}}{1+e^{-2x}}\;.
\end{align}
Its derivative is
\begin{align}
\tanh'(x)&=1-\tanh^2(x)\;.
\end{align}
This activation function has seen its popularity drop due to the use of the activation function presented in the next section.

\begin{figure}[H]
\begin{center}
\begin{tikzpicture}
\node at (0,0) {\includegraphics[scale=1]{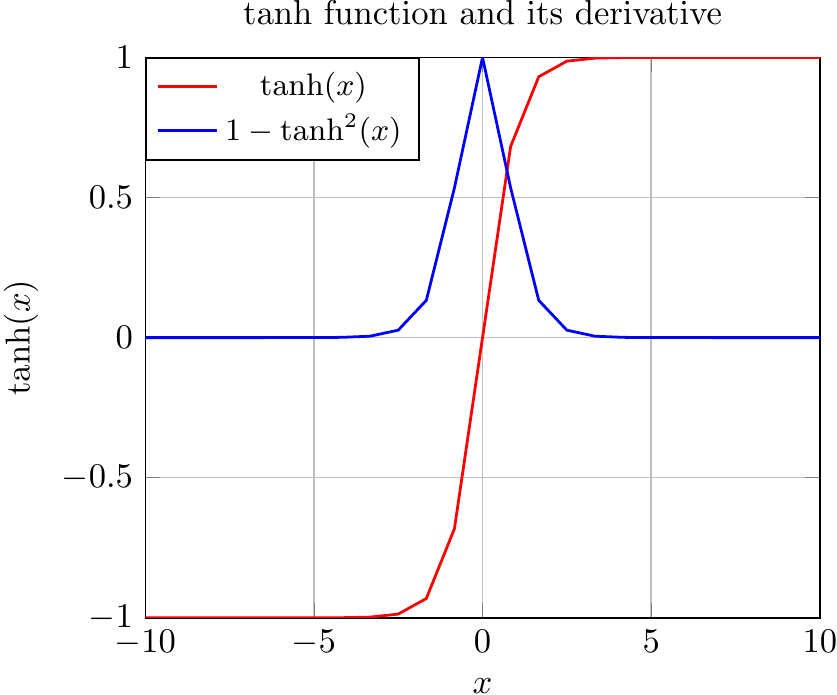}};
\end{tikzpicture}
\end{center}
\caption{\label{fig:tanh} the tanh function and its derivative.}
\end{figure}

It is nevertherless still used in the standard formulation of the RNN-LSTM model (\ref{sec:chapterRNN}).

\subsection{The ReLU function}

The ReLU -- for Rectified Linear Unit -- function takes its value in $[0,+\infty[$ and reads
\begin{align}
g(x)&={\rm ReLU}(x)=\begin{cases}
      x & x\geq 0 \\
      0& x<0
   \end{cases}\;.
\end{align}
Its derivative is
\begin{align}
{\rm ReLU}'(x)&=\begin{cases}
      1 & x\geq 0 \\
      0 & x<0
   \end{cases}\;.
\end{align}

\begin{figure}[H]
\begin{center}
\begin{tikzpicture}
\node at (0,0) {\includegraphics[scale=1]{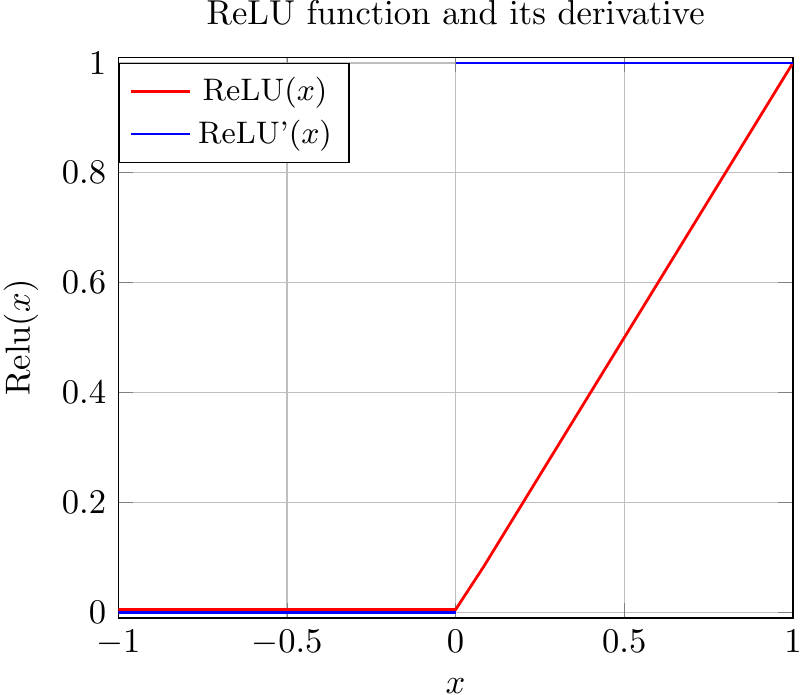}};
\end{tikzpicture}
\end{center}
\caption{\label{fig:relu} the ReLU function and its derivative.}
\end{figure}

This activation function is the most extensively used nowadays. Two of its more common variants can also be found : the leaky ReLU and ELU -- Exponential Linear Unit. They have been introduced because the ReLU activation function tends to "kill" certain hidden neurons: once it has been turned off (zero value), it can never be turned on again.

\subsection{The leaky-ReLU function}

The leaky-ReLU --for Linear Rectified Linear Unit -- function takes its value in $]-\infty,+\infty[$ and is a slight modification of the ReLU that allows non-zero value for the hidden neuron whatever the $x$ value. It reads
\begin{align}
g(x)&={\rm l-ReLU}(x)=\begin{cases}
      x & x\geq 0 \\
      0.01\,x & x<0
   \end{cases}\;.
\end{align}
Its derivative is
\begin{align}
{\rm l-ReLU}'(x)&=\begin{cases}
      1 & x\geq 0 \\
      0.01 & x<0
   \end{cases}\;.
\end{align}

\begin{figure}[H]
\begin{center}
\begin{tikzpicture}
\node at (0,0) {\includegraphics[scale=1]{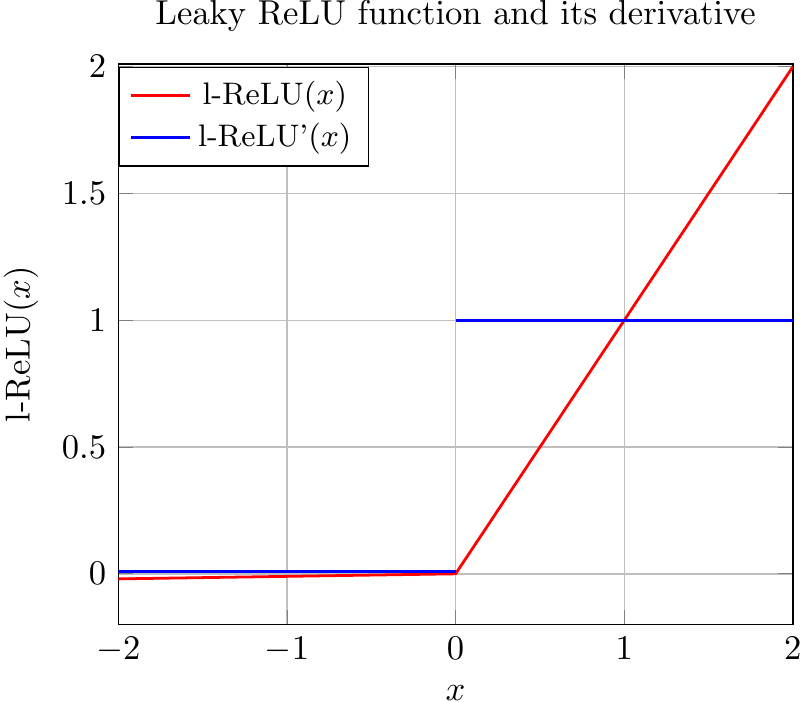}};
\end{tikzpicture}
\end{center}
\caption{\label{fig:lrelu} the leaky-ReLU function and its derivative.}
\end{figure}

A variant of the leaky-ReLU can also be found in the literature : the Parametric-ReLU, where the arbitrary $0.01$ in the definition of the leaky-ReLU is replaced by an $\alpha$ coefficient, that can be
computed via backpropagation.
\begin{align}
g(x)&={\rm Parametric-ReLU}(x)=\begin{cases}
      x & x\geq 0 \\
      \alpha\,x & x<0
   \end{cases}\;.
\end{align}
Its derivative is
\begin{align}
{\rm Parametric-ReLU}'(x)&=\begin{cases}
      1 & x\geq 0 \\
      \alpha & x<0
   \end{cases}\;.
\end{align}

\subsection{The ELU function}

The ELU --for Exponential Linear Unit -- function takes its value between $]-1,+\infty[$ and is inspired by the leaky-ReLU philosophy: non-zero values for all $x$'s. But it presents the advantage of being $\mathcal{C}^1$.
\begin{align}
g(x)&={\rm ELU}(x)=\begin{cases}
      x & x\geq 0 \\
      e^x-1 & x<0
   \end{cases}\;.
\end{align}
Its derivative is
\begin{align}
{\rm ELU}'(x)&=\begin{cases}
      1 & x\geq 0 \\
      e^x & x<0
   \end{cases}\;.
\end{align}

\begin{figure}[H]
\begin{center}
\begin{tikzpicture}
\node at (0,0) {\includegraphics[scale=1]{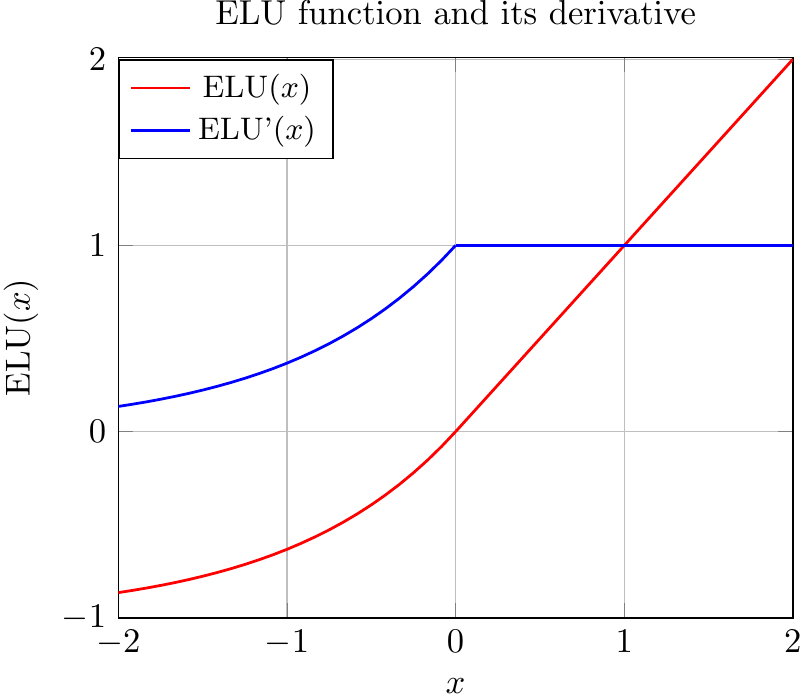}};
\end{tikzpicture}
\end{center}
\caption{\label{fig:elu} the ELU function and its derivative.}
\end{figure}

% From my experience, leay-relu is more than enough.

\section{FNN layers}

As illustrated in figure \ref{fig:1}, a regular FNN is composed by several specific layers. Let us explicit them one by one.

\subsection{Input layer}

The input layer is one of the two places where the data at disposal for the problem at hand come into place. In this chapter, we will be considering a input of size $F_0$, denoted $X^{(t)}_{f}$, with\footnote{
To train the FNN, we jointly compute the forward and backward pass for $T_{{\rm mb}}$ samples of the training set, with $T_{{\rm mb}}\ll T_{{\rm train}}$. In the following we will thus have $t\in \llbracket 0, T_{{\rm mb}}-1\rrbracket$.
}
 $t\in \llbracket 0, T_{{\rm mb}}-1\rrbracket$ (size of the mini-batch, more on that when we will be talking about gradient descent techniques), and $f \in \llbracket 0, F_0-1\rrbracket$. Given the problem at hand, a common procedure could be to center the input following the procedure
\begin{align}
\tilde{X}^{(t)}_{f}&=X^{(t)}_{f}-\mu_{f}\;,
\end{align}
with
\begin{align}
\mu_{f}&=\frac{1}{T_{{\rm train}}}\sum^{T_{{\rm train}}-1}_{t=0}X^{(t)}_{f}\;.
\end{align}
This correspond to compute the mean per data types over the training set. Following our notations, let us recall that
\begin{align}
X^{(t)}_{f}&=h^{(t)(0)}_{f}\;.
\end{align}

\subsection{Fully connected layer}

The fully connected operation is just the conjunction of the weight averaging and the activation procedure. Namely, $\forall \nu\in \llbracket 0,N-1 \rrbracket$
\begin{align}
a_{f}^{(t)(\nu)}&=\sum^{F_\nu-1}_{f'=0}\Theta^{(\nu)f}_{f'}h^{(t)(\nu)}_{f'}\;.\label{eq:Weightavg}
\end{align}
and $\forall \nu\in \llbracket 0,N-2 \rrbracket$
\begin{align}
h_{f}^{(t)(\nu+1)}&=g\left(a_{f}^{(t)(\nu)}\right)\;.
\end{align}
for the case where $\nu=N-1$, the activation function is replaced by an output function.

\subsection{Output layer}

The output of the FNN reads
 \begin{align}
h_{f}^{(t)(N)}&=o(a_{f}^{(t)(N-1)})\;,
\end{align}
where $o$ is called the output function. In the case of the Euclidean loss function, the output function is just the identity. In a classification task, $o$ is the softmax function.
\begin{align}
o\left(a^{(t)(N-1)}_f\right)&=\frac{e^{a^{(t)(N-1)}_f}}{\sum\limits^{F_{N-1}-1}_{f'=0}e^{a^{(t)(N-1)}_{f'}}}
\end{align}

\section{Loss function} \label{sec:FNNlossfunction}

The loss function evaluates the error performed by the FNN when it tries to estimate the data to be predicted (second place where the data make their appearance). For a regression problem, this is simply a mean square error (MSE) evaluation
\begin{align}
J(\Theta)&=\frac{1}{2T_{{\rm mb}}}\sum_{t=0}^{T_{{\rm mb}}-1}\sum_{f=0}^{F_N-1}
\left(y_f^{(t)}-h_{f}^{(t)(N)}\right)^2\;,
\end{align}
while for a classification task, the loss function is called the cross-entropy function
\begin{align}
J(\Theta)&=-\frac{1}{T_{{\rm mb}}}\sum_{t=0}^{T_{{\rm mb}}-1}\sum_{f=0}^{F_N-1}
\delta^f_{y^{(t)}}\ln h_{f}^{(t)(N)}\;,
\end{align}
and for a regression problem transformed into a classification one, calling $C$ the number of bins leads to
\begin{align}
J(\Theta)&=-\frac{1}{T_{{\rm mb}}}\sum_{t=0}^{T_{{\rm mb}}-1}\sum_{f=0}^{F_N-1}\sum_{c=0}^{C-1}
\delta^c_{y_f^{(t)}}\ln h_{fc}^{(t)(N)}\;.
\end{align}
For reasons that will appear clear when talking about the data sample used at each training step, we denote
\begin{align}
J(\Theta)&=\sum_{t=0}^{T_{{\rm mb}}-1}J_{{\rm mb}}(\Theta)\;.
\end{align}

\section{Regularization techniques}

On of the main difficulties when dealing with deep learning techniques is to get the deep neural network to train efficiently. To that end, several regularization techniques have been invented. We will review them in this section

\subsection{L2 regularization}

L2 regularization is the most common regularization technique that on can find in the literature. It amounts to add a regularizing term to the loss function in the following way
\begin{align}
J_{{\rm L2}}(\Theta)&=\lambda_{{\rm L2}} \sum_{\nu=0}^{N-1}\left\|\Theta^{(\nu)}\right\|^2_{{\rm L2}}
=\lambda_{{\rm L2}}\sum_{\nu=0}^{N-1}\sum_{f=0}^{F_{\nu+1}-1}\sum_{f'=0}^{F_\nu-1}
\left(\Theta^{(\nu)f'}_{f}\right)^2\;.\label{eq:l2reg}
\end{align}
This regularization technique is almost always used, but not on its own. A typical value of $\lambda_{{\rm L2}}$ is in the range $10^{-4}-10^{-2}$. Interestingly, this L2 regularization technique has a Bayesian interpretation: it is Bayesian inference with a Gaussian prior on the weights. Indeed, for a given $\nu$, the weight averaging procedure can be considered as
\begin{align}
a_{f}^{(t)(\nu)}&=\sum^{F_\nu-1}_{f'=0}\Theta^{(\nu)f}_{f'}h^{(t)(\nu)}_{f'}+\epsilon\;,
\end{align}
where $\epsilon$ is a noise term of mean $0$ and variance $\sigma^2$. Hence the following Gaussian likelihood for all values of $t$ and $f$:
\begin{align}
\mathcal{N}\left(a_{f}^{(t)(i)}\middle|\sum^{F_\nu-1}_{f'=0}\Theta^{(\nu)f}_{f'}h^{(t)(\nu)}_{f'},\sigma^2\right)\;.
\end{align}
Assuming all the weights to have a Gaussian prior of the form $\mathcal{N}\left(\Theta^{(\nu)f}_{f'}\middle|\lambda_{{\rm L2}}^{-1}\right)$ with the same parameter $\lambda_{{\rm L2}}$, we get the following expression
\begin{align}
\mathcal{P}&=
\prod_{t=0}^{T_{{\rm mb}}-1}\prod_{f=0}^{F_{\nu+1}-1}\left[\mathcal{N}\left(a_{f}^{(t)(\nu)}\middle|
\sum^{F_\nu-1}_{f'=0}\Theta^{(\nu)f}_{f'}h^{(t)(\nu)}_{f'},\sigma^2\right)
\prod_{f'=0}^{F_{\nu}-1}\mathcal{N}\left(\Theta^{(\nu)f}_{f'}
\middle|\lambda_{{\rm L2}}^{-1}\right)\right]\notag\\
&=\prod_{t=0}^{T_{{\rm mb}}-1}\prod_{f=0}^{F_{\nu+1}-1}\left[\frac{1}{\sqrt{2\pi \sigma^2}}
e^{-\frac{\left(a_{f}^{(t)(\nu)}-\sum^{F_i-1}_{f'=0}\Theta^{(\nu)f}_{f'}h^{(t)(\nu)}_{f'}\right)^2}{2\sigma^2}}
\prod_{f'=0}^{F_{\nu}-1}\sqrt{\frac{\lambda_{{\rm L2}}}{2\pi}}e^{-\frac{\left(\Theta^{(\nu)f}_{f'}\right)^2\lambda_{{\rm L2}}}{2}}\right] \;.
\end{align}
Taking the log of it and forgetting most of the constant terms leads to
\begin{align}
\mathcal{L}&\propto\frac{1}{T_{{\rm mb}}\sigma^2}
\sum_{t=0}^{T_{{\rm mb}}-1}\sum_{f=0}^{F_{\nu+1}-1}
\left(a_{f}^{(t)(\nu)}-\sum^{F_\nu-1}_{f'=0}\Theta^{(\nu)f}_{f'}h^{(t)(\nu)}_{f'}\right)^2
+\lambda_{{\rm L2}}\sum_{f=0}^{F_{\nu+1}-1}\sum_{f'=0}^{F_{\nu}-1}\left(\Theta^{(\nu)f}_{f'}\right)^2 \;,
\end{align}
and the last term is exactly the L2 regulator for a given $nu$ value (see formula (\ref{eq:l2reg})).

\subsection{L1 regularization}

L1 regularization amounts to replace the L2 norm by the L1 one in the L2 regularization technique
\begin{align}
J_{{\rm L1}}(\Theta)&=\lambda_{{\rm L1}} \sum_{\nu=0}^{N-1}\left\|\Theta^{(\nu)}\right\|_{{\rm L1}}
=\lambda_{{\rm L1}}\sum_{\nu=0}^{N-1}\sum_{f=0}^{F_{\nu+1}-1}\sum_{f'=0}^{F_\nu-1}
\left|\Theta^{(\nu)f}_{f'}\right|\;.
\end{align}
It can be used in conjunction with L2 regularization, but again these techniques are not sufficient on their own. A typical value of $\lambda_{{\rm L1}}$ is in the range $10^{-4}-10^{-2}$. Following the same line as in the previous section, one can show that L1 regularization is equivalent to Bayesian inference with a Laplacian prior on the weights
\begin{align}
\mathcal{F}\left(\Theta^{(\nu)f}_{f'}\middle| 0,\lambda_{{\rm L1}}^{-1}\right)&=
\frac{\lambda_{{\rm L1}}}{2}e^{-\lambda_{{\rm L1}}\left|\Theta^{(\nu)f}_{f'}\right|}\;.
\end{align}

\subsection{Clipping}

Clipping forbids the L2 norm of the weights to go beyond a pre-determined threshold $C$. Namely after having computed the update rules for the weights, if their L2 norm goes above $C$, it is pushed back to $C$
\begin{align}
{\rm if}\;\left\|\Theta^{(\nu)}\right\|_{{\rm L2}}>C \longrightarrow \Theta^{(\nu)f}_{f'}&=
\Theta^{(\nu)f}_{f'} \times \frac{C}{\left\|\Theta^{(\nu)}\right\|_{{\rm L2}}}\;.
\end{align}

This regularization technique avoids the so-called exploding gradient problem, and is mainly used in RNN-LSTM networks. A typical value of $C$ is in the range $10^{0}-10^{1}$. Let us now turn to the most efficient regularization techniques for a FNN: dropout and Batch-normalization.

\subsection{Dropout}

A simple procedure allows for better backpropagation performance for classification tasks: it amounts to stochastically drop some of the hidden units (and in some instances even some of the input variables) for each training example.

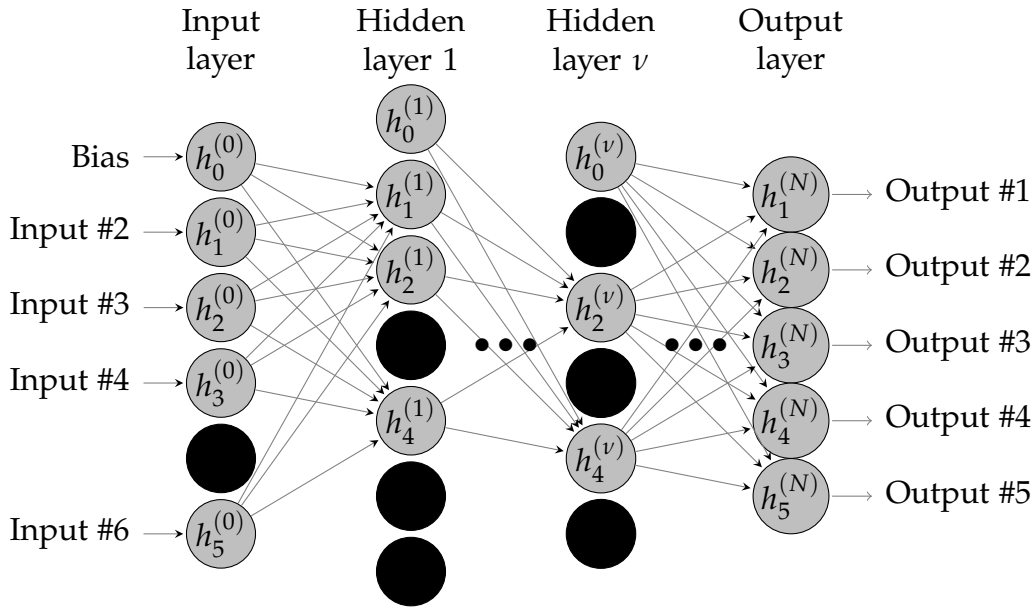
\begin{figure}[H]
\begin{center}
\begin{tikzpicture}[shorten >=1pt,-stealth,draw=black!50, node distance=\layersep]
    \tikzstyle{every pin edge}=[stealth-,shorten <=1pt]
    \tikzstyle{neuron}=[circle,draw=black,fill=black!25,minimum size=17pt,inner sep=0pt]
    \tikzstyle{input neuron}=[neuron, fill=gray!50];
    \tikzstyle{output neuron}=[neuron, fill=gray!50];
    \tikzstyle{dropout neuron}=[neuron, fill=black];
    \tikzstyle{hidden neuron}=[neuron, fill=gray!50];
    \tikzstyle{annot} = [text width=4em, text centered]

    % Draw the input layer nodes
    \foreach \name / \y in {1}
       	\pgfmathtruncatemacro{\m}{int(\y-1)}
    % This is the same as writing \foreach \name / \y in {1/1,2/2,3/3,4/4}
        \node[input neuron, pin=left:Bias] (I-\name) at (0,-\y) {$h_{\m}^{(0)}$};

    \foreach \name / \y in {2,3,4,6}
       	\pgfmathtruncatemacro{\m}{int(\y-1)}
    % This is the same as writing \foreach \name / \y in {1/1,2/2,3/3,4/4}
        \node[input neuron, pin=left:Input \#\y] (I-\name) at (0,-\y) {$h_{\m}^{(0)}$};

        \foreach \name / \y in {5}
       	\pgfmathtruncatemacro{\m}{int(\y-1)}
    % This is the same as writing \foreach \name / \y in {1/1,2/2,3/3,4/4}
        \node[dropout neuron] (I-\name) at (0,-\y) {$h_{\m}^{(0)}$};

    % Draw the hidden layer 1 nodes
    \foreach \name / \y in {1,2,3,5}
    	\pgfmathtruncatemacro{\m}{int(\y-1)}
        \path[yshift=0.5cm]
            node[hidden neuron] (H1-\name) at (\layersep,-\y cm) {$h_{\m}^{(1)}$};

    % Draw the hidden layer 1 nodes
    \foreach \name / \y in {4,6,7}
    	\pgfmathtruncatemacro{\m}{int(\y-1)}
        \path[yshift=0.5cm]
            node[dropout neuron] (H1-\name) at (\layersep,-\y cm) {$h_{\m}^{(1)}$};

    % Draw the hidden layer 1 node
    \foreach \name / \y in {1,3,5}
        \pgfmathtruncatemacro{\m}{int(\y-1)}
        \path[yshift=0.0cm]
            node[hidden neuron] (H2-\name) at (2*\layersep,-\y cm) {$h_{\m}^{(\nu)}$};

    % Draw the hidden layer 1 node
    \foreach \name / \y in {2,4,6}
        \pgfmathtruncatemacro{\m}{int(\y-1)}
        \path[yshift=0.0cm]
            node[dropout neuron] (H2-\name) at (2*\layersep,-\y cm) {$h_{\m}^{(\nu)}$};

    % Draw the output layer node
    \foreach \name / \y in {1,...,5}
        \path[yshift=-0.5cm]
    node[output neuron,pin={[pin edge={->}]right:Output \#\y}] (O-\name) at (3*\layersep,-\y cm) {$h_{\y}^{(N)}$};

    % Connect every node in the input layer with every node in the
    % hidden layer.
    \foreach \source in {1,2,3,4,6}
        \foreach \dest in {2,3,5}
            \path (I-\source) edge (H1-\dest);

     \foreach \source in {1,2,3,5}
       \foreach \dest in {3,5}
           \path (H1-\source) edge (H2-\dest);

    % Connect every node in the hidden layer with the output layer
    \foreach \source in {1,3,5}
       \foreach \dest in {1,...,5}
          \path (H2-\source) edge (O-\dest);

    % Annotate the layers
    \node[annot,above of=H1-1, node distance=1cm] (hl) {Hidden layer 1};
    \node[annot,left of=hl] {Input layer};
    \node[annot,right of=hl] (hm) {Hidden layer $\nu$};
    \node[annot,right of=hm] {Output layer};

    \node at ((1.5*\layersep,-3.5 cm) {$\bullet\bullet\bullet$};
    \node at ((2.5*\layersep,-3.5 cm) {$\bullet\bullet\bullet$};
\end{tikzpicture}
\caption{\label{fig:2}The neural network of figure \ref{fig:1} with dropout taken into account for both the hidden layers and the input. Usually, a different (lower) probability for turning off a neuron is adopted for the input than the one adopted for the hidden layers.}
\end{center}
\end{figure}

This amounts to do the following change: for $\nu\in \llbracket 1,N-1\rrbracket$
\begin{align}
h^{(\nu)}_{f}=\null&m_f^{(\nu)} g\left(a_f^{(\nu)}\right)
\end{align}
with $m_f^{(i)}$ following a $p$ Bernoulli distribution with usually $p=\frac15$ for the mask of the input layer and $p=\frac12$ otherwise. Dropout\cite{Srivastava:2014:DSW:2627435.2670313} has been the most successful regularization technique until the appearance of Batch Normalization.

\subsection{Batch Normalization}

Batch normalization\cite{Ioffe2015} amounts to jointly normalize the mini-batch set per data types, and does so at each input of a FNN layer. In the original paper, the authors argued that this step should be done after the convolutional layers, but in practice it has been shown to be more efficient after the non-linear step. In our case, we will thus consider $\forall i \in \llbracket 0,N-2\rrbracket$
\begin{align}
\tilde{h}_{f}^{(t)(\nu)}&=\frac{h_{f}^{(t)(\nu+1)}-\hat{h}_{f}^{(\nu)}}
{\sqrt{\left(\hat{\sigma}_{f}^{(\nu)}\right)^2+\epsilon}}\;,
\end{align}
with
\begin{align}
\hat{h}_{f}^{(\nu)}&=
\frac{1}{T_{{\rm mb}}}\sum^{T_{{\rm mb}}-1}_{t=0}h_{f}^{(t)(\nu+1)}\\
\left(\hat{\sigma}_{f}^{(\nu)}\right)^2&=\frac{1}{T_{{\rm mb}}}\sum^{T_{{\rm mb}}-1}_{t=0}
\left(h_{f}^{(t)(\nu+1)}-\hat{h}_{f}^{(\nu)}\right)^2\;.
\end{align} To make sure that this transformation can represent the identity transform, we add two additional parameters $(\gamma_f,\beta_f)$ to the model
\begin{align}
y^{(t)(\nu)}_{f}&=\gamma^{(\nu)}_f\,\tilde{h}_{f}^{(t)(\nu)}+\beta^{(\nu)}_f
=\tilde{\gamma}^{(\nu)}_f\,h_{f}^{(t)(\nu)}+\tilde{\beta}^{(\nu)}_f\;.
\end{align}
The presence of the $\beta^{(\nu)}_f$ coefficient is what pushed us to get rid of the bias term, as it is naturally included in batchnorm. During training, one must compute a running sum for the mean and the variance, that will serve for the evaluation of the cross-validation and the test set (calling $e$ the number of iterations/epochs)
\begin{align}
\mathbb{E}\left[h_{f}^{(t)(\nu+1)}\right]_{e+1} &=
\frac{e\mathbb{E}\left[h_{f}^{(t)(\nu)}\right]_{e}+\hat{h}_{f}^{(\nu)}}{e+1}\;,\\
\mathbb{V}ar\left[h_{f}^{(t)(\nu+1)}\right]_{e+1} &=
\frac{e\mathbb{V}ar\left[h_{f}^{(t)(\nu)}\right]_{e}+\left(\hat{\sigma}_{f}^{(\nu)}\right)^2}{e+1}
\end{align}
and what will be used at test time is
\begin{align}
\mathbb{E}\left[h_{f}^{(t)(\nu)}\right]&=\mathbb{E}\left[h_{f}^{(t)(\nu)}\right]\;,&
\mathbb{V}ar\left[h_{f}^{(t)(\nu)}\right]&=
\frac{T_{{\rm mb}}}{T_{{\rm mb}}-1}\mathbb{V}ar\left[h_{f}^{(t)(\nu)}\right]\;.
\end{align}
so that at test time
\begin{align}
y^{(t)(\nu)}_{f}&=\gamma^{(\nu)}_f\,\frac{h_{f}^{(t)(\nu)}-E[h_{f}^{(t)(\nu)}]}{\sqrt{Var\left[h_{f}^{(t)(\nu)}\right]+\epsilon}}+\beta^{(\nu)}_f\;.
\end{align}

In practice, and as advocated in the original paper, on can get rid of dropout without loss of precision when using batch normalization. We will adopt this convention in the following.

\section{Backpropagation}

Backpropagation\cite{LeCun:1998:EB:645754.668382} is the standard technique to decrease the loss function error so as to correctly predict what one needs. As it name suggests, it amounts to backpropagate through the FNN the error performed at the output layer, so as to update the weights. In practice, on has to compute a bunch of gradient terms, and this can be a tedious computational task. Nevertheless, if performed correctly, this is the most useful and important task that one can do in a FN. We will therefore detail how to compute each weight (and Batchnorm coefficients) gradients in the following.

\subsection{Backpropagate through Batch Normalization} \label{sec:Backpropbatchnorm}

Backpropagation introduces a new gradient
\begin{align}
\delta^f_{f'}J^{(tt')(\nu)}_{f}&=\frac{\partial y^{(t')(\nu)}_{f'}}{\partial h_{f}^{(t)(\nu+1)}}\;.
\end{align}
we show in appendix \ref{sec:appenbatchnorm} that
\begin{align}
J^{(tt')(\nu)}_{f}&=\tilde{\gamma}^{(\nu)}_f\ \left[\delta^{t'}_t-
\frac{1+\tilde{h}_{f}^{(t')(\nu)}\tilde{h}_{f}^{(t)(\nu)}}{T_{{\rm mb}}}\right]\;.
\end{align}

\subsection{error updates}

To backpropagate the loss error through the FNN, it is very useful to compute a so-called error rate
\begin{align}
\delta^{(t)(\nu)}_f&= \frac{\partial }{\partial a_{f}^{(t)(\nu)}}J(\Theta)\;,
\end{align}
We show in Appendix \ref{sec:appenbplayers} that $\forall \nu \in \llbracket 0,N-2\rrbracket$
\begin{align}
\delta^{(t)(\nu)}_f&=g'\left(a_{f}^{(t)(\nu)}\right)
\sum_{t'=0}^{T_{{\rm mb}}-1}\sum_{f'=0}^{F_{\nu+1}-1}\Theta^{(\nu+1)f'}_{f}J^{(tt')(\nu)}_{f} \delta^{(t')(\nu+1)}_{f'}\;,
\end{align}
the value of $\delta^{(t)(N-1)}_f$ depends on the loss used. We show also in appendix \ref{sec:appenbpoutput} that for the MSE loss function
\begin{align}
\delta^{(t)(N-1)}_f&= \frac{1}{T_{{\rm mb}}}\left(h_{f}^{(t)(N)}-y_f^{(t)}\right)\;,
\end{align}
and for the cross entropy loss function
\begin{align}
\delta^{(t)(N-1)}_{f}&= \frac{1}{T_{{\rm mb}}}\left(h_{f}^{(t)(N)}-\delta^f_{y^{(t)}}\right)\;.
\end{align}
To unite the notation of chapters \ref{sec:chapterFNN}, \ref{sec:chapterCNN} and \ref{sec:chapterRNN}, we will call
\begin{align}
\mathcal{H}^{(t)(\nu+1)}_{ff'}&=g'\left(a_{f}^{(t)(\nu)}\right)\Theta^{(\nu+1)f'}_{f}\;,
\end{align}
so that the update rule for the error rate reads
\begin{align}
\delta^{(t)(\nu)}_f&=
\sum_{t'=0}^{T_{{\rm mb}}-1}J^{(tt')(\nu)}_{f}\sum_{f'=0}^{F_{\nu+1}-1}\mathcal{H}^{(t)(\nu+1)}_{ff'} \delta^{(t)(\nu+1)}_{f'}\;.
\end{align}

\subsection{Weight update}

Thanks to the computation of the error rates, the derivation of the error rate is straightforward. We indeed get $\forall \nu \in \llbracket 1,N-1\rrbracket$
\begin{align}
\Delta^{\Theta(\nu)f}_{f'}&=\frac{1}{T_{{\rm mb}}}\sum_{t=0}^{T_{{\rm mb}}-1}
\sum^{F_{\nu+1}-1}_{f^{''}=0}\sum^{F_\nu}_{f^{'''}=0}\frac{\partial\Theta^{(\nu)f^{''}}_{f^{'''}}
}{\partial \Theta^{(\nu)f}_{f'}}y^{(t)(\nu-1)}_{f^{'''}}\delta^{(t)(\nu)}_{f^{''}}
=\sum_{t=0}^{T_{{\rm mb}}-1}\delta^{(t)(\nu)}_f y^{(t)(\nu-1)}_{f'}\;.
\end{align}
and
\begin{align}
\Delta^{\Theta(0)f}_{f'}&=\sum_{t=0}^{T_{{\rm mb}}-1}\delta^{(t)(0)}_f h^{(t)(0)}_{f'}\;.
\end{align}

\subsection{Coefficient update}

The update rule for the Batchnorm coefficient can easily be computed thanks to the error rate. It reads
\begin{align}
\Delta_f^{\gamma(\nu)}&=\sum_{t=0}^{T_{{\rm mb}}-1}\sum_{f'=0}^{F_{\nu+1}-1}
\frac{\partial a^{(t)(\nu+1)}_{f'}}{\partial\gamma_f^{(i)}}\delta^{(t)(\nu+1)}_{f'}
=\sum_{t=0}^{T_{{\rm mb}}-1}\sum_{f'=0}^{F_{\nu+1}-1}
\Theta^{(\nu+1)f'}_{f}\tilde{h}^{(t)(i)}_{f}\delta^{(t)(\nu+1)}_{f'}\;,\\
\Delta_f^{\beta(\nu)}&=\sum_{t=0}^{T_{{\rm mb}}-1}\sum_{f'=0}^{F_{\nu+1}-1}
\frac{\partial a^{(t)(\nu+1)}_{f'}}{\partial\beta_f^{(i)}}\delta^{(t)(\nu+1)}_{f'}
=\sum_{t=0}^{T_{{\rm mb}}-1}\sum_{f'=0}^{F_{\nu+1}-1}\Theta^{(\nu+1)f'}_{f}\delta^{(t)(\nu+1)}_{f'}\;,
\end{align}

\section{Which data sample to use for gradient descent?}

From the beginning we have denoted $T_{{\rm mb}}$ the sample of the data from which we train our model. This procedure is repeated a large number of time (each time is called an epoch). But in the literature there exists three way to sample from the data: Full-batch, Stochastic and Mini-batch gradient descent. We explicit these terms in the following sections.

\subsection{Full-batch}

Full-batch takes the whole training set at each epoch, such that the loss function reads
\begin{align}
J(\Theta)&=\sum_{t=0}^{T_{{\rm train}}-1}J_{{\rm train}}(\Theta)\;.
\end{align}
This choice has the advantage to be numerically stable, but it so costly in computation time that it is rarely if ever used.

\subsection{Stochastic Gradient Descent (SGD)}

SGD amounts to take only one exemplary of the training set at each epoch
\begin{align}
J(\Theta)&=J_{{\rm SGD}}(\Theta)\;.
\end{align}
This choice leads to faster computations, but is so numerically unstable that the most standard choice by far is Mini-batch gradient descent.

\subsection{Mini-batch}

Mini-batch gradient descent is a compromise between stability and time efficiency, and is the middle-ground between Full-batch and Stochastic gradient descent: $1\ll T_{{\rm mb}}\ll T_{{\rm train}}$. Hence
\begin{align}
J(\Theta)&=\sum_{t=0}^{T_{{\rm mb}}-1}J_{{\rm mb}}(\Theta)\;.
\end{align}
All the calculations in this note have been performed using this gradient descent technique.

\section{Gradient optimization techniques}

Once the gradients for backpropagation have been computed, the question of how to add them to the existing weights arise. The most natural choice would be to take
\begin{align}
\Theta^{(\nu)f}_{f'}&=\Theta^{(\nu)f}_{f'}-\eta\Delta^{\Theta(i)f}_{f'}\;.
\end{align}
where $\eta$ is a free parameter that is generally initialized thanks to cross-validation. It can also be made epoch dependent (with usually a slow exponentially decaying behaviour). When using Mini-batch gradient descent, this update choice for the weights presents the risk of having the loss function being stuck in a local minimum. Several method have been invented to prevent this risk. We are going to review them in the next sections.

\subsection{Momentum}

Momentum\cite{QIAN1999145} introduces a new vector $v_{{\rm e}}$ and can be seen as keeping a memory of what where the previous updates at prior epochs. Calling $e$ the number of epochs and forgetting the $f,f',\nu$ indices for the gradients to ease the notations, we have
\begin{align}
v_{{\rm e}}&=\gamma v_{{\rm e-1}}+\eta \Delta^{\Theta}\;,
\end{align}
and the weights at epoch $e$ are then updated as
\begin{align}
\Theta_e&=\Theta_{e-1}-v_{{\rm e}}\;.
\end{align}
$\gamma$ is a new parameter of the model, that is usually set to $0.9$ but that could also be fixed thanks to cross-validation.

\subsection{Nesterov accelerated gradient}

Nesterov accelerated gradient\cite{nesterov1983method} is a slight modification of the momentum technique that allows the gradients to escape from local minima. It amounts to take
\begin{align}
v_{{\rm e}}&=\gamma v_{{\rm e-1}}+\eta \Delta^{\Theta-\gamma v_{{\rm e-1}}}\;,
\end{align}
and then again
\begin{align}
\Theta_e&=\Theta_{e-1}-v_{{\rm e}}\;.
\end{align}
Until now, the parameter $\eta$ that controls the magnitude of the update has been set globally. It would be nice to have a fine control of it, so that different weights can be updated with different magnitudes.

\subsection{Adagrad}

Adagrad\cite{Duchi:2011:ASM:1953048.2021068} allows to fine tune the different gradients by having individual learning rates $\eta$. Calling for each value of $f,f',i$
\begin{align}
v_{{\rm e}}&=\sum_{e'=0}^{e-1} \left(\Delta^{\Theta}_{e'}\right)^2\;,
\end{align}
the update rule then reads
\begin{align}
\Theta_e&=\Theta_{e-1}-\frac{\eta}{\sqrt{v_{{\rm e}}+\epsilon}}\Delta^{\Theta}_{e}\;.
\end{align}
One advantage of Adagrad is that the learning rate $\eta$ can be set once and for all (usually to $10^{-2}$) and does not need to be fine tune via cross validation anymore, as it is individually adapted to each weight via the $v_{{\rm e}}$ term. $\epsilon$ is here to avoid division by 0 issues, and is usually set to $10^{-8}$.

\subsection{RMSprop}

Since in Adagrad one adds the gradient from the first epoch, the weight are forced to monotonically decrease. This behaviour can be smoothed via the Adadelta technique, which takes
\begin{align}
v_{{\rm e}}&=\gamma v_{{\rm e-1}}+(1-\gamma )\Delta^{\Theta}_{e}\;,
\end{align}
with $\gamma$ a new parameter of the model, that is usually set to $0.9$. The Adadelta update rule then reads as the Adagrad one
\begin{align}
\Theta_e&=\Theta_{e-1}-\frac{\eta}{\sqrt{v_{{\rm e}}+\epsilon}}\Delta^{\Theta}_{e}\;.
\end{align}
$\eta$ can be set once and for all (usually to $10^{-3}$).

\subsection{Adadelta}

Adadelta\cite{journals/corr/abs-1212-5701} is an extension of RMSprop, that aims at getting rid of the $\eta$ parameter. To do so, a new vector update is introduced
\begin{align}
m_{{\rm e}}&=\gamma m_{{\rm e-1}}+(1-\gamma )
\left(\frac{\sqrt{m_{{\rm e-1}}+\epsilon}}{\sqrt{v_{{\rm e}}+\epsilon}}\Delta^{\Theta}_{e}\right)^2\;,
\end{align}
and the new update rule for the weights reads
\begin{align}
\Theta_e&=\Theta_{e-1}-\frac{\sqrt{m_{{\rm e-1}}+\epsilon}}{\sqrt{v_{{\rm e}}+\epsilon}}\Delta^{\Theta}_{e}\;.
\end{align}
The learning rate has been completely eliminated from the update rule, but the procedure for doing so is ad hoc. The next and last optimization technique presented seems more natural and is the default choice on a number of deep learning algorithms.
\subsection{Adam}

Adam\cite{Kingma2014} keeps track of both the gradient and its square via two epoch dependent vectors
\begin{align}
m_{{\rm e}}&= \beta_1 m_{{\rm e-1}}+ (1-\beta_1)\Delta^{\Theta}_{e}\;,&
v_{{\rm e}}&= \beta_2 v_{{\rm e}}+ (1-\beta_2)\left(\Delta^{\Theta}_{e}\right)^2\;,
\end{align}
with $\beta_1$ and $\beta_2$ parameters usually respectively set to $0.9$ and $0.999$. But the robustness and great strength of Adam is that it makes the whole learning process weakly dependent of their precise value. To avoid numerical problems during the first steps, these vector are rescaled
\begin{align}
\hat{m}_{{\rm e}}&= \frac{m_{{\rm e}}}{1-\beta_1^{e}}\;,&
\hat{v}_{{\rm e}}&= \frac{v_{{\rm e}}}{1-\beta_2^{e}}\;.
\end{align}
before entering into the update rule
\begin{align}
\Theta_e&=\Theta_{e-1}-\frac{\eta }{\sqrt{\hat{v}_{{\rm e}}+\epsilon}}\hat{m}_{{\rm e}}\;.
\end{align}
This is the optimization technique implicitly used throughout this note, alongside with a learning rate decay
\begin{align}
\eta_e&=e^{-\alpha_0}\eta_{e-1}\;,
\end{align}
$\alpha_0$ determined by cross-validation, and $\eta_0$ usually initialized in the range $10^{-3}-10^{-2}$.

\section{Weight initialization}

Without any regularization, training a neural network can be a daunting task because of the fine-tuning of the weight initial conditions. This is one of the reasons why neural networks have experienced out of mode periods. Since dropout and Batch normalization, this issue is less pronounced, but one should not initialize the weight in a symmetric fashion (all zero for instance), nor should one initialize them too large. A good heuristic is
\begin{align}
\left[\Theta^{(\nu)f'}_f\right]_{{\rm init}}&=\sqrt{\frac{6}{F_i+F_{i+1}}}\times\mathcal{N}(0,1)\;.
\end{align}

\begin{subappendices}
\section{Backprop through the output layer} \label{sec:appenbpoutput}

Recalling the MSE loss function
\begin{align}
J(\Theta)&=\frac{1}{2T_{{\rm mb}}}\sum_{t=0}^{T_{{\rm mb}}-1}\sum_{f=0}^{F_N-1}
\left(y_f^{(t)}-h_{f}^{(t)(N)}\right)^2\;,
\end{align}
we instantaneously get
\begin{align}
\delta^{(t)(N-1)}_f&= \frac{1}{T_{{\rm mb}}}\left(h_{f}^{(t)(N)}-y_f^{(t)}\right)\;.
\end{align}
Things are more complicated for the cross-entropy loss function of a regression problem transformed into a multi-classification task.
Assuming that we have $C$ classes for all the values that we are trying to predict, we get
\begin{align}
\delta^{(t)(N-1)}_{fc}&= \frac{\partial }{\partial a_{fc}^{(t)(N-1)}}J(\Theta)
=\sum_{t'=0}^{T_{{\rm mb}}-1}\sum_{f'=0}^{F_N-1}\sum_{d=0}^{C-1}
\frac{\partial h_{f'd}^{(t')(N)}}{\partial a_{fc}^{(t)(N-1)}}
 \frac{\partial }{\partial h_{f'd}^{(t')(N)}}J(\Theta)\;.
\end{align}
Now
\begin{align}
 \frac{\partial }{\partial h_{f'd}^{(t')(N)}}J(\Theta)&=-\frac{\delta^{d}_{ y_{f'}^{(t')}}}{T_{{\rm mb}} h_{f'd}^{(t')(N)}}\;,
\end{align}
and
\begin{align}
\frac{\partial h_{f'd}^{(t')(N)}}{\partial a_{fc}^{(t)(N-1)}}&=
\delta^f_{f'}\delta^{t}_{t'} \left(\delta^c_d h_{fc}^{(t)(N)}- h_{fc}^{(t)(N)} h_{fd}^{(t)(N)}\right)\;,
\end{align}
so that
\begin{align}
\delta^{(t)(N-1)}_{fc}&=-\frac{1}{T_{{\rm mb}}} \sum_{d=0}^{C-1}\frac{\delta^{d}_{ y_f^{(t)}}}{h_{fd}^{(t)(N)}}
\left(\delta^c_d h_{fc}^{(t)(N)}- h_{fc}^{(t)(N)} h_{fd}^{(t)(N)}\right)\notag\\
&=\frac{1}{T_{{\rm mb}}}\left( h_{fc}^{(t)(N)}-\delta^{c}_{ y_f^{(t)}}\right)\;.
\end{align}
For a true classification problem, we easily deduce
\begin{align}
\delta^{(t)(N-1)}_{fc}&=\frac{1}{T_{{\rm mb}}}\left( h_{f}^{(t)(N)}-\delta^{f}_{ y^{(t)}}\right)\;.
\end{align}

\section{Backprop through hidden layers} \label{sec:appenbplayers}

To go further we need
\begin{align}
\delta^{(t)(\nu)}_f&= \frac{\partial }{\partial a_{f}^{(t)(\nu)}}J^{(t)}(\Theta)=
\sum_{t'=0}^{T_{{\rm mb}}-1}\sum_{f'=0}^{F_{\nu+1}-1}
 \frac{\partial a_{f'}^{(t')(\nu+1)}}{\partial a_{f}^{(t)(\nu)}} \delta^{(t')(\nu+1)}_{f'}\notag\\
&=\sum_{t'=0}^{T_{{\rm mb}}-1}\sum_{f'=0}^{F_{\nu+1}-1}\sum^{F_\nu}_{f''=0}\Theta^{(\nu+1)f'}_{f''}
\frac{\partial y^{(t')(\nu)}_{f''} }{\partial a_{f}^{(t)(\nu)}} \delta^{(t')(\nu+1)}_{f'}\notag\\
&=\sum_{t'=0}^{T_{{\rm mb}}-1}\sum_{f'=0}^{F_{\nu+1}-1}\sum^{F_\nu}_{f''=0}\Theta^{(\nu+1)f'}_{f''}
\frac{\partial y^{(t')(\nu)}_{f''} }{\partial h_{f}^{(t)(\nu+1)}}
g'\left(a_{f}^{(t)(\nu)}\right) \delta^{(t')(\nu+1)}_{f'}\;,
\end{align}
so that
\begin{align}
\delta^{(t)(\nu)}_f&=g'\left(a_{f}^{(t)(\nu)}\right)
\sum_{t'=0}^{T_{{\rm mb}}-1}\sum_{f'=0}^{F_{\nu+1}-1}\Theta^{(\nu+1)f'}_{f}J^{(tt')(\nu)}_{f} \delta^{(t)(\nu+1)}_{f'}\;,
\end{align}

\section{Backprop through BatchNorm} \label{sec:appenbatchnorm}

We saw in section \ref{sec:Backpropbatchnorm} that batch normalization implies among other things to compute the following gradient.
\begin{align}
\frac{\partial y^{(t')(\nu)}_{f'}}{\partial h_{f}^{(t)(\nu+1)}}&=
\gamma^{(\nu)}_f\frac{\partial \tilde{h}_{f'}^{(t)(\nu)}}{\partial h_{f}^{(t)(\nu+1)}}\;.
\end{align}
We propose to do just that in this section. Firstly
\begin{align}
\frac{\partial h^{(t')(\nu+1)}_{f'}}{\partial h_{f}^{(t)(\nu+1)}}&=\delta^{t'}_t\delta^{f'}_f\;,&
\frac{\partial \hat{h}_{f'}^{(\nu)}}{\partial h_{f}^{(t)(\nu+1)}}&=\frac{\delta^{f'}_f}{T_{{\rm mb}}}\;.
\end{align}
Secondly
\begin{align}
\frac{\partial \left(\hat{\sigma}_{f'}^{(\nu)}\right)^2}{\partial h_{f}^{(t)(\nu+1)}}&=
\frac{2\delta^{f'}_f}{T_{{\rm mb}}}\left(h_{f}^{(t)(\nu+1)}-\hat{h}_{f}^{(\nu)}\right)\;,
\end{align}
so that we get
\begin{align}
\frac{\partial \tilde{h}_{f'}^{(t)(\nu)}}{\partial h_{f}^{(t)(\nu+1)}}&=
\frac{\delta^{f'}_f}{T_{{\rm mb}}}\left[\frac{T_{{\rm mb}}\delta^{t'}_t-1}
{\left(\left(\hat{\sigma}_{f}^{(\nu)}\right)^2+\epsilon\right)^\frac12}-
\frac{\left(h_{f}^{(t')(\nu+1)}-\hat{h}_{f}^{(\nu)}\right)\left(h_{f}^{(t)(\nu+1)}-\hat{h}_{f}^{(\nu)}\right)}
{\left(\left(\hat{\sigma}_{f}^{(\nu)}\right)^2+\epsilon\right)^\frac32}\right]\notag\\
&=\frac{\delta^{f'}_f}{\left(\left(\hat{\sigma}_{f}^{(\nu)}\right)^2+\epsilon\right)^\frac12}
\left[\delta^{t'}_t-
\frac{1+\tilde{h}_{f}^{(t')(\nu)}\tilde{h}_{f}^{(t)(\nu)}}{T_{{\rm mb}}}\right]\;.
\end{align}
To ease the notation recall that we denoted
\begin{align}
\tilde{\gamma}^{(\nu)}_f&=
\frac{\gamma^{(\nu)}_f}{\left(\left(\hat{\sigma}_{f}^{(\nu)}\right)^2+\epsilon\right)^\frac12}\;.
\end{align}
so that
\begin{align}
\frac{\partial y_{f'}^{(t)(\nu)}}{\partial h_{f}^{(t)(\nu+1)}}&=
\tilde{\gamma}^{(\nu)}_f \delta^{f'}_f\left[\delta^{t'}_t-
\frac{1+\tilde{h}_{f}^{(t')(\nu)}\tilde{h}_{f}^{(t)(\nu)}}{T_{{\rm mb}}}\right]\;.
\end{align}

\section{FNN ResNet (non standard presentation)} \label{sec:ResnetFNN}

The state of the art architecture of convolutional neural networks (CNN, to be explained in chapter \ref{sec:chapterCNN}) is called ResNet\cite{He2015}. Its name comes from its philosophy: each hidden layer output $y$ of the network is a small -- hence the term residual -- modification of its input ($y=x+F(x)$), instead of a total modification ($y=H(x)$) of its input $x$. This philosophy can be imported to the FNN case. Representing the operations of weight averaging, activation function and batch normalization in the following way

\begin{figure}[H]
\begin{center}
\begin{tikzpicture}
\node at (0,0) {\includegraphics[scale=1]{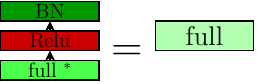}};
\end{tikzpicture}
\end{center}
\caption{\label{fig:fc_equiv} Schematic representation of one FNN fully connected layer.}
\end{figure}

In its non standard form presented in this section, the residual operation amounts to add a skip connection to two consecutive full layers

\begin{figure}[H]
\begin{center}
\begin{tikzpicture}
\node at (0,0) {\includegraphics[scale=1]{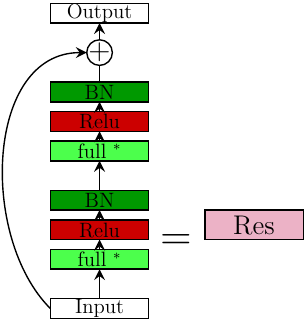}};
\end{tikzpicture}
\end{center}
\caption{\label{fig:fc_resnet_2} Residual connection in a FNN.}
\end{figure}

Mathematically, we had before (calling the input $y^{(t)(\nu-1)}$)

\begin{align}
y_{f}^{(t)(\nu+1)}&=\gamma_f^{(\nu+1)}\tilde{h}_f^{(t)(\nu+2)}+\beta_f^{(\nu+1)}\;,&
a_{f}^{(t)(\nu+1)}&=\sum^{F_{\nu}-1}_{f'=0}\Theta^{(\nu+1)f}_{f'}y_{f}^{(t)(\nu)}\notag\\
y_{f}^{(t)(\nu)}&=\gamma_f^{(\nu)}\tilde{h}_f^{(t)(\nu+1)}+\beta_f^{(\nu)}\;,&
a_{f}^{(t)(\nu)}&=\sum^{F_{\nu-1}-1}_{f'=0}\Theta^{(\nu)f}_{f'}y_{f}^{(t)(\nu-1)}\;,
\end{align}
as well as $h^{(t)(\nu+2)}_f=g\left(a_{f}^{(t)(\nu+1)}\right)$ and $h^{(t)(\nu+1)}_f=g\left(a_{f}^{(t)(\nu)}\right)$. In ResNet, we now have the slight modification
\begin{align}
y_{f}^{(t)(\nu+1)}&=\gamma_f^{(\nu+1)}\tilde{h}_f^{\nu+2}+\beta_f^{(\nu+1)}+y^{(t)(\nu-1)}_{f}\;.
\end{align}
The choice of skipping two and not just one layer has become a standard for empirical reasons, so as the decision not to weight the two paths (the trivial skip one and the two FNN layer one) by a parameter to be learned by backpropagation
\begin{align}
y_{f}^{(t)(\nu+1)}&=\alpha\left(\gamma_f^{(\nu+1)}\tilde{h}_f^{(t)(\nu+2)}+\beta_f^{(\nu+1)}\right)
+\left( 1-\alpha\right)y^{(t)(\nu-1)}_{f'}\;.
\end{align}
This choice is called highway nets\cite{citeulike:14070430}, and it remains to be theoretically understood why it leads to worse performance than ResNet, as the latter is a particular instance of the former. Going back to the ResNet backpropagation algorithm, this changes the gradient through the skip connection in the following way
\begin{align}
\delta^{(t)(\nu-1)}_f&=
\sum_{t'=0}^{T_{{\rm mb}}-1}\sum_{f'=0}^{F_{\nu}-1}
 \frac{\partial a_{f'}^{(t')(\nu)}}{\partial a_{f}^{(t)(\nu-1)}} \delta^{(t')(\nu)}_{f'}
+\sum_{t'=0}^{T_{{\rm mb}}-1}\sum_{f'=0}^{F_{\nu+2}-1}
 \frac{\partial a_{f'}^{(t')(\nu+2)}}{\partial a_{f}^{(t)(\nu-1)}} \delta^{(t')(\nu+2)}_{f'}\notag\\
&=\sum_{t'=0}^{T_{{\rm mb}}-1}\sum_{f'=0}^{F_{\nu}-1}\sum^{F_{\nu-1}-1}_{f''=0}\Theta^{(\nu)f'}_{f''}
\frac{\partial y^{(t')(\nu-1)}_{f''} }{\partial a_{f}^{(t)(\nu-1)}} \delta^{(t')(\nu)}_{f'}\notag\\
&+\sum_{t'=0}^{T_{{\rm mb}}-1}\sum_{f'=0}^{F_{\nu+2}-1}\sum^{F_{\nu-1}-1}_{f''=0}\Theta^{(\nu+2)f'}_{f''}
\frac{\partial y^{(t')(\nu+1)}_{f''} }{\partial a_{f}^{(t)(\nu-1)}} \delta^{(t')(\nu+2)}_{f'}\notag\\
&=g'\left(a_{f}^{(t)(\nu-1)}\right)
\sum_{t'=0}^{T_{{\rm mb}}-1}\sum_{f'=0}^{F_{\nu}-1}\sum^{F_{\nu-1}-1}_{f''=0}\Theta^{(\nu)f'}_{f''}
J^{(tt')(\nu)}_{f} \delta^{(t')(\nu)}_{f'}\notag\\
&+g'\left(a_{f}^{(t)(\nu-1)}\right)
\sum_{t'=0}^{T_{{\rm mb}}-1}\sum_{f'=0}^{F_{\nu+2}-1}\sum^{F_{\nu-1}-1}_{f''=0}\Theta^{(\nu+2)f'}_{f''}
J^{(tt')(\nu)}_{f} \delta^{(t')(\nu+2)}_{f'}\;,
\end{align}
so that
\begin{align}
\delta^{(t)(\nu-1)}_f&=g'\left(a_{f}^{(t)(\nu-1)}\right)
\sum_{t'=0}^{T_{{\rm mb}}-1}\sum^{F_{\nu-1}-1}_{f''=0}J^{(tt')(\nu)}_{f}\notag\\
&\times\left[\sum_{f'=0}^{F_{\nu}-1}\Theta^{(\nu)f'}_{f''}\delta^{(t')(\nu)}_{f'}+
\sum_{f'=0}^{F_{\nu+2}-1}\Theta^{(\nu+2)f'}_{f''}\delta^{(t')(\nu+2)}_{f'}\right]\;.
\end{align}

This formulation has one advantage: it totally preserves the usual FNN layer structure of a weight averaging (WA) followed by an activation function (AF) and then a batch normalization operation (BN). It nevertheless has one disadvantage: the backpropagation gradient does not really flow smoothly from one error rate to the other. In the following section we will present the standard ResNet formulation of that takes the problem the other way around : it allows the gradient to flow smoothly at the cost of "breaking" the natural FNN building block.

\section{FNN ResNet (more standard presentation)} \label{sec:ResnetFNN2}

\begin{figure}[H]
\begin{center}
\begin{tikzpicture}
\node at (0,0) {\includegraphics[scale=1]{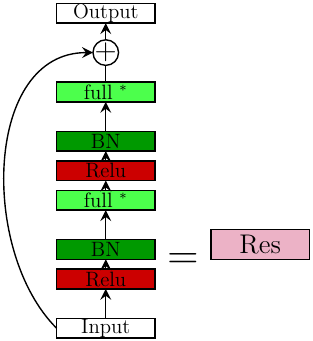}};
\end{tikzpicture}
\end{center}
\caption{\label{fig:fc_resnet_3} Residual connection in a FNN, trivial gradient flow through error rates.}
\end{figure}

In the more standard form of ResNet, the skip connections reads
\begin{align}
 a_{f}^{(t)(\nu+2)}&= a_{f}^{(t)(\nu+2)}+ a_{f}^{(t)(\nu)}\;,
\end{align}
and the updated error rate reads
\begin{align}
\delta^{(t)(\nu)}_f&=g'\left(a_{f}^{(t)(\nu)}\right)
\sum_{t'=0}^{T_{{\rm mb}}-1}\sum^{F_{\nu}-1}_{f''=0}J^{(tt')(\nu)}_{f}
\sum_{f'=0}^{F_{\nu+1}-1}\Theta^{(\nu+1)f'}_{f''}\delta^{(t')(\nu+1)}_{f'}+\delta^{(t')(\nu+2)}_{f}\;.
\end{align}

\section{Matrix formulation}

In all this chapter, we adopted an "index" formulation of the FNN. This has upsides and downsides. On the positive side, one can take the formula as written here and go implement them. On the downside, they can be quite cumbersome to read.

\vspace{0.2cm}

Another FNN formulation is therefore possible: a matrix one. To do so, one has to rewrite
\begin{align}
h_f^{(t)(\nu)}\mapsto h^{(\nu)}_{ft}&\mapsto h^{(\nu)}\in \mathcal{M}(F_\nu,T_{\rm mb})\;.
\end{align}
In this case the weight averaging procedure (\ref{eq:Weightavg}) can be written as
\begin{align}
a_f^{(t)(\nu)}=\sum_{f'=0}^{F_\nu-1}\Theta^{(\nu)f}_{f'}h^{(\nu)}_{f't}&\mapsto a^{(\nu)}=\Theta^{(\nu)}h^{(\nu)}\;.
\end{align}
The upsides and downsides of this formulation are the exact opposite of the index one: what we gained in readability, we lost in terms of direct implementation in low level programming languages (C for instance). For FNN, one can use a high level programming language (like python), but this will get quite intractable when we talk about Convolutional networks. Since the whole point of the present work was to introduce the index notation, and as one can easily find numerous derivation of the backpropagation update rules in matrix form, we will stick with the index notation in all the following, and now turn our attention to convolutional networks.
\end{subappendices}

\chapter{Convolutional Neural Networks} \label{sec:chapterCNN}

\minitoc

\section{Introduction}

\yinipar{\fontsize{60pt}{72pt}\usefont{U}{Kramer}{xl}{n}I}n this chapter we review a second type of neural network that is presumably the most popular one: Convolutional Neural Networks (CNN). CNN are particularly adapted for image classification, be it numbers or animal/car/... category. We will review the novelty involved when dealing with CNN when compared to FNN. Among them are the fundamental building blocks of CNN: convolution and pooling. We will in addition see what modification have to be taken into account for the regularization techniques introduced in the FNN part. Finally, we will present the most common CNN architectures that are used in the literature: from LeNet to ResNet.

\section{CNN architecture}

A CNN is formed by several convolution and pooling operations, usually followed by one or more fully connected layers (those being similar to the traditional FNN layers). We will clarify the new terms introduced thus far in the following sections.

\begin{figure}[H]
\begin{center}
\begin{tikzpicture}
\node[] at (0,0) {\includegraphics[scale=0.5]{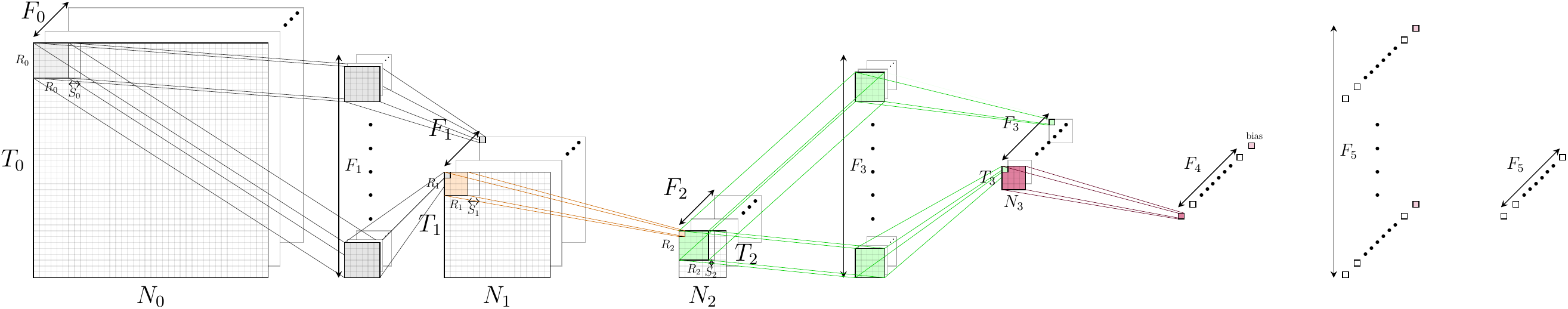}};
\end{tikzpicture}
\caption{\label{fig:lenet-CNN}A typical CNN architecture (in this case LeNet inspired): convolution operations are followed by pooling operations, until the size of each feature map is reduced to one. Fully connected layers can then be introduced.}
\end{center}
\end{figure}

\section{CNN specificities}

\subsection{Feature map}

In each layer of a CNN, the data are no longer labeled by a single index as in a FNN. One should see the FNN index as equivalent to the label of a given image in a layer of a CNN. This label is the feature map. In each feature map $f\in \llbracket 0,F_\nu-1\rrbracket$ of the $\nu$'th layer, the image is fully characterized by two additional indices corresponding to its height $k\in T_\nu-1$ and its width $j\in N_\nu-1$. A given $f,j,k$ thus characterizes a unique pixel of a given feature map. Let us now review the different layers of a CNN

\subsection{Input layer}

\begin{figure}[H]
\begin{center}
\begin{tikzpicture}
\node[] at (0,0) {\includegraphics[scale=1]{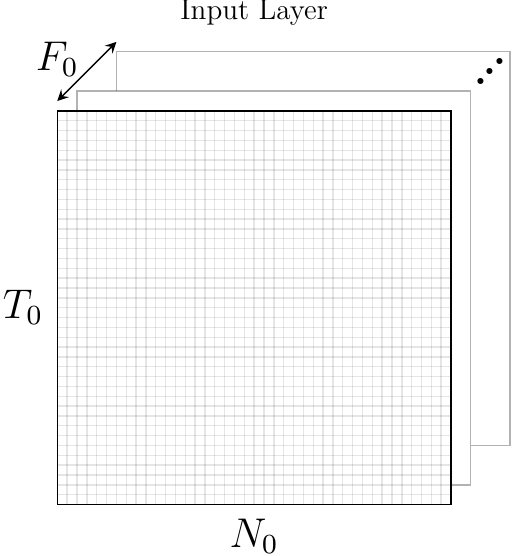}};
\end{tikzpicture}
\caption{\label{fig:input_layer} The Input layer}
\end{center}
\end{figure}

We will be considering a input of $F_0$ channels. In the standard image treatment, these channels can correspond to the RGB colors ($F_0=3$). Each image in each channel will be of size $N_0\times T_0$ (width$\times$height). The input will be denoted $X^{(t)}_{f\,j\,k}$, with $t\in \llbracket 0, T_{{\rm mb}}-1\rrbracket$ (size of the Mini-batch set, see chapter \ref{sec:chapterFNN}), $j \in \llbracket 0, N_0-1\rrbracket$ and $k \in \llbracket 0, T_0-1\rrbracket$. A standard input treatment is to center the data following either one of the two following procedure
\begin{align}
\tilde{X}^{(t)}_{f\,j\,k}&=X^{(t)}_{i\,j\,k}-\mu_{f}\;,&
\tilde{X}^{(t)}_{f\,j\,k}&=X^{(t)}_{i\,j\,k}-\mu_{f\,j\,k}\;
\end{align}
with
\begin{align}
\mu_{f}&=\frac{1}{T_{{\rm train}}T_0 N_0}\sum^{T_{{\rm train}}-1}_{t=0}\sum_j^{N_0-1}
\sum_k^{T_0-1}X^{(t)}_{f\,j\,k}\;,\\
\mu_{f\,j\,k}&=\frac{1}{T_{{\rm train}}}\sum^{T_{{\rm train}}-1}_{t=0}X^{(t)}_{f\,j\,k}\;.
\end{align}
This correspond to either compute the mean per pixel over the training set, or to also average over every pixel. This procedure should not be followed for regression tasks. To conclude, figure \ref{fig:input_layer} shows what the input layer looks like.

\subsection{Padding}

As we will see when we proceed, it may be convenient to "pad" the feature maps in order to preserve the width and the height of the images though several hidden layers. The padding operation amounts to add $0$'s around the original image. With a padding of size $P$, we add $P$ zeros at the beginning of each row and column of a given feature map. This is illustrated in the following figure

\begin{figure}[H]
\begin{center}
\begin{tikzpicture}
\node[] at (0,0) {\includegraphics[scale=1]{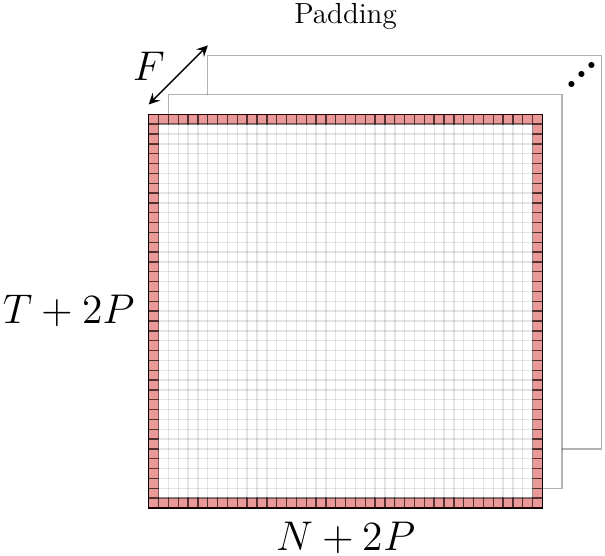}};
\end{tikzpicture}
\caption{Padding of the feature maps. The zeros added correspond to the red tiles, hence a padding of size $P=1$.}
\end{center}
\end{figure}

\subsection{Convolution}

The convolution operation that gives its name to the CNN is the fundamental building block of this type of network. It amounts to convolute a feature map of an input hidden layer with a weight matrix to give rise to an output feature map. The weight is really a four dimensional tensor, one dimension ($F$) being the number of feature maps of the convolutional input layer, another ($F_p$) the number of feature maps of the convolutional output layer. The two others gives the size of the receptive field in the width and the height direction. The receptive field allows one to convolute a subset instead of the whole input image. It aims at searching similar patterns in the input image, no matter where the pattern is (translational invariance). The width and the height of the output image are also determined by the stride: it is simply the number of pixel by which one slides in the vertical and/or the horizontal direction before applying again the convolution operation. A good picture being worth a thousand words, here is the convolution operation in a nutshell

\begin{figure}[H]
\begin{center}
\begin{tikzpicture}
\node[] at (0,0) {\includegraphics[scale=1]{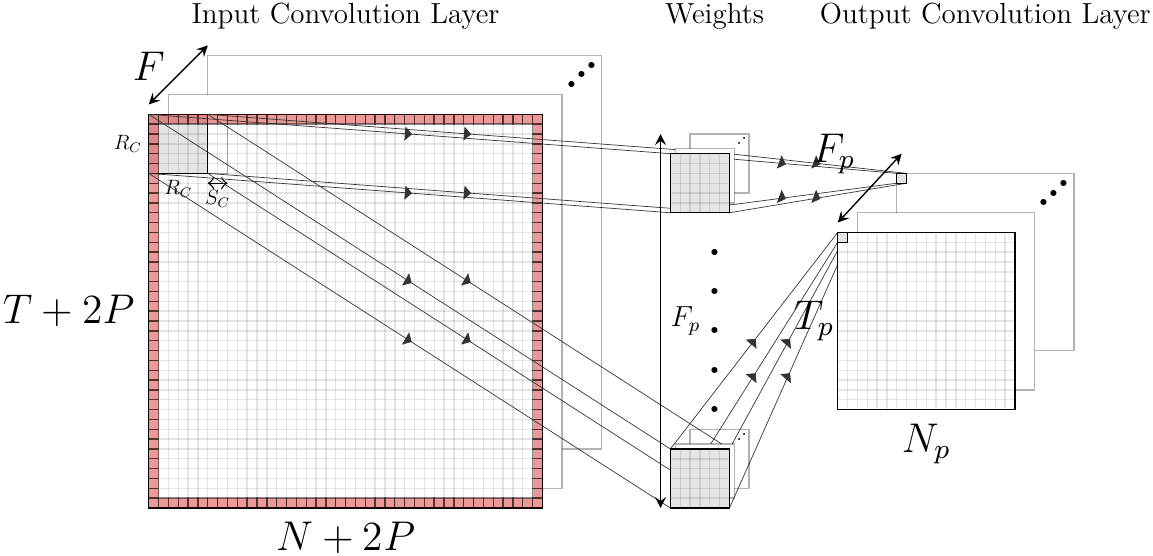}};
\end{tikzpicture}
\caption{The convolution operation}
\end{center}
\end{figure}

Here $R_C$ is the size of the convolutional receptive field (we will see that the pooling operation also has a receptive field and a stride) and $S_C$ the convolutional stride. The widths and heights of the output image can be computed thanks to the input height $T$ and output width $N$
\begin{align}
N_p&=\frac{N+2P-R_C}{S_C}+1 \;,&
T_p&=\frac{T+2P-R_C}{S_C}+1\;.
\end{align}
It is common to introduce a padding to preserve the widths and heights of the input image $N=N_p=T=T_p$, so that in these cases $S_C=1$ and
\begin{align}
P&=\frac{R_C-1}{2}\;.
\end{align}
For a given layer $n$, the convolution operation mathematically reads (similar in spirit to the weight averaging procedure of a FNN)
\begin{align}
a_{f\,l\,m}^{(t)(\nu)}&=\sum^{F_\nu-1}_{f'=0}\sum^{R_C-1}_{j=0}\sum^{R_C-1}_{k=0}
\Theta^{(o)f}_{f'\,j\,k}h^{(t)(\nu)}_{f'\,S_Cl+j\,S_Cm+k}\;,
\end{align}
where $o$ characterizes the $o+1$ convolution in the network. Here $\nu$ denotes the $\nu$'th hidden layer of the network (and thus belongs to $\llbracket0,N-1 \rrbracket$), and $f\in\llbracket0,F_{\nu+1}-1\rrbracket$, $l\in\llbracket0,N_{\nu+1}-1 \rrbracket$ and $m\in\llbracket0,T_{\nu+1}-1 \rrbracket$. Thus $S_Cl+j\in\llbracket0,N_\nu-1 \rrbracket$ and $S_Cl+j\in\llbracket0,T_\nu-1 \rrbracket$. One then obtains the hidden units via a ReLU (or other, see chapter \ref{sec:chapterFNN}) activation function application. Taking padding into account, it reads
\begin{align}
h_{f\,l+P\,m+P}^{(t)(\nu+1)}&=g\left(a_{f\,l\,m}^{(t)(\nu)}\right)\;.
\end{align}

\subsection{Pooling}

The pooling operation, less and less used in the current state of the art CNN, is fundamentally a dimension reduction operation. It amounts either to average or to take the maximum of a sub-image -- characterized by a pooling receptive field $R_P$ and a stride $S_P$ -- of the input feature map $F$ to obtain an output feature map $F_p=F$ of width $N_p<N$ and height $T_p<T$. To be noted: the padded values of the input hidden layers are not taken into account during the pooling operation (hence the $+P$ indices in the following formulas)

\begin{figure}[H]
\begin{center}
\begin{tikzpicture}
\node[] at (0,0) {\includegraphics[scale=1]{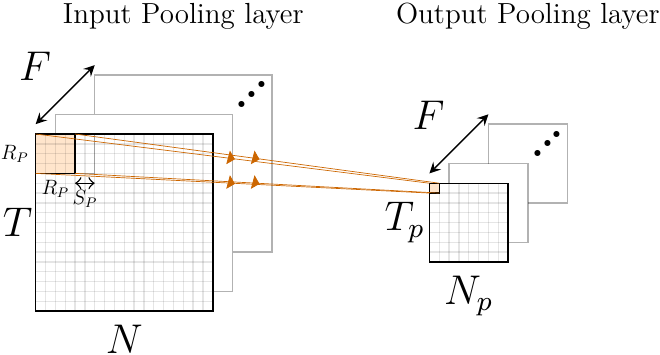}};
\end{tikzpicture}
\caption{The pooling operation}
\end{center}
\end{figure}

The average pooling procedure reads for a given $\nu$'th pooling operation
\begin{align}
a_{f\,l\,m}^{(t)(\nu)}&=\sum^{R_P-1}_{j,k=0} h_{f\,S_P l+j+P\,S_Pm+k+P}^{(t)(\nu)}\;,
\end{align}
while the max pooling reads
\begin{align}
a_{f\,l\,m}^{(t)(\nu)}&=\max^{R_P-1}_{j,k=0} h_{f\,S_P l+j+P\,S_Pm+k+P}^{(t)(\nu)}\;.
\end{align}
Here $\nu$ denotes the $\nu$'th hidden layer of the network (and thus belongs to $\llbracket0,N-1 \rrbracket$), and $f\in\llbracket0,F_{\nu+1}-1\rrbracket$, $l\in\llbracket0,N_{\nu+1}-1 \rrbracket$ and $m\in\llbracket0,T_{\nu+1}-1 \rrbracket$. Thus $S_Pl+j\in\llbracket0,N_\nu-1 \rrbracket$ and $S_Pl+j\in\llbracket0,T_\nu-1 \rrbracket$. Max pooling is extensively used in the literature, and we will therefore adopt it in all the following. Denoting $j^{(t)(p)}_{_{flm}},\,k^{(t)(p)}_{_{flm}}$ the indices at which the $l,m$ maximum of the $f$ feature map of the $t$'th batch sample is reached, we have
\begin{align}
h_{f\,l+P\,m+P}^{(t)(\nu+1)}&=a_{f\,l\,m}^{(t)(\nu)}=
h^{(t)(\nu)}_{f\,S_P l+j^{(t)(p)}_{_{flm}}+P\,S_Pm+k^{(t)(p)}_{_{flm}}+P}\;.
\end{align}

\subsection{Towards fully connected layers}

At some point in a CNN the convolutional receptive field is equal to the width and the height of the input image. In this case, the convolution operation becomes a kind of weight averaging procedure (as in a FNN)

\begin{figure}[H]
\begin{center}
\begin{tikzpicture}
\node[] at (0,0) {\includegraphics[scale=1]{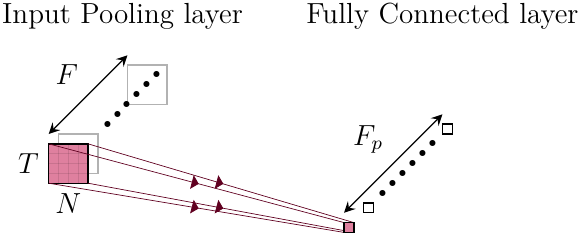}};
\end{tikzpicture}
\caption{Fully connected operation to get images of width and height $1$.}
\end{center}
\end{figure}

This weight averaging procedure reads
\begin{align}
a_{f}^{(t)(\nu)}&=\sum^{F_\nu-1}_{f'=0}\sum^{N-1}_{l=0}
\sum^{T-1}_{m=0}\Theta^{(o)f}_{f'lm}h^{(t)(\nu)}_{f'l+Pm+P}\;,
\end{align}
and is followed by the activation function
\begin{align}
h_{f}^{(t)(\nu+1)}&=g\left(a_{f}^{(t)(\nu)}\right)\;,
\end{align}
\subsection{fully connected layers}

After the previous operation, the remaining network is just a FNN one. The weigh averaging procedure reads

\begin{align}
a_{f}^{(t)(\nu)}&=\sum^{F_\nu-1}_{f'=0}\Theta^{(o)f}_{f'}h^{(t)(\nu)}_{f'}\;,
\end{align}
and is followed as usual by the activation function
 \begin{align}
h_{f}^{(t)(\nu+1)}&=g\left(a_{f}^{(t)(\nu)}\right)\;,
\end{align}

\begin{figure}[H]
\begin{center}
\begin{tikzpicture}
\node[] at (0,0) {\includegraphics[scale=1]{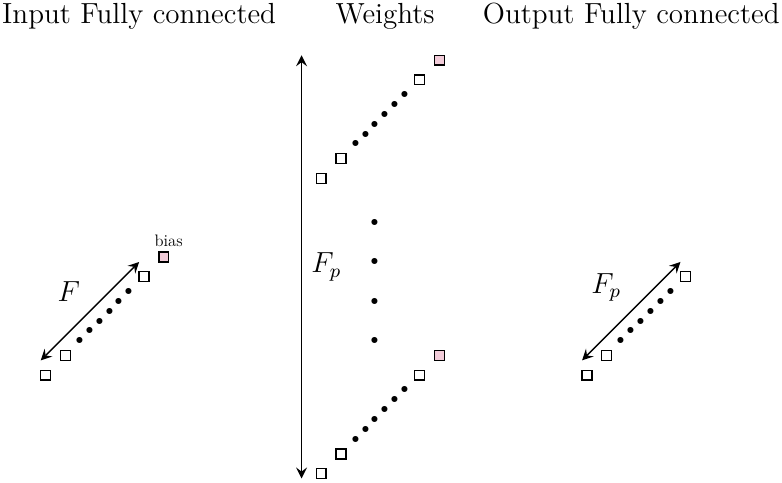}};
\end{tikzpicture}
\caption{Fully connected operation, identical to the FNN operations.}
\end{center}
\end{figure}

\subsection{Output connected layer}

Finally, the output is computed as in a FNN
\begin{align}
a_{f}^{(t)(N-1)}&=\sum^{F_{N}-1}_{f'=0}\Theta^{(o)f}_{f'}h^{(t)(N-1)}_{f'}\;,
&
h_{f}^{(t)(N)}&=o\left(a_{f}^{(t)(N-1)}\right)\;,
\end{align}
where as in a FNN, $o$ is either the L2 or the cross-entropy loss function (see chapter \ref{sec:chapterFNN}).

\section{Modification to Batch Normalization}

In a CNN, Batch normalization is modified in the following way (here, contrary to a regular FNN, not all the hidden layers need to be Batch normalized. Indeed this operation is not performed on the output of the pooling layers. We will hence use different names $\nu$ and $n$ for the regular and batch normalized hidden layers)
\begin{align}
\tilde{h}_{f\,l\,m}^{(t)(n)}&=\frac{h_{f\,l\,m}^{(t)(\nu)}-\hat{h}_{f}^{(n)}}
{\sqrt{\left(\hat{\sigma}_{f}^{(n)}\right)^2+\epsilon}}\;,
\end{align}
with
\begin{align}
\hat{h}_{f}^{(n)}&=
\frac{1}{T_{{\rm mb}}N_nT_n}\sum^{T_{{\rm mb}}-1}_{t=0}\sum^{N_n-1}_{l=0}\sum^{T_n-1}_{m=0}h_{f\,l\,m}^{(t)(\nu)}\\
\left(\hat{\sigma}_{f}^{(n)}\right)^2&=\frac{1}{T_{{\rm mb}}N_nT_n}\sum^{T_{{\rm mb}}-1}_{t=0}
\sum^{N_n-1}_{l=0}\sum^{T_n-1}_{m=0}\left(h_{f\,l\,m}^{(t)(\nu)}-\hat{h}_{f}^{(n)}\right)^2\;.
\end{align}
The identity transform can be implemented thanks to the two additional parameters $(\gamma_f,\beta_f)$
\begin{align}
y^{(t)(n)}_{f\,l\,m}&=\gamma^{(n)}_f\,\tilde{h}_{f\,l\,m}^{(t)(n)}+\beta^{(n)}_f\;.
\end{align}
For the evaluation of the cross-validation and the test set (calling e the number of iterations/epochs), one has to compute
\begin{align}
\mathbb{E}\left[h_{f\,l\,m}^{(t)(\nu)}\right]_{e+1} &=
\frac{e\mathbb{E}\left[h_{f\,l\,m}^{(t)(\nu)}\right]_{e}+\hat{h}_{f}^{(n)}}{e+1}\;,\\
\mathbb{V}ar\left[h_{f\,l\,m}^{(t)(\nu)}\right]_{e+1} &=
\frac{i\mathbb{V}ar\left[h_{f\,l\,m}^{(t)(\nu)}\right]_{e}+\left(\hat{\sigma}_{f}^{(n)}\right)^2}{e+1}
\end{align}
and what will be used at test time is $\mathbb{E}\left[h_{f\,l\,m}^{(t)(\nu)}\right]$ and $\frac{T_{{\rm mb}}}{T_{{\rm mb}}-1}\mathbb{V}ar\left[h_{f\,l\,m}^{(t)(\nu)}\right]$.

\section{Network architectures}

We will now review the standard CNN architectures that have been used in the literature in the past 20 years, from old to very recent (end of 2015) ones. To allow for an easy graphical representation, we will adopt the following schematic representation of the different layers.

\begin{figure}[H]
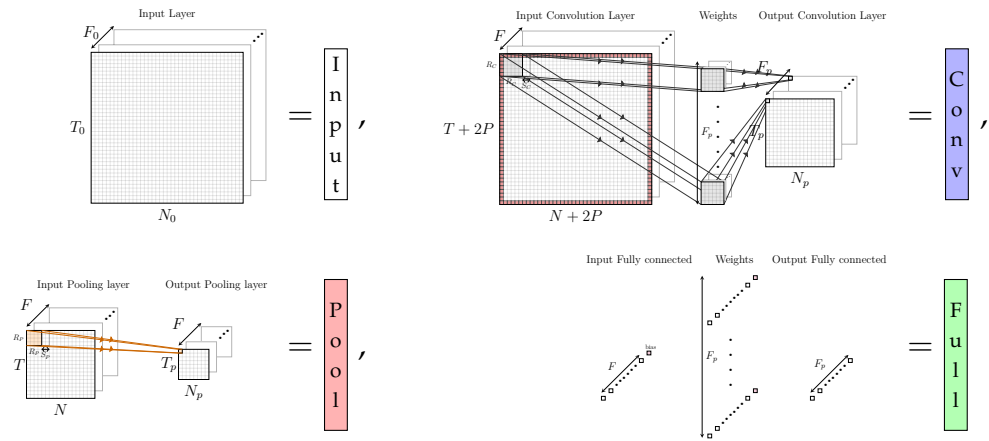

\begin{align*}
\begin{tikzpicture}[baseline=-3pt]
\node[] at (0,0) {\includegraphics[scale=0.5]{input_layer}};
\end{tikzpicture}&=
\begin{tikzpicture}[baseline=-0pt]
\draw (0,-1) rectangle (0+0.3,1);
\node[align=center,scale = 0.65] at (0+0.15,0) {I \\ n \\ p \\ u \\ t};
\end{tikzpicture}\;,&
\begin{tikzpicture}[baseline=-3pt]
\node[] at (0,0) {\includegraphics[scale=0.5]{VGG-conv}};
\end{tikzpicture}&=
\begin{tikzpicture}[baseline=-0pt]
\filldraw[fill=blue!30!white] (0,-1) rectangle (0+0.3,1);
\node[align=center,scale = 0.65] at (0+0.15,0) {C \\ o \\ n \\ v};
\end{tikzpicture}\;,\notag\\
\begin{tikzpicture}[baseline=-3pt]
\node[] at (0,0) {\includegraphics[scale=0.5]{VGG-pool}};
\end{tikzpicture}&=
\begin{tikzpicture}[baseline=-0pt]
\filldraw[fill=red!30!white] (0,-1) rectangle (0+0.3,1);
\node[align=center,scale = 0.65] at (0+0.15,0) {P \\ o \\ o \\ l};
\end{tikzpicture}\;,&
\begin{tikzpicture}[baseline=-3pt]
\node[] at (0,0) {\includegraphics[scale=0.5]{VGG-fc}};
\end{tikzpicture}&=
\begin{tikzpicture}[baseline=-0pt]
\filldraw[fill=green!30!white] (0,-1) rectangle (0+0.3,1);
\node[align=center,scale = 0.65] at (0+0.15,0) {F \\ u \\ l \\ l};
\end{tikzpicture}
\end{align*}
\begin{center}
\caption{Schematic representation of the different layer}
\end{center}
\end{figure}

\subsection{Realistic architectures}

In realistic architectures, every fully connected layer (except the last one related to the output) is followed by a ReLU (or other) activation and then a batch normalization step (these two data processing steps can be inverted, as it was the case in the original BN implementation).
\begin{figure}[H]
\begin{center}
\begin{tikzpicture}
\node[] at (0,0) {\includegraphics[scale=1.5]{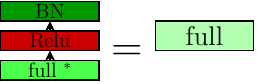}};
\end{tikzpicture}
\caption{Realistic Fully connected operation}
\end{center}
\end{figure}
 The same holds for convolutional layers
\begin{figure}[H]
\begin{center}
\begin{tikzpicture}
\node[] at (0,0) {\includegraphics[scale=1.5]{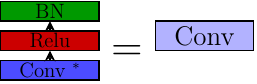}};
\end{tikzpicture}
\caption{Realistic Convolution operation}
\end{center}
\end{figure}

We will adopt the simplified right hand side representation, keeping in mind that the true structure of a CNN is richer. With this in mind -- and mentioning in passing \cite{Gu2015RecentAI} that details recent CNN advances, let us now turn to the first popular CNN used by the deep learning community.

\subsection{LeNet}

The LeNet\cite{Lecun98gradient-basedlearning} (end of the 90's) network is formed by an input, followed by two conv-pool layers and then a fully-connected layer before the final output. It can be seen in figure \ref{fig:lenet-CNN}
\begin{figure}[H]
\begin{center}
\begin{tikzpicture}
\node[] at (0,0) {\includegraphics[scale=1]{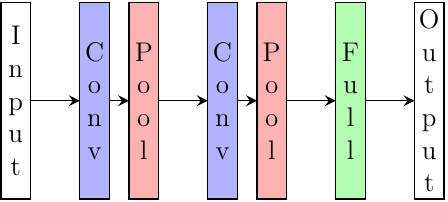}};
\end{tikzpicture}
\caption{The LeNet CNN}
\end{center}
\end{figure}

When treating large images ($224\times 224$), this implies to use large size of receptive fields and strides. This has two downsides. Firstly, the number or parameter in a given weight matrix is proportional to the size of the receptive field, hence the larger it is the larger the number of parameter. The network can thus be more prone to overfit. Second, a large stride and receptive field means a less subtle analysis of the fine structures of the images. All the subsequent CNN implementations aim at reducing one of these two issues.

\subsection{AlexNet}

The AlexNet\cite{NIPS2012_4824} (2012) saw no qualitative leap in the CNN theory, but due to better processors was able to deal with more hidden layers.

\begin{figure}[H]
\begin{center}
\begin{tikzpicture}
\node[] at (0,0) {\includegraphics[scale=1]{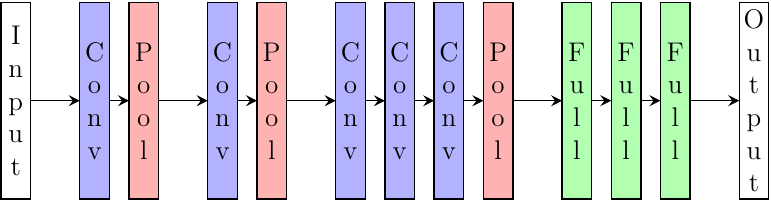}};
\end{tikzpicture}
\caption{The AlexNet CNN}
\end{center}
\end{figure}

This network is still commonly used, though less since the arrival of the VGG network.

\subsection{VGG}

The VGG\cite{DBLP:journals/corr/SimonyanZ14a} network (2014) adopted a simple criteria: only $2 \times 2$ paddings of stride $2$ and $3\times 3$ convolutions of stride $1$ with a padding of size $1$, hence preserving the size of the image's width and height through the convolution operations.

\begin{figure}[H]
\begin{center}
\begin{tikzpicture}
\node[] at (0,0) {\includegraphics[scale=1]{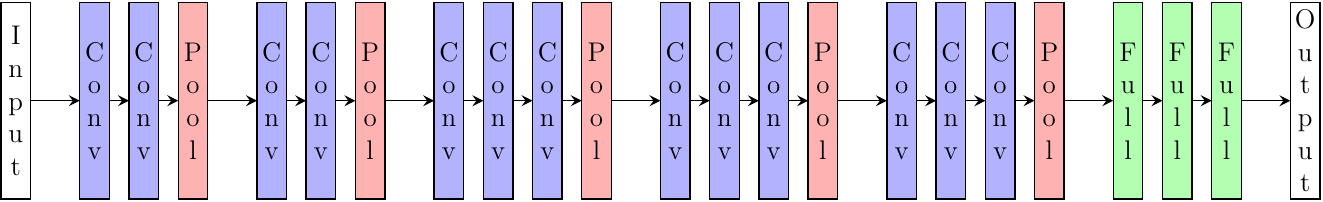}};
\end{tikzpicture}
\caption{The VGG CNN}
\end{center}
\end{figure}

This network is the standard one in most of the deep learning packages dealing with CNN. It is no longer the state of the art though, as a design innovation has taken place since its creation.

\subsection{GoogleNet}

The GoogleNet\cite{43022} introduced a new type of "layer" (which is in reality a combination of existing layers): the inception layer (in reference to the movie by Christopher Nolan). Instead of passing from one layer of a CNN to the next by a simple pool, conv or fully-connected (fc) operation, one averages the result of the following architecture.

\begin{figure}[H]
\begin{center}
\begin{tikzpicture}
\node[] at (0,0) {\includegraphics[scale=1]{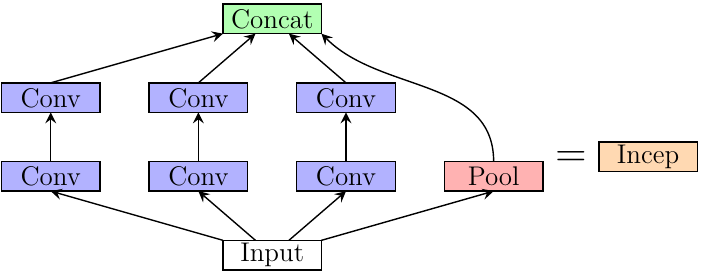}};
\end{tikzpicture}
\caption{The Inception module}
\end{center}
\end{figure}

We won't enter into the details of the concat layer, as the Google Net illustrated on the following figure is (already!) no longer state of the art.

\begin{figure}[H]
\begin{center}
\begin{tikzpicture}
\node[] at (0,0) {\includegraphics[scale=1]{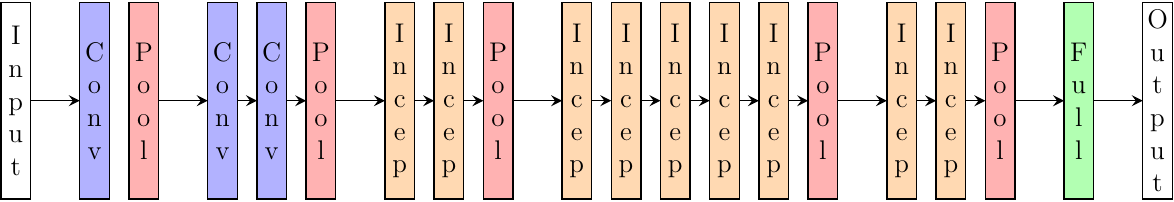}};
\end{tikzpicture}
\caption{The GoogleNet CNN}
\end{center}
\end{figure}

Indeed, the idea of averaging the result of several conv-pool operations to obtain the next hidden layer of a CNN as been exploited but greatly simplified by the state of the art CNN : The ResNet.

\subsection{ResNet}

\begin{figure}[H]
\begin{center}
\begin{align*}
&\begin{tikzpicture}
\node[] at (0,0) {\includegraphics[scale=1.3]{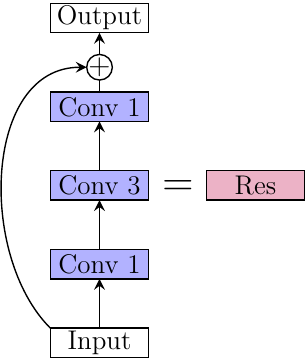}};
\end{tikzpicture}&
\begin{tikzpicture}
\node[] at (0,0) {\includegraphics[scale=1]{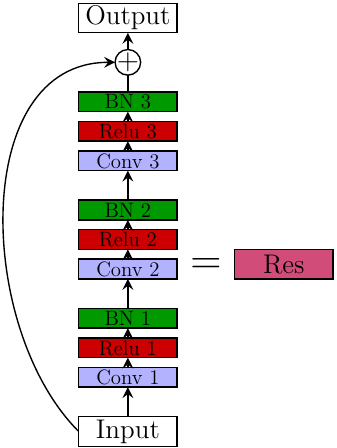}};
\end{tikzpicture}
\end{align*}
\caption{\label{fig:Bottleneck_BN}The Bottleneck Residual architecture. Schematic representation on the left, realistic one on the right. It amounts to a $1\times 1$ conv of stride $1$ and padding $0$, then a standard VGG conv and again a $1 \times 1$ conv. Two main modifications in our presentation of ResNet: BN operations have been put after ReLU ones, and the final ReLU is before the plus operation.}
\end{center}
\end{figure}

The ResNet\cite{He2015} takes back the simple idea of the VGG net to always use the same size for the convolution operations (except for the first one). It also takes into account an experimental fact: the fully connected layer (that usually contains most of the parameters given their size) are not really necessary to perform well. Removing them leads to a great decrease of the number of parameters of a CNN. In addition, the pooling operation is also less and less popular and tend to be replaced by convolution operations. This gives the basic ingredients of the ResNet fundamental building block, the Residual module of figure \ref{fig:Bottleneck_BN}.

\vspace{0.2cm}

Two important points have to be mentioned concerning the Residual module. Firstly, a usual conv-conv-conv structure would lead to the following output (forgetting about batch normalization for simplicity and only for the time being, and denoting that there is no need for padding in $1\times 1$ convolution operations)
\begin{align}
h^{(t)(1)}_{fl+Pm+P}&=
g\left(\sum^{F_0-1}_{f'=0}\sum^{R_C-1}_{j=0}\sum^{R_C-1}_{k=0}
\Theta^{(0)f}_{f'\,j\,k}h^{(t)(0)}_{f'\,S_Cl+j\,S_Cm+k}\right)\notag\\
h^{(t)(2)}_{flm}&=
g\left(\sum^{F_1-1}_{f'=0}\sum^{R_C-1}_{j=0}\sum^{R_C-1}_{k=0}
\Theta^{(0)f}_{f'\,j\,k}h^{(t)(1)}_{f'\,S_Cl+j\,S_Cm+k}\right)\notag\\
h^{(t)(3)}_{flm}&=
g\left(\sum^{F_2-1}_{f'=0}\sum^{R_C-1}_{j=0}\sum^{R_C-1}_{k=0}
\Theta^{(0)f}_{f'\,j\,k}h^{(t)(2)}_{f'\,S_Cl+j\,S_Cm+k}\right)\;,
\end{align}
whereas the Residual module modifies the last previous equation to (implying that the width, the size and the number of feature size of the input and the output being the same)
\begin{align}
h^{(t)(4)}_{flm}&=h^{(t)(0)}_{flm}+g\left(
\sum^{F_2-1}_{f'=0}\sum^{R_C-1}_{j=0}\sum^{R_C-1}_{k=0}
\Theta^{(0)f}_{f'\,j\,k}h^{(t)(2)}_{f'\,S_Cl+j\,S_Cm+k}\right)\notag\\
&=h^{(t)(0)}_{flm}+\delta h^{(t)(0)}_{flm}\;.
\end{align}
Instead of trying to fit the input, one is trying to fit a tiny modification of the input, hence the name residual. This allows the network to minimally modify the input when necessary, contrary to traditional architectures. Secondly, if the number of feature maps is important, a $3\times 3$ convolution with stride 1 could be very costly in term of execution time and prone to overfit (large number of parameters). This is the reason of the presence of the $1 \times 1$ convolution, whose aim is just to prepare the input to the $3\times 3$ conv to reduce the number of feature maps, number which is then restored with the final $1\times 1$ conv of the Residual module. The first $1\times 1$ convolution thus reads as a weight averaging operation
\begin{align}
h^{(t)(1)}_{fl+Pm+P}&=
g\left(\sum^{F_0-1}_{f'=0}
\Theta^{(0)f}_{f'}h^{(t)(0)}_{f'\,l\,m}\right)\;,
\end{align}
but is designed such that $f\in F_1\ll F_0$. The second $1\times 1$ convolution reads
\begin{align}
h^{(t)(3)}_{flm}&=g\left(\sum^{F_1-1}_{i=0}\Theta^{(2)f}_{i}h^{(t)(1)}_{i\,l\,m}\right)\;,
\end{align}
with $f \in F_0$, restoring the initial feature map size. The ResNet architecture is then the stacking of a large number (usually 50) of Residual modules, preceded by a conv-pool layer and ended by a pooling operation to obtain a fully connected layer, to which the output function is directly applied. This is illustrated in the following figure.

\begin{figure}[H]
\begin{center}
\begin{tikzpicture}
\node[] at (0,0) {\includegraphics[scale=1]{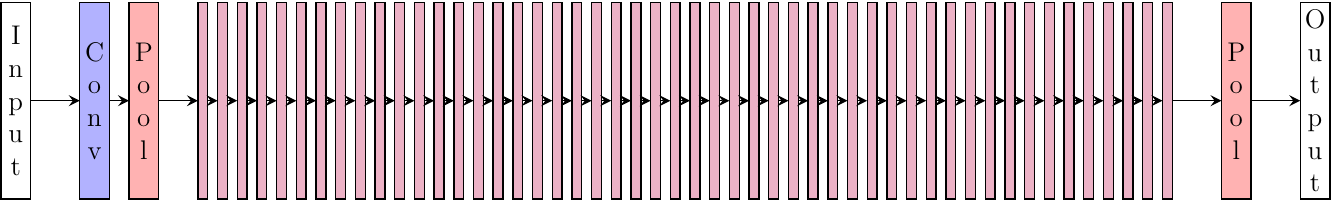}};
\end{tikzpicture}
\caption{The ResNet CNN}
\end{center}
\end{figure}

The ResNet CNN has accomplished state of the art results on a number of popular training sets (CIFAR, MNIST...). In practice, we will present in the following the backpropagation algorithm for CNN having standard (like VGG) architectures in mind.

\section{Backpropagation}

In a FNN, one just has to compute two kind of backpropagations : from output to fully connected (fc) layer and from fc to fc. In a traditional CNN, 4 new kind of propagations have to be computed: fc to pool, pool to conv, conv to conv and conv to pool. We present the corresponding error rates in the next sections, postponing their derivation to the appendix. We will consider as in a FNN a network with an input layer labelled $0$, N-1 hidden layers labelled $i$ and an output layer labelled $N$ ($N+1$ layers in total in the network).

\subsection{Backpropagate through Batch Normalization} \label{sec:BackpropbatchnormCNN}

As in FNN, backpropagation introduces a new gradient
\begin{align}
\delta^f_{f'}J^{(tt')(n)}_{fll'mm'}&=
\frac{\partial y^{(t')(n)}_{f'\,l'\,m'}}{\partial h_{f\,l\,m}^{(t)(\nu)}}\;.
\end{align}
we show in appendix \ref{sec:appenbatchnorm-vgg} that for pool and conv layers
\begin{align}
J^{(tt')(n)}_{fll'mm'}&=\tilde{\gamma}^{(n)}_f \left[\delta^{t'}_t\delta^{l'}_l\delta^{m'}_m-
\frac{1+\tilde{h}_{f\,l'\,m'}^{(t')(n)}\tilde{h}_{f\,l\,m}^{(t)(n)}}{T_{{\rm mb}}N_nT_n}\right]\;,
\end{align}
while we find the FNN result as expected for fc layers
\begin{align}
J^{(tt')(n)}_{f}&=\tilde{\gamma}^{(n)}_f \left[\delta^{t'}_t-
\frac{1+\tilde{h}_{f}^{(t')(n)}\tilde{h}_{f}^{(t)(n)}}{T_{{\rm mb}}}\right]\;.
\end{align}

\subsection{Error updates}

We will call the specific CNN error rates (depending on whether we need padding or not)
\begin{align}
\delta^{(t)(\nu)}_{fl(+P)m(+P)}&=\frac{\partial }{\partial a_{f\,l\,m}^{(t)(i)}}J(\Theta)\;,
\end{align}

\subsubsection{Backpropagate from output to fc}

Backpropagate from output to fc is schematically illustrated on the following plot

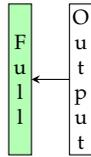
\begin{figure}[H]
\begin{center}
\begin{tikzpicture}[baseline=-0pt]
\draw (0.8,-1) rectangle (0.8+0.3,1);
\node[align=center,scale = 0.65] at (0.8+0.15,0) {O \\ u \\ t \\ p \\ u \\ t};
\draw[-stealth] (0.8,0) -- (0+0.3,0);
\filldraw[fill=green!30!white] (0,-1) rectangle (0+0.3,1);
\node[align=center,scale = 0.65] at (0+0.15,0) {F \\ u \\ l \\ l};
\end{tikzpicture}
\caption{Backpropagate from output to fc.}
\end{center}
\end{figure}
As in FNN, we find for the L2 loss function
\begin{align}
\delta^{(t)(N-1)}_f&= \frac{1}{T_{{\rm mb}}}\left(h_{f}^{(t)(N)}-y_f^{(t)}\right)\;,
\end{align}
and for the cross-entropy one
\begin{align}
\delta^{(t)(N-1)}_f&= \frac{1}{T_{{\rm mb}}}\left(h_{f}^{(t)(N)}-\delta_{y^{(t)}}^f\right)\;,
\end{align}

\subsubsection{Backpropagate from fc to fc}

Backpropagate from fc to fc is schematically illustrated on the following plot

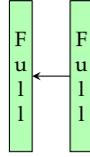
\begin{figure}[H]
\begin{center}
\begin{tikzpicture}[baseline=-0pt]
\filldraw[fill=green!30!white] (0.8,-1) rectangle (0.8+0.3,1);
\node[align=center,scale = 0.65] at (0.8+0.15,0) {F \\ u \\ l \\ l};
\draw[-stealth] (0.8,0) -- (0+0.3,0);
\filldraw[fill=green!30!white] (0,-1) rectangle (0+0.3,1);
\node[align=center,scale = 0.65] at (0+0.15,0) {F \\ u \\ l \\ l};
\end{tikzpicture}
\caption{Backpropagate from fc to fc.}
\end{center}
\end{figure}
As in FNN, we find
\begin{align}
\delta^{(t)(\nu)}_f &=g'\left(a_{f}^{(t)(\nu)}\right)
\sum_{t'=0}^{T_{{\rm mb}}-1}\sum_{f'=0}^{F_{\nu+1}-1}\Theta^{(o)f'}_{f}J^{(tt')(n)}_{f} \delta^{(t)(\nu+1)}_{f'}\;,
\end{align}

\subsubsection{Backpropagate from fc to pool}

Backpropagate from fc to pool is schematically illustrated on the following plot

\begin{figure}[H]
\begin{center}
\begin{tikzpicture}[baseline=-0pt]
\filldraw[fill=green!30!white] (0.8,-1) rectangle (0.8+0.3,1);
\node[align=center,scale = 0.65] at (0.8+0.15,0) {F \\ u \\ l \\ l};
\draw[-stealth] (0.8,0) -- (0+0.3,0);
\filldraw[fill=red!30!white] (0,-1) rectangle (0+0.3,1);
\node[align=center,scale = 0.65] at (0+0.15,0) {P \\ o \\ o \\ l};
\end{tikzpicture}
\caption{Backpropagate from fc to pool.}
\end{center}
\end{figure}
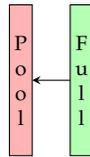
We show in appendix \ref{sec:appenderrorrate} that this induces the following error rate
\begin{align}
\delta^{(t)(\nu)}_{flm}&=\sum_{f'=0}^{F_{\nu+1}-1}\Theta^{(o)f'}_{f\,l\,m} \delta^{(t)(\nu+1)}_{f'}\;,
\end{align}

\subsubsection{Backpropagate from pool to conv}

Backpropagate from pool to conv is schematically illustrated on the following plot

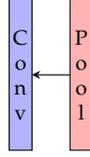
\begin{figure}[H]
\begin{center}
\begin{tikzpicture}[baseline=-0pt]
\filldraw[fill=red!30!white] (0.8,-1) rectangle (0.8+0.3,1);
\node[align=center,scale = 0.65] at (0.8+0.15,0) {P \\ o \\ o \\ l};
\draw[-stealth] (0.8,0) -- (0+0.3,0);
\filldraw[fill=blue!30!white] (0,-1) rectangle (0+0.3,1);
\node[align=center,scale = 0.65] at (0+0.15,0) {C \\ o \\ n \\ v};
\end{tikzpicture}
\caption{Backpropagate from pool to conv.}
\end{center}
\end{figure}
We show in appendix \ref{sec:appenbatchnorm-vgg} that this induces the following error rate (calling the pooling layer the $p$th one)
\begin{align}
\delta^{(t)(\nu)}_{fl+Pm+P}&=g'\left(a_{f\,l\,m}^{(t)(\nu)}\right)
\sum_{t'=0}^{T_{{\rm mb}}-1}\sum^{N_{\nu+1}-1}_{l'=0}\sum^{T_{\nu+1}-1}_{m'=0}\delta_{fl'm'}^{(t')(\nu+1)}\notag\\
&\times J^{(tt')(n)}_{f\,S_P l'+j_{fl'm'}^{(t')(p)}+P\,S_Pm'+k_{fl'm'}^{(t')(p)}+P\,l+P\,m+P}\;.
\end{align}
Note that we have padded this error rate.

\subsubsection{Backpropagate from conv to conv}

Backpropagate from conv to conv is schematically illustrated on the following plot

\begin{figure}[H]
\begin{center}
\begin{tikzpicture}[baseline=-0pt]
\filldraw[fill=blue!30!white] (0.8,-1) rectangle (0.8+0.3,1);
\node[align=center,scale = 0.65] at (0.8+0.15,0) {C \\ o \\ n \\ v};
\draw[-stealth] (0.8,0) -- (0+0.3,0);
\filldraw[fill=blue!30!white] (0,-1) rectangle (0+0.3,1);
\node[align=center,scale = 0.65] at (0+0.15,0) {C \\ o \\ n \\ v};
\end{tikzpicture}
\caption{Backpropagate from conv to conv.}
\end{center}
\end{figure}
We show in appendix \ref{sec:appenderrorrate} that this induces the following error rate
\begin{align}
\delta^{(t)(\nu)}_{fl+Pm+P}&=g'\left(a_{f\,l\,m}^{(t)(\nu)}\right)
\sum_{t'=0}^{T_{{\rm mb}}-1}\sum^{F_{\nu+1}-1}_{f'=0}\sum^{N_{\nu+1}-1}_{l'=0}\sum^{T_{\nu+1}-1}_{m'=0}
\sum^{R_C-1}_{j=0}\sum^{R_C-1}_{k=0}\delta_{f'l'+Pm'+P}^{(t')(\nu+1)}\notag\\
&\times\Theta^{(o)f'}_{f\,j\,k}J^{(tt')(n)}_{f\, S_Cl'+j\,S_Cm'+k\,l+P\,m+P}
\end{align}
Note that we have padded this error rate.

\subsubsection{Backpropagate from conv to pool}

Backpropagate from conv to pool is schematically illustrated on the following plot

\begin{figure}[H]
\begin{center}
\begin{tikzpicture}[baseline=-0pt]
\filldraw[fill=blue!30!white] (0.8,-1) rectangle (0.8+0.3,1);
\node[align=center,scale = 0.65] at (0.8+0.15,0) {C \\ o \\ n \\ v};
\draw[-stealth] (0.8,0) -- (0+0.3,0);
\filldraw[fill=red!30!white] (0,-1) rectangle (0+0.3,1);
\node[align=center,scale = 0.65] at (0+0.15,0) {P \\ o \\ o \\ l};
\end{tikzpicture}
\caption{Backpropagate from conv to pool.}
\end{center}
\end{figure}
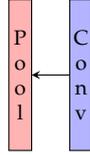
We show in appendix \ref{sec:appenderrorrate} that this induces the following error rate
\begin{align}
\delta^{(t)(\nu)}_{flm}&=
\sum^{F_{\nu+1}-1}_{f'=0}\sum^{R_C-1}_{j=0}\sum^{R_C-1}_{k=0}
\Theta^{(o)f}_{f'\,j\,k}\delta_{f\,\frac{l+P-j}{S_C}+P\,\frac{m+P-k}{S_C}+P}^{(t)(\nu+1)}\;.
\end{align}

\subsection{Weight update}

For the weight updates, we will also consider separately the weights between fc to fc layer, fc to pool, conv to conv, conv to pool and conv to input.

\subsubsection{Weight update from fc to fc}

For the two layer interactions

\begin{figure}[H]
\begin{center}
\begin{tikzpicture}[baseline=-0pt]
\filldraw[fill=green!30!white] (0.8,-1) rectangle (0.8+0.3,1);
\node[align=center,scale = 0.65] at (0.8+0.15,0) {F \\ u \\ l \\ l};
\draw[-stealth] (0.8,0) -- (0+0.3,0);
\filldraw[fill=green!30!white] (0,-1) rectangle (0+0.3,1);
\node[align=center,scale = 0.65] at (0+0.15,0) {F \\ u \\ l \\ l};
\end{tikzpicture}
\caption{Weight update between two fc layers.}
\end{center}
\end{figure}
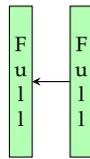

We have the weight update that reads
\begin{align}
\Delta^{\Theta(o)f}_{f'}&=\sum_{t=0}^{T_{{\rm mb}}-1} y^{(t)(n)}_{f'}\delta^{(t)(\nu)}_f
\end{align}

\subsubsection{Weight update from fc to pool}

For the two layer interactions

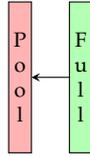
\begin{figure}[H]
\begin{center}
\begin{tikzpicture}[baseline=-0pt]
\filldraw[fill=green!30!white] (0.8,-1) rectangle (0.8+0.3,1);
\node[align=center,scale = 0.65] at (0.8+0.15,0) {F \\ u \\ l \\ l};
\draw[-stealth] (0.8,0) -- (0+0.3,0);
\filldraw[fill=red!30!white] (0,-1) rectangle (0+0.3,1);
\node[align=center,scale = 0.65] at (0+0.15,0) {P \\ o \\ o \\ l};
\end{tikzpicture}
\caption{Weight update between a fc layer and a pool layer.}
\end{center}
\end{figure}

We have the weight update that reads
\begin{align}
\Delta^{\Theta(o)f}_{f'jk}&=\sum_{t=0}^{T_{{\rm mb}}-1} h^{(t)(\nu)}_{f'j+Pk+P}\delta^{(t)(\nu)}_f
\end{align}

\subsubsection{Weight update from conv to conv}

For the two layer interactions

\begin{figure}[H]
\begin{center}
\begin{tikzpicture}[baseline=-0pt]
\filldraw[fill=blue!30!white] (0.8,-1) rectangle (0.8+0.3,1);
\node[align=center,scale = 0.65] at (0.8+0.15,0) {C \\ o \\ n \\ v};
\draw[-stealth] (0.8,0) -- (0+0.3,0);
\filldraw[fill=blue!30!white] (0,-1) rectangle (0+0.3,1);
\node[align=center,scale = 0.65] at (0+0.15,0) {C \\ o \\ n \\ v};
\end{tikzpicture}
\caption{Weight update between two conv layers.}
\end{center}
\end{figure}
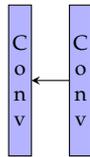

We have the weight update that reads
\begin{align}
\Delta^{\Theta(o)f}_{f'jk}&=\sum_{t=0}^{T_{{\rm mb}}-1}\sum^{T_{\nu+1}-1}_{l=0}\sum^{N_{\nu+1}-1}_{m=0}
y^{(t)(n)}_{f'\,l+j\,m+k}\delta_{f\,l+P\,m+P}^{(t)(\nu)}
\end{align}

\subsubsection{Weight update from conv to pool and conv to input}

For the two layer interactions

\begin{figure}[H]
\begin{center}
\begin{tikzpicture}[baseline=-0pt]
\filldraw[fill=blue!30!white] (0.8,-1) rectangle (0.8+0.3,1);
\node[align=center,scale = 0.65] at (0.8+0.15,0) {C \\ o \\ n \\ v};
\draw[-stealth] (0.8,0) -- (0+0.3,0);
\filldraw[fill=red!30!white] (0,-1) rectangle (0+0.3,1);
\node[align=center,scale = 0.65] at (0+0.15,0) {P \\ o \\ o \\ l};
\filldraw[fill=blue!30!white] (2.8,-1) rectangle (2.8+0.3,1);
\node[align=center,scale = 0.65] at (2.8+0.15,0) {C \\ o \\ n \\ v};
\draw[-stealth] (2.8,0) -- (2+0.3,0);
\draw (2,-1) rectangle (2+0.3,1);
\node[align=center,scale = 0.65] at (2+0.15,0) {I \\ n \\ p \\ u \\ t};
\end{tikzpicture}
\caption{Weight update between a conv and a pool layer, as well as between a conv and the input layer.}
\end{center}
\end{figure}
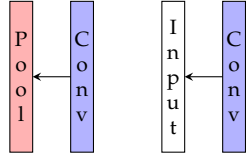
\begin{align}
\Delta^{\Theta(o)f}_{f'jk}&=\sum_{t=0}^{T_{{\rm mb}}-1}\sum^{T_{\nu+1}-1}_{l=0}\sum^{N_{\nu+1}-1}_{m=0}
h^{(t)(\nu)}_{f'\,l+j\,m+k}\delta_{f\,l+P\,m+P}^{(t)(\nu)}
\end{align}

\subsection{Coefficient update}

For the Coefficient updates, we will also consider separately the weights between fc to fc layer, fc to pool, cont to pool and conv to conv.

\subsubsection{Coefficient update from fc to fc}

For the two layer interactions

\begin{figure}[H]
\begin{center}
\begin{tikzpicture}[baseline=-0pt]
\filldraw[fill=green!30!white] (0.8,-1) rectangle (0.8+0.3,1);
\node[align=center,scale = 0.65] at (0.8+0.15,0) {F \\ u \\ l \\ l};
\draw[-stealth] (0.8,0) -- (0+0.3,0);
\filldraw[fill=green!30!white] (0,-1) rectangle (0+0.3,1);
\node[align=center,scale = 0.65] at (0+0.15,0) {F \\ u \\ l \\ l};
\end{tikzpicture}
\caption{Coefficient update between two fc layers.}
\end{center}
\end{figure}
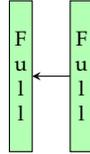

We have

\begin{align}
\Delta_f^{\gamma(n)}&=\sum_{t=0}^{T_{{\rm mb}}-1}\sum_{f'=0}^{F_{\nu+1}-1}
\Theta^{(o)f'}_{f}\tilde{h}^{(t)(n)}_{f}\delta^{(t)(\nu)}_{f'}\;,\notag\\
\Delta_f^{\beta(n)}&=
\sum_{t=0}^{T_{{\rm mb}}-1}\sum_{f'=0}^{F_{\nu+1}-1}\Theta^{(o)f'}_{f}\delta^{(t)(\nu)}_{f'}\;,
\end{align}

\subsubsection{Coefficient update from fc to pool and conv to pool}

For the two layer interactions

\begin{figure}[H]
\begin{center}
\begin{tikzpicture}[baseline=-0pt]
\filldraw[fill=green!30!white] (0.8,-1) rectangle (0.8+0.3,1);
\node[align=center,scale = 0.65] at (0.8+0.15,0) {F \\ u \\ l \\ l};
\draw[-stealth] (0.8,0) -- (0+0.3,0);
\filldraw[fill=red!30!white] (0,-1) rectangle (0+0.3,1);
\node[align=center,scale = 0.65] at (0+0.15,0) {P \\ o \\ o \\ l};
\filldraw[fill=blue!30!white] (2.8,-1) rectangle (2.8+0.3,1);
\node[align=center,scale = 0.65] at (2.8+0.15,0) {C \\ o \\ n \\ v};
\draw[-stealth] (2.8,0) -- (2+0.3,0);
\filldraw[fill=red!30!white] (2,-1) rectangle (2+0.3,1);
\node[align=center,scale = 0.65] at (2+0.15,0) {P \\ o \\ o \\ l};
\end{tikzpicture}
\caption{Coefficient update between a fc layer and a pool as well as a conv and a pool layer.}
\end{center}
\end{figure}
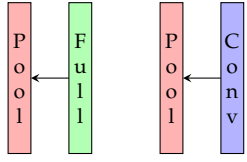

\begin{align}
\Delta_f^{\gamma(n)}&=
\sum_{t=0}^{T_{{\rm mb}}-1}\sum_{l=0}^{N_{\nu+1}-1}\sum_{m=0}^{T_{\nu+1}-1}
\tilde{h}_{f\,S_P l+j_{flm}^{(t)(p)}+P\,S_Pm+k_{flm}^{(t)(p)}+P}^{(t)(n)}\delta^{(t)(\nu)}_{flm}\;,\notag\\
\Delta_f^{\beta(n)}&=
\sum_{t=0}^{T_{{\rm mb}}-1}\sum_{l=0}^{N_{\nu+1}-1}\sum_{m=0}^{T_{\nu+1}-1}\delta^{(t)(\nu)}_{flm}\;,
\end{align}

\subsubsection{Coefficient update from conv to conv}

For the two layer interactions

\begin{figure}[H]
\begin{center}
\begin{tikzpicture}[baseline=-0pt]
\filldraw[fill=blue!30!white] (0.8,-1) rectangle (0.8+0.3,1);
\node[align=center,scale = 0.65] at (0.8+0.15,0) {C \\ o \\ n \\ v};
\draw[-stealth] (0.8,0) -- (0+0.3,0);
\filldraw[fill=blue!30!white] (0,-1) rectangle (0+0.3,1);
\node[align=center,scale = 0.65] at (0+0.15,0) {C \\ o \\ n \\ v};
\end{tikzpicture}
\caption{Coefficient update between two conv layers.}
\end{center}
\end{figure}

We have

\begin{align}
\Delta_f^{\gamma(n)}&=
\sum_{t=0}^{T_{{\rm mb}}-1}\sum_{f'=0}^{F_{\nu+1}-1}\sum_{l=0}^{N_{\nu+1}-1}\sum_{m=0}^{T_{\nu+1}-1}
\sum_{j=0}^{R_C-1}\sum_{k=0}^{R_C-1}\Theta^{(o)f'}_{fjk}
\tilde{h}^{(t)(n)}_{fl+j\,m+k}\delta^{(t)(\nu)}_{f'\,l+P\,m+P}\;,\notag\\
\Delta_f^{\beta(n)}&=
\sum_{t=0}^{T_{{\rm mb}}-1}\sum_{f'=0}^{F_{\nu+1}-1}\sum_{l=0}^{N_{\nu+1}-1}\sum_{m=0}^{T_{\nu+1}-1}
\sum_{j=0}^{R_C-1}\sum_{k=0}^{R_C-1}\Theta^{(o)f'}_{fjk}\delta^{(t)(\nu)}_{f'\,l+P\,m+P}\;.
\end{align}

Let us now demonstrate all these formulas!

\begin{subappendices}
\section{Backprop through BatchNorm} \label{sec:appenbatchnorm-vgg}

For Backpropagation, we will need
\begin{align}
\frac{\partial y^{(t')(n)}_{f'\,l'\,m'}}{\partial h_{f\,l\,m}^{(t)(\nu)}}&=
\gamma^{(n)}_f\frac{\partial \tilde{h}_{f'\,l'\,m'}^{(t')(n)}}{\partial h_{f\,l\,m}^{(t)(\nu)}}\;.
\end{align}
Since
\begin{align}
\frac{\partial h^{(t')(\nu)}_{f'\,l'\,m'}}{\partial h_{f\,l\,m}^{(t)(\nu)}}&=\delta^{t'}_t\delta^{f'}_f
\delta^{l'}_l\delta^{m'}_m\;,&
\frac{\partial \hat{h}_{f'}^{(n)}}{\partial h_{f\,l\,m}^{(t)(\nu)}}&=\frac{\delta^{f'}_f}{T_{{\rm mb}}N_nT_n}\:;
\end{align}
and
\begin{align}
\frac{\partial \left(\hat{\sigma}_{f'}^{(n)}\right)^2}{\partial h_{f\,l\,m}^{(t)(\nu)}}&=
\frac{2\delta^{f'}_f}{T_{{\rm mb}}N_nT_n}\left(h_{f\,l\,m}^{(t)(\nu)}-\hat{h}_{f}^{(n)}\right)\;,
\end{align}
we get
\begin{align}
\frac{\partial \tilde{h}_{f'\,l'\,m'}^{(t')(n)}}{\partial h_{f\,l\,m}^{(t)(\nu)}}&=
\frac{\delta^{f'}_f}{T_{{\rm mb}}N_nT_n}\left[\frac{T_{{\rm mb}}N_nT_n\delta^{t'}_t
\delta^{l'}_l\delta^{m'}_m-1}
{\left(\left(\hat{\sigma}_{f}^{(n)}\right)^2+\epsilon\right)^\frac12}-
\frac{\left(h_{f\,l'\,m'}^{(t')(\nu)}-\hat{h}_{f}^{(n)}\right)\left(h_{f\,l\,m}^{(t)(\nu)}-\hat{h}_{f}^{(n)}\right)}
{\left(\left(\hat{\sigma}_{f}^{(n)}\right)^2+\epsilon\right)^\frac32}\right]\notag\\
&=\frac{\delta^{f'}_f}{\left(\left(\hat{\sigma}_{f}^{(n)}\right)^2+\epsilon\right)^\frac12}
\left[\delta^{t'}_t\delta^{l'}_l\delta^{m'}_m-
\frac{1+\tilde{h}_{f\,l'\,m'}^{(t')(n)}\tilde{h}_{f\,l\,m}^{(t)(n)}}{T_{{\rm mb}}N_nT_n}\right]\;.
\end{align}
To ease the notation we will denote
\begin{align}
\tilde{\gamma}^{(n)}_f&=
\frac{\gamma^{(n)}_f}{\left(\left(\hat{\sigma}_{f}^{(n)}\right)^2+\epsilon\right)^\frac12}\;.
\end{align}
so that
\begin{align}
\delta^f_{f'}J^{(tt')(n)}_{f\,l\,m\,l'\,m'}&=
\frac{\partial y_{f'\,l'\,m'}^{(t')(n)}}{\partial h_{f\,l\,m}^{(t)(\nu)}}=
\tilde{\gamma}^{(n)}_f \delta^{f'}_f\left[\delta^{t'}_t\delta^{l'}_l\delta^{m'}_m-
\frac{1+\tilde{h}_{f\,l'\,m'}^{(t')(n)}\tilde{h}_{f\,l\,m}^{(t)(n)}}{T_{{\rm mb}}N_nT_n}\right]\;.
\end{align}

\section{Error rate updates: details} \label{sec:appenderrorrate}

We have for backpropagation from fc to pool
\begin{align}
\delta^{(t)(\nu)}_{flm}&= \frac{\partial }{\partial a_{flm}^{(t)(\nu)}}J^{(t)}(\Theta)=
\sum_{t'=0}^{T_{{\rm mb}}-1}\sum_{f'=0}^{F_{\nu+1}-1}
 \frac{\partial a_{f'}^{(t')(\nu+1)}}{\partial a_{flm}^{(t)(\nu)}} \delta^{(t')(\nu+1)}_{f'}\notag\\
&=\sum_{t'=0}^{T_{{\rm mb}}-1}\sum_{f'=0}^{F_{\nu+1}-1}\sum^{F_\nu-1}_{f''=0}
\sum^{N_{\nu+1}}_{j=0}\sum^{T_{\nu+1}}_{k=0}\Theta^{(o)f'}_{f''\,j\,k}
\frac{\partial h^{(t)(\nu+1)}_{f''j+Pk+P} }{\partial h_{fl+Pm+P}^{(t)(\nu+1)}} \delta^{(t')(\nu+1)}_{f'}\notag\\
&=\sum_{f'=0}^{F_{\nu+1}-1}\Theta^{(o)f'}_{f\,l\,m} \delta^{(t)(\nu+1)}_{f'}\;,
\end{align}

For backpropagation from pool to conv

\begin{align}
\delta^{(t)(\nu)}_{fl+Pm+P}&=
\sum_{t'=0}^{T_{{\rm mb}}-1}\sum^{F_{\nu+1}-1}_{f'=0}\sum^{N_{\nu+1}-1}_{l'=0}\sum^{T_{\nu+1}-1}_{m'=0}
\frac{\partial a_{f'l'm'}^{(t')(\nu+1)}}{\partial a_{f\,l\,m}^{(t)(\nu)}}
\delta_{f'l'm'}^{(t')(\nu+1)}
=\notag\\
&=\sum_{t'=0}^{T_{{\rm mb}}-1}\sum^{F_{\nu+1}-1}_{f'=0}\sum^{N_{\nu+1}-1}_{l'=0}\sum^{T_{\nu+1}-1}_{m'=0}
\frac{\partial y_{f'\,S_P l'+j_{f'l'm'}^{(t')(p)}+P\,S_Pm'+k_{f'l'm'}^{(t')(p)}+P}^{(t')(n)}}{\partial h_{f\,l+P\,m+P}^{(t)(\nu+1)}}
g'\left(a_{f\,l\,m}^{(t)(\nu)}\right)\delta_{f'l'm'}^{(t')(\nu+1)}\notag\\
&=\tilde{\gamma}^{(n)}_fg'\left(a_{f\,l\,m}^{(t)(\nu)}\right)
\sum_{t'=0}^{T_{{\rm mb}}-1}\sum^{N_{\nu+1}-1}_{l'=0}\sum^{T_{\nu+1}-1}_{m'=0}\delta_{fl'm'}^{(t')(\nu+1)}\notag\\
&\left[\delta^{t'}_t\delta^{S_P l'+j_{t'fl'm'}^{(p)}}_l\delta^{S_Pm'+k_{t'fl'm'}^{(p)}}_m-
\frac{1+\tilde{h}_{f\,S_P l'+j_{fl'm'}^{(t')(p)}+P\,S_Pm'+k_{fl'm'}^{(t')(p)}+P}^{(t')(n)}\tilde{h}_{f\,l+P\,m+P}^{(t)(n)}}{T_{{\rm mb}}N_{n}T_{n}}\right]\notag\\
&=g'\left(a_{f\,l\,m}^{(t)(\nu)}\right)\sum_{t'=0}^{T_{{\rm mb}}-1}\sum^{N_{\nu+1}-1}_{l'=0}
\sum^{T_{\nu+1}-1}_{m'=0}\delta_{fl'm'}^{(t')(\nu+1)}\notag\\
&\times J^{(tt')(n)}_{f\,S_P l'+j_{fl'm'}^{(t')(p)}+P\,S_Pm'+k_{fl'm'}^{(t')(p)}+P\,l+P\,m+P}
\end{align}

For backpropagation from conv to conv

\begin{align}
\delta^{(t)(\nu)}_{fl+Pm+P}&=
\sum_{t'=0}^{T_{{\rm mb}}-1}\sum^{F_{\nu+1}-1}_{f'=0}\sum^{N_{\nu+1}-1}_{l'=0}\sum^{T_{\nu+1}-1}_{m'=0}
\frac{\partial a_{f'l'm'}^{(t')(\nu+1)}}{\partial a_{f\,l\,m}^{(t)(\nu)}}
\delta_{f'l'+Pm'+P}^{(t')(\nu+1)}\notag\\
&=\sum_{t'=0}^{T_{{\rm mb}}-1}\sum^{F_{\nu+1}-1}_{f'=0}\sum^{N_{\nu+1}-1}_{l'=0}\sum^{T_{\nu+1}-1}_{m'=0}
\sum^{F_{\nu+1}-1}_{f''=0}\sum^{R_C-1}_{j=0}\sum^{R_C-1}_{k=0}
\Theta^{(o)f'}_{f''\,j\,k}\notag\\
&\times\frac{\partial y^{(t')(n)}_{f''\,l'+j\,m'+k}}{\partial h_{f\,l+P\,m+P}^{(t)(\nu+1)}}
g'\left(a_{f\,l\,m}^{(t)(\nu)}\right)\delta_{f'l'+Pm'+P}^{(t')(\nu+1)}\;,
\end{align}
so
\begin{align}
\delta^{(t)(\nu)}_{fl+Pm+P}&=\tilde{\gamma}^{(n)}_fg'\left(a_{f\,l\,m}^{(t)(\nu)}\right)
\sum_{t'=0}^{T_{{\rm mb}}-1}\sum^{F_{\nu+1}-1}_{f'=0}\sum^{N_{\nu+1}-1}_{l'=0}\sum^{T_{\nu+1}-1}_{m'=0}
\sum^{R_C-1}_{j=0}\sum^{R_C-1}_{k=0}\Theta^{(o)f'}_{f\,j\,k}\delta_{f'l'+Pm'+P}^{(t')(\nu+1)}\notag\\
&\times \left[\delta^{t'}_t\delta^{ l'+j}_{l+P}\delta^{m'+k}_{m+P}-
\frac{1+\tilde{h}_{f\, l'+j\,m'+k}^{(t')(n)}\tilde{h}_{f\,l+P\,m+P}^{(t)(n)}}{T_{{\rm mb}}N_nT_n}\right]\notag\\
&=g'\left(a_{f\,l\,m}^{(t)(\nu)}\right)
\sum_{t'=0}^{T_{{\rm mb}}-1}\sum^{F_{\nu+1}-1}_{f'=0}\sum^{N_{\nu+1}-1}_{l'=0}\sum^{T_{\nu+1}-1}_{m'=0}
\sum^{R_C-1}_{j=0}\sum^{R_C-1}_{k=0}\delta_{f'l'+Pm'+P}^{(t')(\nu+1)}\notag\\
&\times \Theta^{(o)f'}_{f\,j\,k} J^{(tt')(n)}_{f\, l'+j\,m'+k\,l+P\,m+P}\;,
\end{align}

and for backpropagation from conv to pool (taking the stride equal to 1 to simplify the derivation)

\begin{align}
\delta^{(t)(\nu)}_{flm}&=
\sum_{t'=0}^{T_{{\rm mb}}-1}\sum^{F_{\nu+1}-1}_{f'=0}\sum^{N_{\nu+1}-1}_{l'=0}\sum^{T_{\nu+1}-1}_{m'=0}
\frac{\partial a_{f'l'm'}^{(t')(\nu+1)}}{\partial a_{f\,l\,m}^{(t)(\nu)}}
\delta_{f'l'+Pm'+P}^{(t')(\nu+1)}\notag\\
&=\sum_{t'=0}^{T_{{\rm mb}}-1}\sum^{F_{\nu+1}-1}_{f'=0}\sum^{N_{\nu+1}-1}_{l'=0}\sum^{T_{\nu+1}-1}_{m'=0}
\sum^{F_{\nu+1}-1}_{f''=0}\sum^{R_C-1}_{j=0}\sum^{R_C-1}_{k=0}
\Theta^{(o)f'}_{f''\,j\,k}
\frac{\partial h^{(t')(\nu+1)}_{f''\,l'+j\,m'+k}}{\partial h_{f\,l+P\,m+P}^{(t)(\nu+1)}}
\delta_{f'l'+Pm'+P}^{(t')(\nu+1)}\notag\\
&=\sum^{F_{\nu+1}-1}_{f'=0}\sum^{R_C-1}_{j=0}\sum^{R_C-1}_{k=0}
\Theta^{(o)f'}_{f\,j\,k}\delta_{f'l+2P-j\,m+2P-k}^{(t)(\nu+1)}\;.
\end{align}

And so on and so forth.

\section{Weight update: details}
Fc to Fc
\begin{align}
\Delta^{\Theta(o)f}_{f'}&=\frac{1}{T_{{\rm mb}}}\sum_{t=0}^{T_{{\rm mb}}-1}
\sum^{F_{\nu+1}-1}_{f''=0}\sum^{F_\nu}_{f'''=0}\frac{\partial\Theta^{(o)f''}_{f'''}
}{\partial \Theta^{(o)f}_{f'}}y^{(t)(n)}_{f'''}\delta^{(t)(\nu)}_{f''}
=\sum_{t=0}^{T_{{\rm mb}}-1}\delta^{(t)(\nu)}_f y^{(t)(n)}_{f'}\;.
\end{align}
Fc to pool
\begin{align}
\Delta^{\Theta(o)f}_{f'jk}&=\frac{1}{T_{{\rm mb}}}\sum_{t=0}^{T_{{\rm mb}}-1}
\sum^{F_{\nu+1}-1}_{f''=0}\sum^{F_\nu}_{f'''=0}\sum^{N_{\nu+1}}_{j'=0}\sum^{T_{\nu+1}}_{k'=0}
\frac{\partial\Theta^{(13)f''}_{f'''j'k'}
}{\partial \Theta^{(o)f}_{f'jk}}h^{(t)(\nu)}_{f'''j'+Pk'+P}\delta^{(t)(\nu)}_{f''}\notag\\
&=\sum_{t=0}^{T_{{\rm mb}}-1}\delta^{(t)(\nu)}_f h^{(t)(\nu)}_{f'j+Pk+P}\;.
\end{align}
and for conv to conv
\begin{align}
\Delta^{\Theta(o)f}_{f'jk}&=\sum_{t=0}^{T_{{\rm mb}}-1}\sum^{F_{\nu+1}-1}_{f''=0}\sum^{T_{\nu+1}-1}_{l=0}\sum^{N_{\nu+1}-1}_{m=0}
\frac{\partial a_{f''\,l\,m}^{(t)(\nu)}}{\partial \Theta^{(o)f}_{f'\,j\,k}}
\delta_{f''\,l+P\,m+P}^{(t)(\nu)}\notag\\
&=\sum_{t=0}^{T_{{\rm mb}}-1}\sum^{F_{\nu+1}-1}_{f''=0}\sum^{T_{\nu+1}-1}_{l=0}\sum^{N_{\nu+1}-1}_{m=0}
\sum^{F_{\nu+1}-1}_{f'''=0}\sum^{R_C-1}_{j'=0}\sum^{R_C-1}_{k'=0}
\frac{\partial \Theta^{(o)f''}_{f'''\,j'\,k'}}{\partial \Theta^{(o)f}_{f'\,j\,k}}\notag\\
&\times y^{(t)(n)}_{f'''\,S_Cl+j'\,S_Cm+k'}
\delta_{f''\,l+P\,m+P}^{(t)(\nu)}\notag\\
&=\sum_{t=0}^{T_{{\rm mb}}-1}\sum^{T_{\nu+1}-1}_{l=0}\sum^{N_{\nu+1}-1}_{m=0}
y^{(t)(n)}_{f'\,S_Cl+j\,S_Cm+k}\delta_{f\,l+P\,m+P}^{(t)(\nu)}\;.
\end{align}
similarly for conv to pool and conv to input
\begin{align}
\Delta^{\Theta(o)f}_{f'jk}&=\sum_{t=0}^{T_{{\rm mb}}-1}\sum^{T_{\nu+1}-1}_{l=0}\sum^{N_{\nu+1}-1}_{m=0}
h^{(t)(\nu)}_{f'\,S_Cl+j\,S_Cm+k}\delta_{f\,l+P\,m+P}^{(t)(\nu)}\;.
\end{align}
\section{Coefficient update: details}
Fc to Fc
\begin{align}
\Delta_f^{\gamma(n)}&=\sum_{t=0}^{T_{{\rm mb}}-1}\sum_{f'=0}^{F_{\nu+1}-1}
\frac{\partial a^{(t)(\nu+1)}_{f'}}{\partial\gamma_f^{(n)}}\delta^{(t)(\nu+1)}_{f'}
=\sum_{t=0}^{T_{{\rm mb}}-1}\sum_{f'=0}^{F_{\nu+1}-1}
\Theta^{(o)f'}_{f}\tilde{h}^{(t)(n)}_{f}\delta^{(t)({\nu+1})}_{f'}\;,\\
\Delta_f^{\beta(n)}&=\sum_{t=0}^{T_{{\rm mb}}-1}\sum_{f'=0}^{F_{\nu+1}-1}
\frac{\partial a^{(t)(\nu+1)}_{f'}}{\partial\beta_f^{(n)}}\delta^{(t)(\nu+1)}_{f'}
=\sum_{t=0}^{T_{{\rm mb}}-1}\sum_{f'=0}^{F_{\nu+1}-1}\Theta^{(o)f'}_{f}\delta^{(t)(\nu+1)}_{f'}\;,
\end{align}
fc to pool and conv to pool
\begin{align}
\Delta_f^{\gamma(n)}&=
\sum_{t=0}^{T_{{\rm mb}}-1}\sum_{l=0}^{N_{\nu+1}-1}\sum_{m=0}^{T_{\nu+1}-1}
\tilde{h}_{f\,S_P l+j_{flm}^{(t)(p)}+P\,S_Pm+k_{flm}^{(t)(p)}+P}^{(t)(n)}\delta^{(t)(\nu+1)}_{flm}\\
\Delta_f^{\beta(n)}&=
\sum_{t=0}^{T_{{\rm mb}}-1}\sum_{l=0}^{N_{\nu+1}-1}\sum_{m=0}^{T_{\nu+1}-1}\delta^{(t)(\nu+1)}_{flm}\;,
\end{align}
conv to conv
\begin{align}
\Delta_f^{\gamma(n)}&=
\sum_{t=0}^{T_{{\rm mb}}-1}\sum_{f'=0}^{F_{\nu+1}-1}\sum_{l=0}^{N_{\nu+1}-1}\sum_{m=0}^{T_{\nu+1}-1}
\frac{a^{(t)(\nu+1)}_{f'lm}}{\partial \gamma_f^{(n)}}\delta^{(t)(\nu+1)}_{f'lm}\\
&=\sum_{t=0}^{T_{{\rm mb}}-1}\sum_{f'=0}^{F_{\nu+1}-1}\sum_{l=0}^{N_{\nu+1}-1}\sum_{m=0}^{T_{\nu+1}-1}
\sum_{j=0}^{R_C-1}\sum_{k=0}^{R_C-1}\Theta^{(o)f'}_{fjk}
\tilde{h}^{(t)(n)}_{fl+j\,m+k}\delta^{(t)(\nu+1)}_{f'lm}\\
\Delta_f^{\beta(n)}&=
\sum_{t=0}^{T_{{\rm mb}}-1}\sum_{f'=0}^{F_{\nu+1}-1}\sum_{l=0}^{N_{\nu+1}-1}\sum_{m=0}^{T_{\nu+1}-1}
\sum_{j=0}^{R_C-1}\sum_{k=0}^{R_C-1}\Theta^{(o)f'}_{fjk}\delta^{(t)(\nu+1)}_{f'lm}\;,
\end{align}

\section{Practical Simplification} \label{sec:pracsimpl}

When implementing a CNN, it turns out that some of the error rate computation can be very costly (in term of execution time) if naively encoded. In this section, we sketch some improvement that can be performed on the pool to conv, conv to conv error rate implementation, as well as ones on coefficient updates.

\subsection{pool to conv Simplification}

Let us expand the batch normalization term of the pool to conv error rate to see how we can simplify it (calling the pooling the $p$'th one)
\begin{align}
\delta^{(t)(\nu)}_{flm}&=\tilde{\gamma}^{(n)}_fg'\left(a_{f\,l\,m}^{(t)(\nu)}\right)
\sum_{t'=0}^{T_{{\rm mb}}-1}\sum^{N_{\nu+1}-1}_{l'=0}\sum^{T_{\nu+1}-1}_{m'=0}\delta_{fl'm'}^{(t')(\nu+1)}\notag\\
&\left[\delta^{t'}_t\delta^{S_P l'+j_{t'fl'm'}^{(p)}}_l\delta^{S_Pm'+k_{t'fl'm'}^{(p)}}_m-
\frac{1+\tilde{h}_{f\,S_P l'+j_{fl'm'}^{(t')(p)}+P\,S_Pm'+k_{fl'm'}^{(t')(p)}+P}^{(t')(n)}\tilde{h}_{f\,l+P\,m+P}^{(t)(n)}}{T_{{\rm mb}}N_nT_n}\right]\;.
\end{align}
Numerically, this implies that for each $t,f,l,m$ one needs to perform 3 loops (on $t',l',m'$), hence a 7 loop process. This can be reduced to 4 at most in the following way. Defining
\begin{align}
\mu_f^{(1)}&=
\sum_{t'=0}^{T_{{\rm mb}}-1}\sum^{N_{\nu+1}-1}_{l'=0}\sum^{T_{\nu+1}-1}_{m'=0}
\delta_{fl'm'}^{(t')(\nu+1)}\;,
\end{align}
and
\begin{align}
\mu_f^{(2)}&=
\sum_{t'=0}^{T_{{\rm mb}}-1}\sum^{N_{\nu+1}-1}_{l'=0}\sum^{T_{\nu+1}-1}_{m'=0}\delta_{fl'm'}^{(t')(\nu+1)}
\tilde{h}_{f\,S_P l'+j_{fl'm'}^{(t')(p)}+P\,S_Pm'+k_{fl'm'}^{(t')(p)}+P}^{(t')(n)}\;,
\end{align}
we have introduced two new variables that can be computed in four loops, but three of them are independent of the ones needed to compute $\delta^{(t)(\nu)}_{flm}$. For the last term, the $\delta$ functions "kill" 3 loops and we are left with
\begin{align}
\delta^{(t)(\nu)}_{flm}=\tilde{\gamma}^{(n)}_fg'\left(a_{f\,l\,m}^{(t)(\nu)}\right)&\left\{
\sum^{N_{\nu+1}-1}_{l'=0}\sum^{T_{\nu+1}-1}_{m'=0}\delta_{fl'm'}^{(t')(\nu+1)}
\delta^{S_P l'+j_{t'fl'm'}^{(p)}}_l\delta^{S_Pm'+k_{t'fl'm'}^{(p)}}_m\right.\notag\\
&-\left.
\frac{\mu_f^{(1)}+\mu_f^{(2)}\tilde{h}_{f\,l+P\,m+P}^{(t)(n)}}{T_{{\rm mb}}N_nT_n}\right\}\;,
\end{align}
which requires only 4 loops to be computed.

\subsection{Convolution Simplification}

Let us expand the batch normalization term of the conv to conv error rate to see how we can simplify it
\begin{align}
\delta^{(t)(\nu)}_{flm}&=\tilde{\gamma}^{(n)}_fg'\left(a_{f\,l\,m}^{(t)(i\nu)}\right)
\sum_{t'=0}^{T_{{\rm mb}}-1}\sum^{F_{\nu+1}-1}_{f'=0}\sum^{N_{\nu+1}-1}_{l'=0}\sum^{T_{\nu+1}-1}_{m'=0}
\sum^{R_C-1}_{j=0}\sum^{R_C-1}_{k=0}\Theta^{(o)f'}_{f\,j\,k}\delta_{f'l'm'}^{(t')(\nu+1)}\notag\\
&\left[\delta^{t'}_t\delta^{ l'+j}_{l+P}\delta^{m'+k}_{m+P}-
\frac{1+\tilde{h}_{f\, l'+j\,m'+k}^{(t')(n)}\tilde{h}_{f\,l+P\,m+P}^{(t)(n)}}{T_{{\rm mb}}N_nT_n}\right]\;.
\end{align}
If left untouched, one now needs 10 loops (on $t,f,l,m$ and $t',f',l',m',j,k$) to compute $\delta^{(t)(\nu)}_{flm}$ ! This can be reduced to 7 loops at most in the following way. First, we define
\begin{align}
\lambda^{(t)(1)}_{flm}&=\sum^{F_{\nu+1}-1}_{f'=0}\sum^{R_C-1}_{j=0}\sum^{R_C-1}_{k=0}
\Theta^{(o)f'}_{f\,j\,k}\delta_{f'l+P-j\,m+P-k}^{(t')(\nu+1)}\;,
\end{align}
which is a convolution operation on a shifted index $\delta^{(\nu+1)}$. This is second most expensive operation, but libraries are optimized for this and it implies two of the smallest loops (those on $R_C$). Then we compute (four loops each)
\begin{align}
\lambda^{(2)}_{ff'}&=\sum^{R_C-1}_{j=0}\sum^{R_C-1}_{k=0}\Theta^{(o)f'}_{f\,j\,k}\;,&
\lambda^{(3)}_{f'}&=\sum_{t'=0}^{T_{{\rm mb}}-1}\sum^{N_{\nu+1}-1}_{l'=0}\sum^{T_{\nu+1}-1}_{m'=0}
\delta_{f'l'm'}^{(t')(\nu+1)}\;,
\end{align}
and (2 loops)
\begin{align}
\lambda^{(4)}_{f}&=\sum^{F_{\nu+1}-1}_{f'=0}\lambda^{(2)}_{ff'}\lambda^{(3)}_{f'}\;.
\end{align}
Finally, we compute
\begin{align}
\lambda^{(5)}_{ff'jk}&=\sum_{t'=0}^{T_{{\rm mb}}-1}\sum^{N_{\nu+1}-1}_{l'=0}\sum^{T_{\nu+1}-1}_{m'=0}
\delta_{f'l'm'}^{(t')(\nu+1)}\tilde{h}_{f\, l'+j\,m'+k}^{(t')(n)}\;,\\
\lambda^{(6)}_{f}&=\sum^{F_{\nu+1}-1}_{f'=0}\sum^{R_C-1}_{j=0}\sum^{R_C-1}_{k=0}
\Theta^{(o)f'}_{f\,j\,k}\lambda^{(5)}_{ff'jk}\;.
\end{align}
$\lambda^{(6)}_{f}$ only requires four loops, and $\lambda^{(5)}_{ff'jk}$ is the most expensive operation. But it is also a convolution operation that implies two of the smallest loops (those on $R_C$). With all these newly introduced $\lambda$, we obtain
\begin{align}
\delta^{(t)(\nu)}_{flm}&=\tilde{\gamma}^{(n)}_fg'\left(a_{f\,l\,m}^{(t)(\nu)}\right)\left\{
\lambda^{(t)(1)}_{flm}-
\frac{\lambda^{(4)}_{f}+\lambda^{(6)}_{f}\tilde{h}_{f\,l+P\,m+P}^{(t)(n)}}{T_{{\rm mb}}N_nT_n}\right\}\;,
\end{align}
which only requires four loops to be computed.

\subsection{Coefficient Simplification}

To compute
\begin{align}
\Delta_f^{\gamma(n)}
&=\sum_{t=0}^{T_{{\rm mb}}-1}\sum_{f'=0}^{F_{\nu+1}-1}\sum_{l=0}^{N_{\nu+1}-1}\sum_{m=0}^{T_{\nu+1}-1}
\sum_{j=0}^{R_C-1}\sum_{k=0}^{R_C-1}\Theta^{(o)f'}_{fjk}
\tilde{h}^{(t)(n)}_{fl+j\,m+k}\delta^{(t)(\nu+1)}_{f'lm}\\
\Delta_f^{\beta(n)}&=
\sum_{t=0}^{T_{{\rm mb}}-1}\sum_{f'=0}^{F_{\nu+1}-1}\sum_{l=0}^{N_{\nu+1}-1}\sum_{m=0}^{T_{\nu+1}-1}
\sum_{j=0}^{R_C-1}\sum_{k=0}^{R_C-1}\Theta^{(o)f'}_{fjk}\delta^{(t)(\nu+1)}_{f'lm}\;,
\end{align}
we will first define
\begin{align}
\nu^{(1)}_{f'fjk}&=\sum_{t=0}^{T_{{\rm mb}}-1}\sum_{l=0}^{N_{\nu+1}-1}\sum_{m=0}^{T_{\nu+1}-1}
\tilde{h}^{(t)(n)}_{fl+j\,m+k}\delta^{(t)(\nu+1)}_{f'lm}\;,&
\nu^{(2)}_{f'}&=\sum_{t=0}^{T_{{\rm mb}}-1}\sum_{l=0}^{N_{\nu+1}-1}\sum_{m=0}^{T_{\nu+1}-1}
\delta^{(t)(\nu+1)}_{f'lm}\;,
\end{align}
so that
\begin{align}
\Delta_f^{\gamma(n)}&=\sum_{f'=0}^{F_{\nu+1}-1}\sum_{j=0}^{R_C-1}
\sum_{k=0}^{R_C-1}\nu^{(1)}_{f'fjk}\Theta^{(o)f'}_{fjk}\;\\
\Delta_f^{\beta(n)}&=
\sum_{f'=0}^{F_{\nu+1}-1}\sum_{j=0}^{R_C-1}\sum_{k=0}^{R_C-1}\nu^{(2)}_{f'}\Theta^{(o)f'}_{fjk}\;,
\end{align}

\section{Batchpropagation through a ResNet module}

For pedagogical reasons, we introduced the ResNet structure without "breaking" the conv layers. Nevertheless, a more standard choice is depicted in the following figure

\begin{figure}[H]
\begin{center}
\begin{tikzpicture}[baseline=-0pt]
\node at (0,0) {\includegraphics[scale=1]{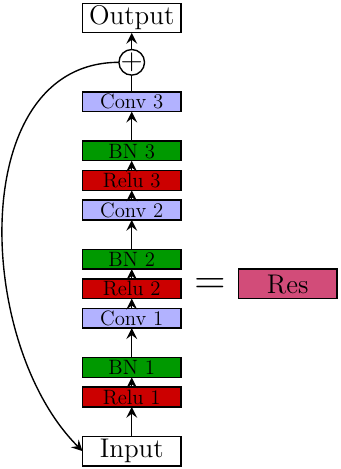}};
\end{tikzpicture}
\caption{Batchpropagation through a ResNet module.}
\end{center}
\end{figure}

Batchpropagation through this ResNet module presents no particular difficulty. Indeed, the update rules imply the usual conv to conv backpropagation derived in the main part of this note. The only novelty is the error rate update of the input layer of the ResNet, as it now reads (assuming that the input of this ResNet module is the output of another one)

\begin{align}
\delta^{(t)(\nu)}_{fl+Pm+P}&=
\sum_{t'=0}^{T_{{\rm mb}}-1}\sum^{F_{\nu+1}-1}_{f'=0}\sum^{N_{\nu+1}-1}_{l'=0}\sum^{T_{\nu+1}-1}_{m'=0}
\frac{\partial a_{f'l'm'}^{(t')(\nu+1)}}{\partial a_{f\,l\,m}^{(t)(\nu)}}
\delta_{f'l'+Pm'+P}^{(t')(\nu+1)}\notag\\
&+\sum_{t'=0}^{T_{{\rm mb}}-1}\sum^{F_{\nu+3}-1}_{f'=0}\sum^{N_{\nu+3}-1}_{l'=0}\sum^{T_{\nu+3}-1}_{m'=0}
\frac{\partial a_{f'l'm'}^{(t')(\nu+3)}}{\partial a_{f\,l\,m}^{(t)(\nu)}}
\delta_{f'l'+Pm'+P}^{(t')(\nu+3)}\notag\\
&=g'\left(a_{f\,l\,m}^{(t)(\nu)}\right)
\sum_{t'=0}^{T_{{\rm mb}}-1}\sum^{F_{\nu+1}-1}_{f'=0}\sum^{N_{\nu+1}-1}_{l'=0}\sum^{T_{\nu+1}-1}_{m'=0}
\sum^{R_C-1}_{j=0}\sum^{R_C-1}_{k=0}\delta_{f'l'+Pm'+P}^{(t')(\nu+1)}\notag\\
&\times \Theta^{(o)f'}_{f\,j\,k} J^{(tt')(n)}_{f\, l'+j\,m'+k\,l+P\,m+P}
+\delta_{fl+Pm+P}^{(t)(\nu+3)}\;.
\end{align}
This new term is what allows the error rate to flow smoothly from the output to the input in the ResNet CNN, as the additional connexion in a ResNet is like a skip path to the convolution chains. Let us mention in passing that some architecture connects every hidden layer to each others\cite{HuangGLZLW}.

\section{Convolution as a matrix multiplication}

Thanks to the simplifications introduced in appendix \ref{sec:pracsimpl}, we have reduced all convolution, pooling and tensor multiplication operations to at most 7 loops operations. Nevertheless, in high abstraction programming languages such as python, it is still way too much to be handled smoothly. But there exists additional tricks to "reduce" the dimension of the convolution operation, such as one has only to encode three for loops (2D matrix multiplication) at the end of the day. Let us begin our presentation of these tricks with a 2D convolution example

\subsection{2D Convolution}

\begin{figure}[H]
\begin{center}
\begin{tikzpicture}[baseline=-0pt]
\node at (0,0) {\includegraphics[scale=0.8]{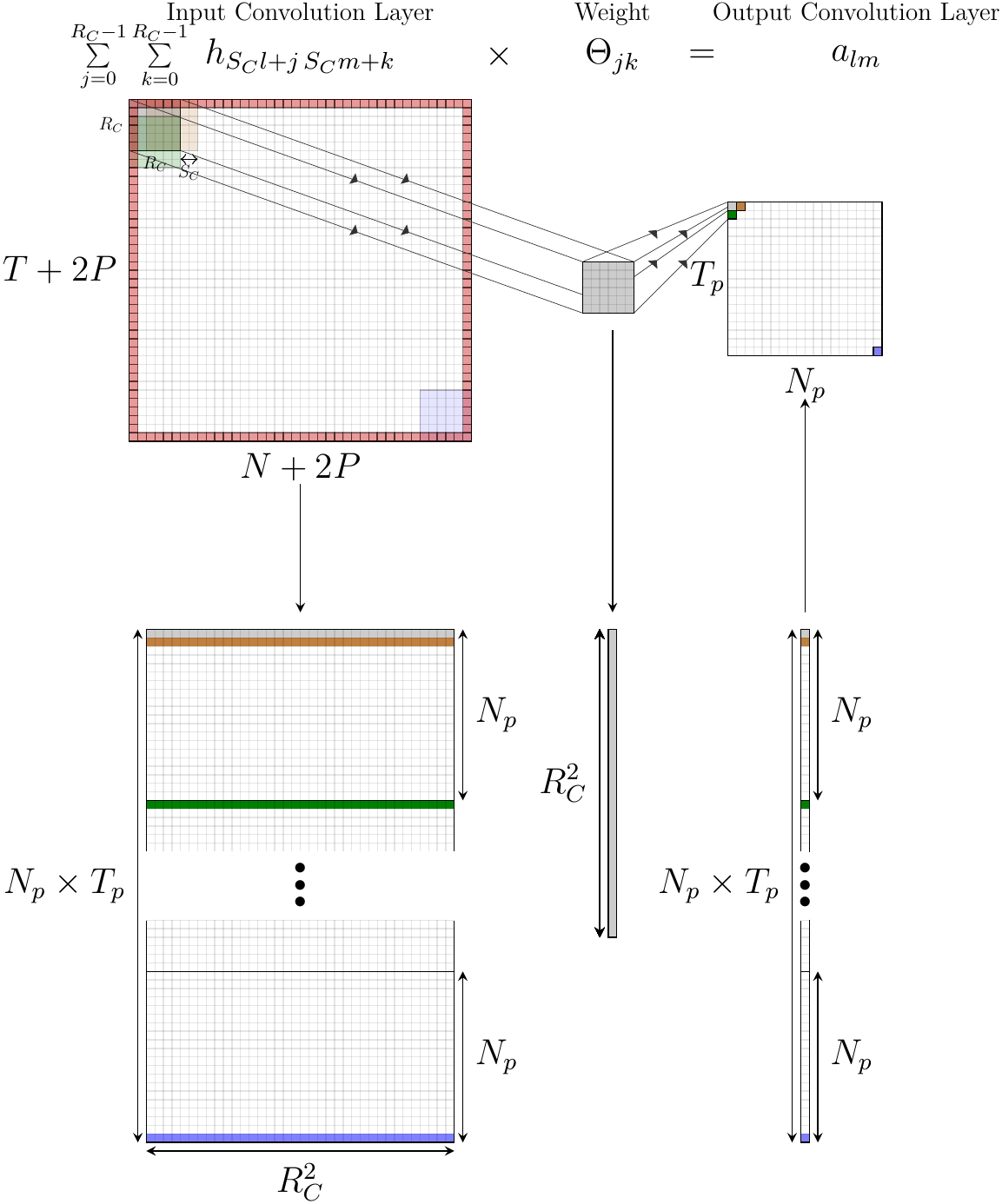}};
\end{tikzpicture}
\caption{\label{fig:2dconv}:2D convolution as a 2D matrix multiplication}
\end{center}
\end{figure}

A 2D convolution operation reads
\begin{align}
a_{lm}&=\sum_{j=0}^{R_C-1}\sum_{k=0}^{R_C-1}\Theta_{jk}h_{S_Cl+j\,S_Cm+k}\;.
\end{align}
and it involves 4 loops. The trick to reduce this operation to a 2D matrix multiplication is to redefine the $h$ matrix by looking at each $h$ indices are going to be multiplied by $\Theta$ for each value of $l$ and $m$. Then the associated $h$ values are stored into a $N_pT_p\times R_C^2$ matrix. Flattening out the $\Theta$ matrix, we are left with a matrix multiplication, the flattened $\Theta$ matrix being of size $R_C^2\times 1$. This is illustrated on figure \ref{fig:2dconv}

\subsection{4D Convolution}

\begin{figure}[H]
\begin{center}
\begin{tikzpicture}[baseline=-0pt]
\node at (0,0) {\includegraphics[scale=0.75]{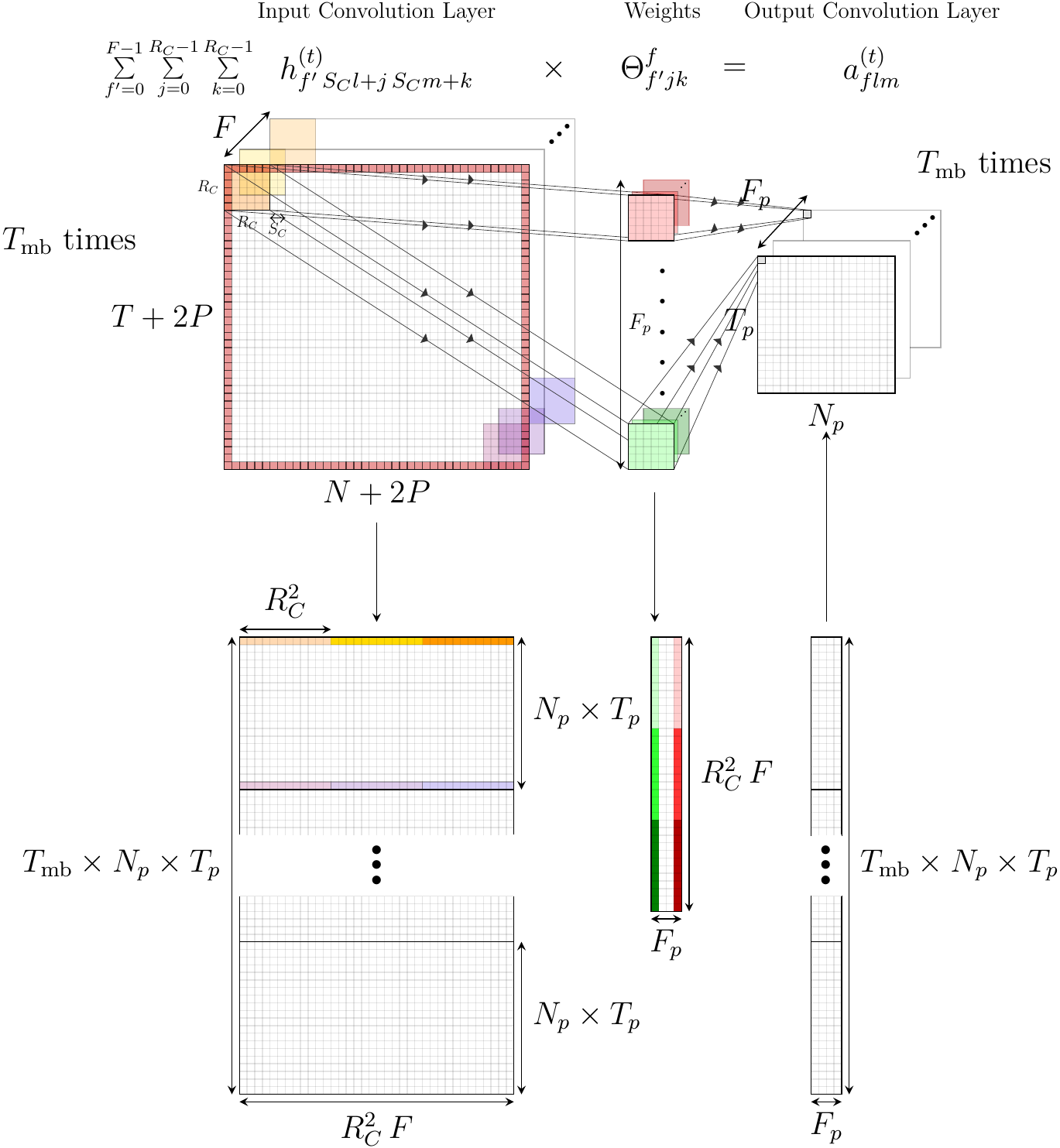}};
\end{tikzpicture}
\caption{\label{fig:4dconv}4D convolution as a 2D matrix multiplication}
\end{center}
\end{figure}

Following the same lines, the adding of the input and output feature maps as well as the batch size poses no particular conceptual difficulty, as illustrated on figure \ref{fig:4dconv}, corresponding to the 4D convolution
\begin{align}
a^{(t)}_{flm}&=
\sum_{f'=0}^{F_p-1}\sum_{j=0}^{R_C-1}\sum_{k=0}^{R_C-1}\Theta^f_{f'jk}h^{(t)}_{f'S_Cl+j\,S_Cm+k}\;.
\end{align}

\section{Pooling as a row matrix maximum}

\begin{figure}[H]
\begin{center}
\begin{tikzpicture}[baseline=-0pt]
\node at (0,0) {\includegraphics[scale=0.75]{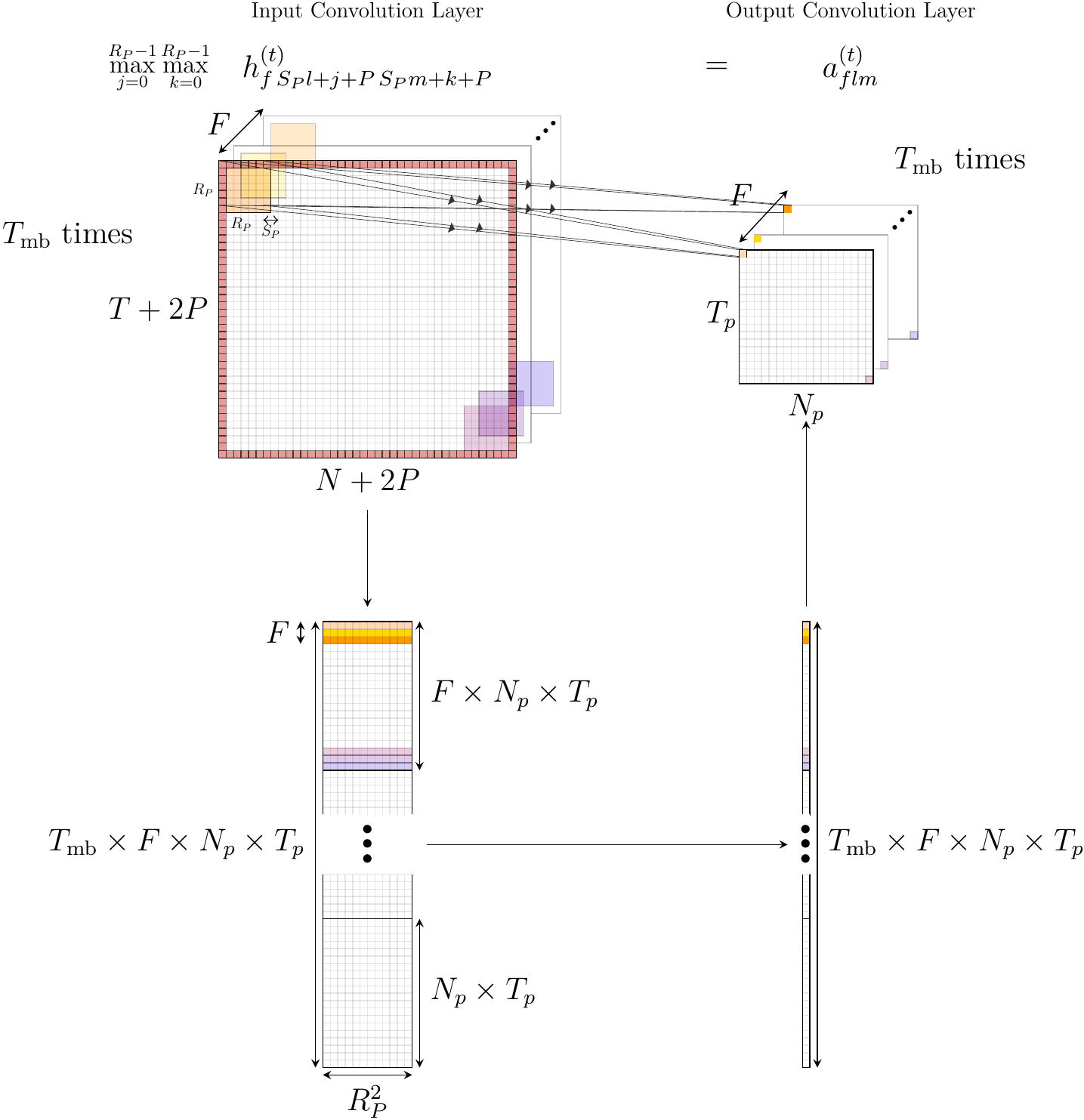}};
\end{tikzpicture}
\caption{\label{fig:4dpool}4D pooling as a 2D matrix multiplication}
\end{center}
\end{figure}

The pooling operation can also be simplified, seeing it as the maximum search on the rows of a flattened 2D matrix. This is illustrated on figure \ref{fig:4dpool}
\begin{align}
a^{(t)}_{flm}&=
\max_{j=0}^{R_P-1}\max_{k=0}^{R_P-1}h^{(t)}_{fS_Pl+j\,S_Pm+k}\;.
\end{align}
\end{subappendices}

\chapter{Recurrent Neural Networks} \label{sec:chapterRNN}

\minitoc

\section{Introduction}

\yinipar{\fontsize{60pt}{72pt}\usefont{U}{Kramer}{xl}{n}I}n this chapter, we review a third kind of Neural Network architecture: Recurrent Neural Networks\cite{GravesA2016}. By contrast with the CNN, this kind of network introduces a real architecture novelty : instead of forwarding only in a "spatial" direction, the data are also forwarded in a new -- time dependent -- direction. We will present the first Recurrent Neural Network (RNN) architecture, as well as the current most popular one: the Long Short Term Memory (LSTM) Neural Network. 

\section{RNN-LSTM architecture}

\subsection{Forward pass in a RNN-LSTM}

In figure \ref{fig:1}, we present the RNN architecture in a schematic way

\begin{figure}[H]
\begin{center}
\begin{tikzpicture}
\node at (0,0) [rectangle,draw,fill=gray!0!white] (h00) {$h^{(00)}$};
\node at (0,1.5) [rectangle,draw,fill=gray!30!white] (h01) {$h^{(10)}$};
\node at (0,3) [rectangle,draw,fill=gray!30!white] (h02) {$h^{(20)}$};
\node at (0,4.5) [rectangle,draw,fill=gray!30!white] (h03) {$h^{(30)}$};
\node at (0,6) [rectangle,draw,fill=gray!0!white] (h04) {$h^{(40)}$};
\node at (1.8,0) [rectangle,draw,fill=gray!0!white] (h10) {$h^{(01)}$};
\node at (1.8,1.5) [rectangle,draw,fill=gray!70!white] (h11) {$h^{(11)}$};
\node at (1.8,3) [rectangle,draw,fill=gray!70!white] (h12) {$h^{(21)}$};
\node at (1.8,4.5) [rectangle,draw,fill=gray!70!white] (h13) {$h^{(31)}$};
\node at (1.8,6) [rectangle,draw,fill=gray!0!white] (h14) {$h^{(41)}$};
\node at (3.6,0) [rectangle,draw,fill=gray!0!white] (h20) {$h^{(02)}$};
\node at (3.6,1.5) [rectangle,draw,fill=gray!70!white] (h21) {$h^{(12)}$};
\node at (3.6,3) [rectangle,draw,fill=gray!70!white] (h22) {$h^{(22)}$};
\node at (3.6,4.5) [rectangle,draw,fill=gray!70!white] (h23) {$h^{(32)}$};
\node at (3.6,6) [rectangle,draw,fill=gray!0!white] (h24) {$h^{(42)}$};
\node at (5.4,0) [rectangle,draw,fill=gray!0!white] (h30) {$h^{(03)}$};
\node at (5.4,1.5) [rectangle,draw,fill=gray!70!white] (h31) {$h^{(13)}$};
\node at (5.4,3) [rectangle,draw,fill=gray!70!white] (h32) {$h^{(23)}$};
\node at (5.4,4.5) [rectangle,draw,fill=gray!70!white] (h33) {$h^{(33)}$};
\node at (5.4,6) [rectangle,draw,fill=gray!0!white] (h34) {$h^{(43)}$};
\node at (7.2,0) [rectangle,draw,fill=gray!0!white] (h40) {$h^{(04)}$};
\node at (7.2,1.5) [rectangle,draw,fill=gray!70!white] (h41) {$h^{(14)}$};
\node at (7.2,3) [rectangle,draw,fill=gray!70!white] (h42) {$h^{(24)}$};
\node at (7.2,4.5) [rectangle,draw,fill=gray!70!white] (h43) {$h^{(34)}$};
\node at (7.2,6) [rectangle,draw,fill=gray!0!white] (h44) {$h^{(44)}$};
\node at (9,0) [rectangle,draw,fill=gray!0!white] (h50) {$h^{(05)}$};
\node at (9,1.5) [rectangle,draw,fill=gray!70!white] (h51) {$h^{(15)}$};
\node at (9,3) [rectangle,draw,fill=gray!70!white] (h52) {$h^{(25)}$};
\node at (9,4.5) [rectangle,draw,fill=gray!70!white] (h53) {$h^{(35)}$};
\node at (9,6) [rectangle,draw,fill=gray!0!white] (h54) {$h^{(45)}$};
\node at (10.8,0) [rectangle,draw,fill=gray!0!white] (h60) {$h^{(06)}$};
\node at (10.8,1.5) [rectangle,draw,fill=gray!70!white] (h61) {$h^{(16)}$};
\node at (10.8,3) [rectangle,draw,fill=gray!70!white] (h62) {$h^{(26)}$};
\node at (10.8,4.5) [rectangle,draw,fill=gray!70!white] (h63) {$h^{(36)}$};
\node at (10.8,6) [rectangle,draw,fill=gray!0!white] (h64) {$h^{(46)}$};
\node at (12.6,0) [rectangle,draw,fill=gray!0!white] (h70) {$h^{(07)}$};
\node at (12.6,1.5) [rectangle,draw,fill=gray!70!white] (h71) {$h^{(17)}$};
\node at (12.6,3) [rectangle,draw,fill=gray!70!white] (h72) {$h^{(27)}$};
\node at (12.6,4.5) [rectangle,draw,fill=gray!70!white] (h73) {$h^{(37)}$};
\node at (12.6,6) [rectangle,draw,fill=gray!0!white] (h74) {$h^{(47)}$};
\draw[-stealth] (h00) -- node[pos=0.5,anchor=east,scale=1] {$\Theta^{\nu(1)}$} (h01);
\draw[-stealth] (h01) -- node[pos=0.5,anchor=east,scale=1] {$\Theta^{\nu(2)}$} (h02);
\draw[-stealth] (h02) -- node[pos=0.5,anchor=east,scale=1] {$\Theta^{\nu(3)}$} (h03);
\draw[dotted,-stealth] (h03) -- node [pos=0.5,anchor = east] {$\Theta$} (h04);
\draw[-stealth] (h10) -- (h11);
\draw[-stealth] (h11) -- (h12);
\draw[-stealth] (h12) -- (h13);
\draw[dotted,-stealth] (h13) -- (h14);
\draw[-stealth] (h20) -- (h21);
\draw[-stealth] (h21) -- (h22);
\draw[-stealth] (h22) -- (h23);
\draw[dotted,-stealth] (h23) -- (h24);
\draw[-stealth] (h30) -- (h31);
\draw[-stealth] (h31) -- (h32);
\draw[-stealth] (h32) -- (h33);
\draw[dotted,-stealth] (h33) -- (h34);
\draw[-stealth] (h40) -- (h41);
\draw[-stealth] (h41) -- (h42);
\draw[-stealth] (h42) -- (h43);
\draw[dotted,-stealth] (h43) -- (h44);
\draw[-stealth] (h50) -- (h51);
\draw[-stealth] (h51) -- (h52);
\draw[-stealth] (h52) -- (h53);
\draw[dotted,-stealth] (h53) -- (h54);
\draw[-stealth] (h60) -- (h61);
\draw[-stealth] (h61) -- (h62);
\draw[-stealth] (h62) -- (h63);
\draw[dotted,-stealth] (h63) -- (h64);
\draw[-stealth] (h70) -- (h71);
\draw[-stealth] (h71) -- (h72);
\draw[-stealth] (h72) -- (h73);
\draw[dotted,-stealth] (h73) -- (h74);
\draw[-stealth] (h01) --   node[pos=0.5,above=7pt,scale=1] {$\Theta^{\tau(1)}$} (h11);
\draw[-stealth] (h11) -- (h21);
\draw[-stealth] (h21) -- (h31);
\draw[-stealth] (h31) -- (h41);
\draw[-stealth] (h41) -- (h51);
\draw[-stealth] (h51) -- (h61);
\draw[-stealth] (h61) -- (h71);
\draw[-stealth] (h02) -- node[pos=0.5,above=7pt,scale=1] {$\Theta^{\tau(2)}$} (h12);
\draw[-stealth] (h12) -- (h22);
\draw[-stealth] (h22) -- (h32);
\draw[-stealth] (h32) -- (h42);
\draw[-stealth] (h42) -- (h52);
\draw[-stealth] (h52) -- (h62);
\draw[-stealth] (h62) -- (h72);
\draw[-stealth] (h03) -- node[pos=0.5,above=7pt,scale=1] {$\Theta^{\tau(3)}$} (h13);
\draw[-stealth] (h13) -- (h23);
\draw[-stealth] (h23) -- (h33);
\draw[-stealth] (h33) -- (h43);
\draw[-stealth] (h43) -- (h53);
\draw[-stealth] (h53) -- (h63);
\draw[-stealth] (h63) -- (h73);
\draw[very thin,densely dashed,-stealth] (h04) to[out=45,in=225] (h10);
\draw[very thin,densely dashed,-stealth] (h14) to[out=45,in=225] (h20);
\draw[very thin,densely dashed,-stealth] (h24) to[out=45,in=225] (h30);
\draw[very thin,densely dashed,-stealth] (h34) to[out=45,in=225] (h40);
\draw[very thin,densely dashed,-stealth] (h44) to[out=45,in=225] (h50);
\draw[very thin,densely dashed,-stealth] (h54) to[out=45,in=225] (h60);
\draw[very thin,densely dashed,-stealth] (h64) to[out=45,in=225] (h70);
\end{tikzpicture}
\caption{\label{fig:RNN architecture}RNN architecture, with data propagating both in "space" and in "time". In our exemple, the time dimension is of size 8 while the "spatial" one is of size 4.}
\end{center}
\end{figure}
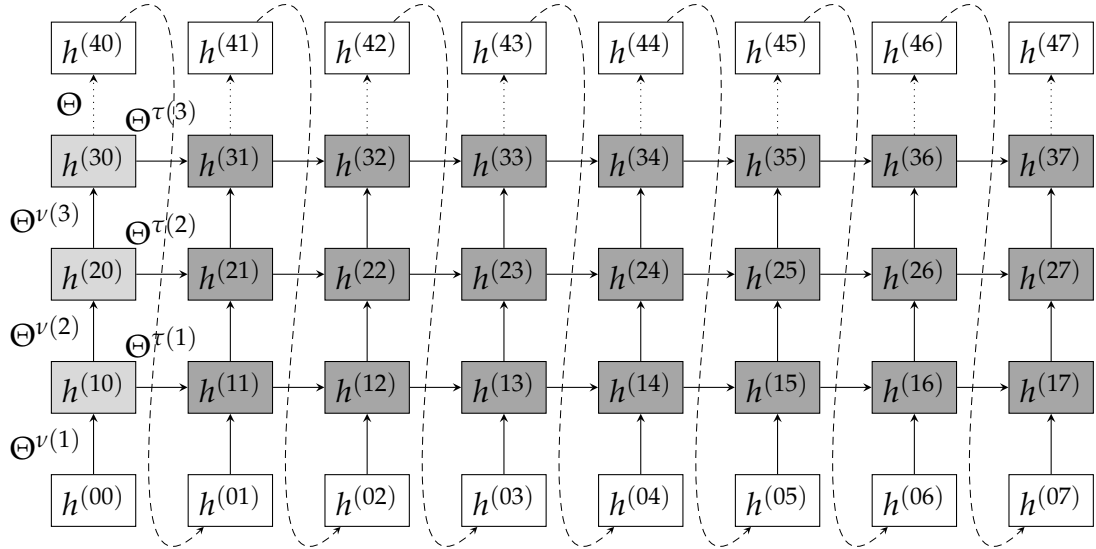

The real novelty of this type of neural network is that the fact that we are trying to predict a time serie is encoded in the very architecture of the network. RNN have first been introduced mostly to predict the next words in a sentance (classification task), hence the notion of ordering in time of the prediction. But this kind of network architecture can also be applied to regression problems. Among others things one can think of stock prices evolution, or temperature forecasting. In contrast to the precedent neural networks that we introduced, where we defined (denoting $\nu$ as in previous chapters the layer index in the spatial direction)
\begin{align}
a^{(t)(\nu)}_{f}&= \text{ Weight Averaging } \left(h^{(t)(\nu)}_{f}\right)\;,\notag\\
h^{(t)(\nu+1)}_{f}&= \text{ Activation function } \left(a^{(t)(\nu)}_{f}\right)\;,
\end{align}
we now have the hidden layers that are indexed by both a "spatial" and a "temporal" index (with $T$ being the network dimension in this new direction), and the general philosophy of the RNN is (now the $a$ is usually characterized by a $c$ for cell state, this denotation, trivial for the basic RNN architecture will make more sense when we talk about LSTM networks)
\begin{align}
c^{(t)(\nu \tau )}_{f}&= \text{ Weight Averaging } \left(h^{(t)(\nu \tau-1)}_{f},h^{(t)(\nu-1\tau)}_{f}\right)\;,\notag\\
h^{(t)(\nu\tau)}_{f}&= \text{ Activation function } \left(c^{(t)(\nu \tau)}_{f}\right)\;,
\end{align}

\subsection{Backward pass in a RNN-LSTM}

The backward pass in a RNN-LSTM has to respect a certain time order, as illustrated in the following figure

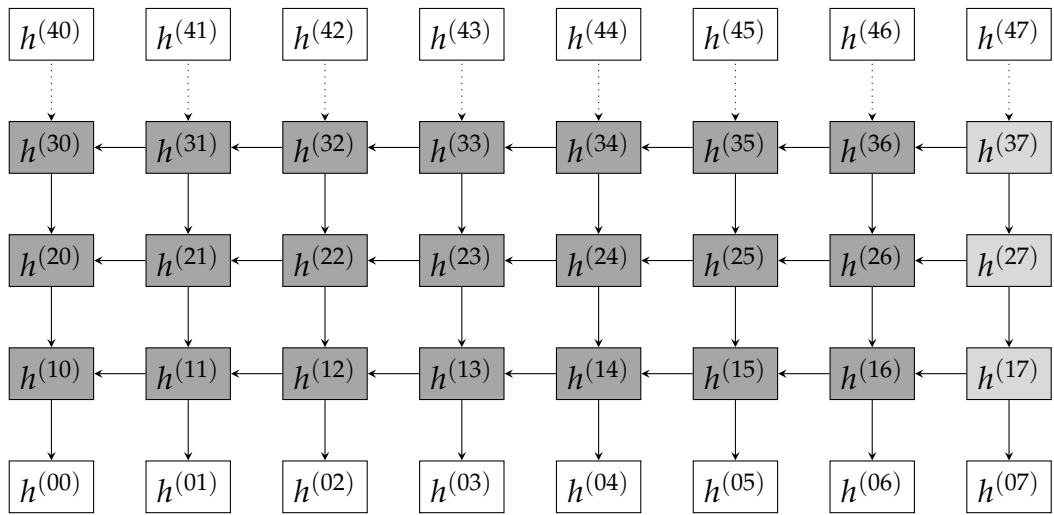
\begin{figure}[H]
\begin{center}
\begin{tikzpicture}
\node at (0,0) [rectangle,draw,fill=gray!0!white] (h00) {$h^{(00)}$};
\node at (0,1.5) [rectangle,draw,fill=gray!70!white] (h01) {$h^{(10)}$};
\node at (0,3) [rectangle,draw,fill=gray!70!white] (h02) {$h^{(20)}$};
\node at (0,4.5) [rectangle,draw,fill=gray!70!white] (h03) {$h^{(30)}$};
\node at (0,6) [rectangle,draw,fill=gray!0!white] (h04) {$h^{(40)}$};
\node at (1.8,0) [rectangle,draw,fill=gray!0!white] (h10) {$h^{(01)}$};
\node at (1.8,1.5) [rectangle,draw,fill=gray!70!white] (h11) {$h^{(11)}$};
\node at (1.8,3) [rectangle,draw,fill=gray!70!white] (h12) {$h^{(21)}$};
\node at (1.8,4.5) [rectangle,draw,fill=gray!70!white] (h13) {$h^{(31)}$};
\node at (1.8,6) [rectangle,draw,fill=gray!0!white] (h14) {$h^{(41)}$};
\node at (3.6,0) [rectangle,draw,fill=gray!0!white] (h20) {$h^{(02)}$};
\node at (3.6,1.5) [rectangle,draw,fill=gray!70!white] (h21) {$h^{(12)}$};
\node at (3.6,3) [rectangle,draw,fill=gray!70!white] (h22) {$h^{(22)}$};
\node at (3.6,4.5) [rectangle,draw,fill=gray!70!white] (h23) {$h^{(32)}$};
\node at (3.6,6) [rectangle,draw,fill=gray!0!white] (h24) {$h^{(42)}$};
\node at (5.4,0) [rectangle,draw,fill=gray!0!white] (h30) {$h^{(03)}$};
\node at (5.4,1.5) [rectangle,draw,fill=gray!70!white] (h31) {$h^{(13)}$};
\node at (5.4,3) [rectangle,draw,fill=gray!70!white] (h32) {$h^{(23)}$};
\node at (5.4,4.5) [rectangle,draw,fill=gray!70!white] (h33) {$h^{(33)}$};
\node at (5.4,6) [rectangle,draw,fill=gray!0!white] (h34) {$h^{(43)}$};
\node at (7.2,0) [rectangle,draw,fill=gray!0!white] (h40) {$h^{(04)}$};
\node at (7.2,1.5) [rectangle,draw,fill=gray!70!white] (h41) {$h^{(14)}$};
\node at (7.2,3) [rectangle,draw,fill=gray!70!white] (h42) {$h^{(24)}$};
\node at (7.2,4.5) [rectangle,draw,fill=gray!70!white] (h43) {$h^{(34)}$};
\node at (7.2,6) [rectangle,draw,fill=gray!0!white] (h44) {$h^{(44)}$};
\node at (9,0) [rectangle,draw,fill=gray!0!white] (h50) {$h^{(05)}$};
\node at (9,1.5) [rectangle,draw,fill=gray!70!white] (h51) {$h^{(15)}$};
\node at (9,3) [rectangle,draw,fill=gray!70!white] (h52) {$h^{(25)}$};
\node at (9,4.5) [rectangle,draw,fill=gray!70!white] (h53) {$h^{(35)}$};
\node at (9,6) [rectangle,draw,fill=gray!0!white] (h54) {$h^{(45)}$};
\node at (10.8,0) [rectangle,draw,fill=gray!0!white] (h60) {$h^{(06)}$};
\node at (10.8,1.5) [rectangle,draw,fill=gray!70!white] (h61) {$h^{(16)}$};
\node at (10.8,3) [rectangle,draw,fill=gray!70!white] (h62) {$h^{(26)}$};
\node at (10.8,4.5) [rectangle,draw,fill=gray!70!white] (h63) {$h^{(36)}$};
\node at (10.8,6) [rectangle,draw,fill=gray!0!white] (h64) {$h^{(46)}$};
\node at (12.6,0) [rectangle,draw,fill=gray!0!white] (h70) {$h^{(07)}$};
\node at (12.6,1.5) [rectangle,draw,fill=gray!30!white] (h71) {$h^{(17)}$};
\node at (12.6,3) [rectangle,draw,fill=gray!30!white] (h72) {$h^{(27)}$};
\node at (12.6,4.5) [rectangle,draw,fill=gray!30!white] (h73) {$h^{(37)}$};
\node at (12.6,6) [rectangle,draw,fill=gray!0!white] (h74) {$h^{(47)}$};
\draw[stealth-] (h00) -- (h01);
\draw[stealth-] (h01) -- (h02);
\draw[stealth-] (h02) -- (h03);
\draw[dotted,stealth-] (h03) -- (h04);
\draw[stealth-] (h10) -- (h11);
\draw[stealth-] (h11) -- (h12);
\draw[stealth-] (h12) -- (h13);
\draw[dotted,stealth-] (h13) -- (h14);
\draw[stealth-] (h20) -- (h21);
\draw[stealth-] (h21) -- (h22);
\draw[stealth-] (h22) -- (h23);
\draw[dotted,stealth-] (h23) -- (h24);
\draw[stealth-] (h30) -- (h31);
\draw[stealth-] (h31) -- (h32);
\draw[stealth-] (h32) -- (h33);
\draw[dotted,stealth-] (h33) -- (h34);
\draw[stealth-] (h40) -- (h41);
\draw[stealth-] (h41) -- (h42);
\draw[stealth-] (h42) -- (h43);
\draw[dotted,stealth-] (h43) -- (h44);
\draw[stealth-] (h50) -- (h51);
\draw[stealth-] (h51) -- (h52);
\draw[stealth-] (h52) -- (h53);
\draw[dotted,stealth-] (h53) -- (h54);
\draw[stealth-] (h60) -- (h61);
\draw[stealth-] (h61) -- (h62);
\draw[stealth-] (h62) -- (h63);
\draw[dotted,stealth-] (h63) -- (h64);
\draw[stealth-] (h70) -- (h71);
\draw[stealth-] (h71) -- (h72);
\draw[stealth-] (h72) -- (h73);
\draw[dotted,stealth-] (h73) -- (h74);
\draw[stealth-] (h01) -- (h11);
\draw[stealth-] (h11) -- (h21);
\draw[stealth-] (h21) -- (h31);
\draw[stealth-] (h31) -- (h41);
\draw[stealth-] (h41) -- (h51);
\draw[stealth-] (h51) -- (h61);
\draw[stealth-] (h61) -- (h71);
\draw[stealth-] (h02) -- (h12);
\draw[stealth-] (h12) -- (h22);
\draw[stealth-] (h22) -- (h32);
\draw[stealth-] (h32) -- (h42);
\draw[stealth-] (h42) -- (h52);
\draw[stealth-] (h52) -- (h62);
\draw[stealth-] (h62) -- (h72);
\draw[stealth-] (h03) -- (h13);
\draw[stealth-] (h13) -- (h23);
\draw[stealth-] (h23) -- (h33);
\draw[stealth-] (h33) -- (h43);
\draw[stealth-] (h43) -- (h53);
\draw[stealth-] (h53) -- (h63);
\draw[stealth-] (h63) -- (h73);
\end{tikzpicture}
\caption{\label{fig:rnnback}Architecture taken, backward pass. Here what cannot compute the gradient of a layer without having computed the ones that flow into it}
\end{center}
\end{figure}

With this in mind, let us now see in details the implementation of a RNN and its advanced cousin, the Long Short Term Memory (LSTM)-RNN.

\section{Extreme Layers and loss function}

These part of the RNN-LSTM networks just experiences trivial modifications. Let us see them

\subsection{Input layer}

In a RNN-LSTM, the input layer is recursively defined as 
\begin{align}
h^{(t)(0\tau+1)}_{f}&=\left(\tilde{h}^{(t)(0\tau)}_{f},h^{(t)(N-1\tau)}_{f}\right)\;.
\end{align}
where $\tilde{h}^{(t)(0\tau)}_{f}$ is $h^{(t)(0\tau)}_{f}$ with the first time column removed.

\subsection{Output layer }

The output layer of a RNN-LSTM reads
\begin{align}
h^{(t)(N\tau)}_{f}&=o\left(\sum_{f'=0}^{F_{N-1}-1}\Theta^f_{f'} h^{(t)(N-1\tau)}_{f}\right)\;,
\end{align}
where the output function $o$ is as for FNN's and CNN's is either the identity (regression task) or the cross-entropy function (classification task).

\subsection{Loss function}

The loss function for a regression task reads
\begin{align}
J(\Theta)&=\frac{1}{2T_{{\rm mb}}}\sum_{t=0}^{T_{{\rm mb}}-1}\sum_{\tau=0}^{T-1}\sum_{f=0}^{F_N-1}
\left(h^{(t)( N\tau)}_f-y^{(t)(\tau)}_f\right)^2\;.
\end{align}
and for a classification task
\begin{align}
J(\Theta)&=-\frac{1}{T_{{\rm mb}}}\sum_{t=0}^{T_{{\rm mb}}-1}\sum_{\tau=0}^{T-1}\sum_{c=0}^{C-1}
\delta^{c}_{y^{(t)(\tau)}_c}  \ln \left(h^{(t)( N\tau)}_f\right)\;.
\end{align}

\section{RNN specificities}

\subsection{RNN structure} \label{sec:rnnstructure}

RNN is the most basic architecture that takes -- thanks to the way it is built in -- into account the time structure of the data to be predicted. Zooming on one hidden layer of \ref{fig:RNN architecture}, here is what we see for a simple Recurrent Neural Network.

\begin{figure}[H]
\begin{center}
\begin{tikzpicture}
\node[] at (0,0) {\includegraphics[scale=1.5]{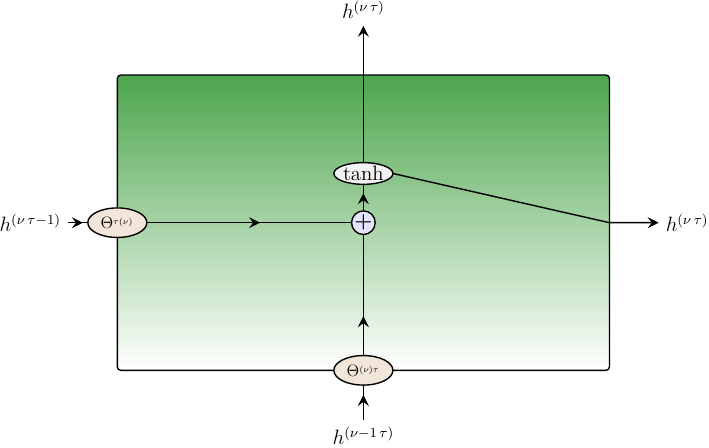}};
\end{tikzpicture}
\caption{\label{fig:RNN hidden unit}RNN hidden unit details}
\end{center}
\end{figure}

And here is how the output of the hidden layer represented in \ref{fig:RNN hidden unit} enters into the subsequent hidden units

\begin{figure}[H]
\begin{center}
\begin{tikzpicture}
\node[] at (0,0) {\includegraphics[scale=0.8]{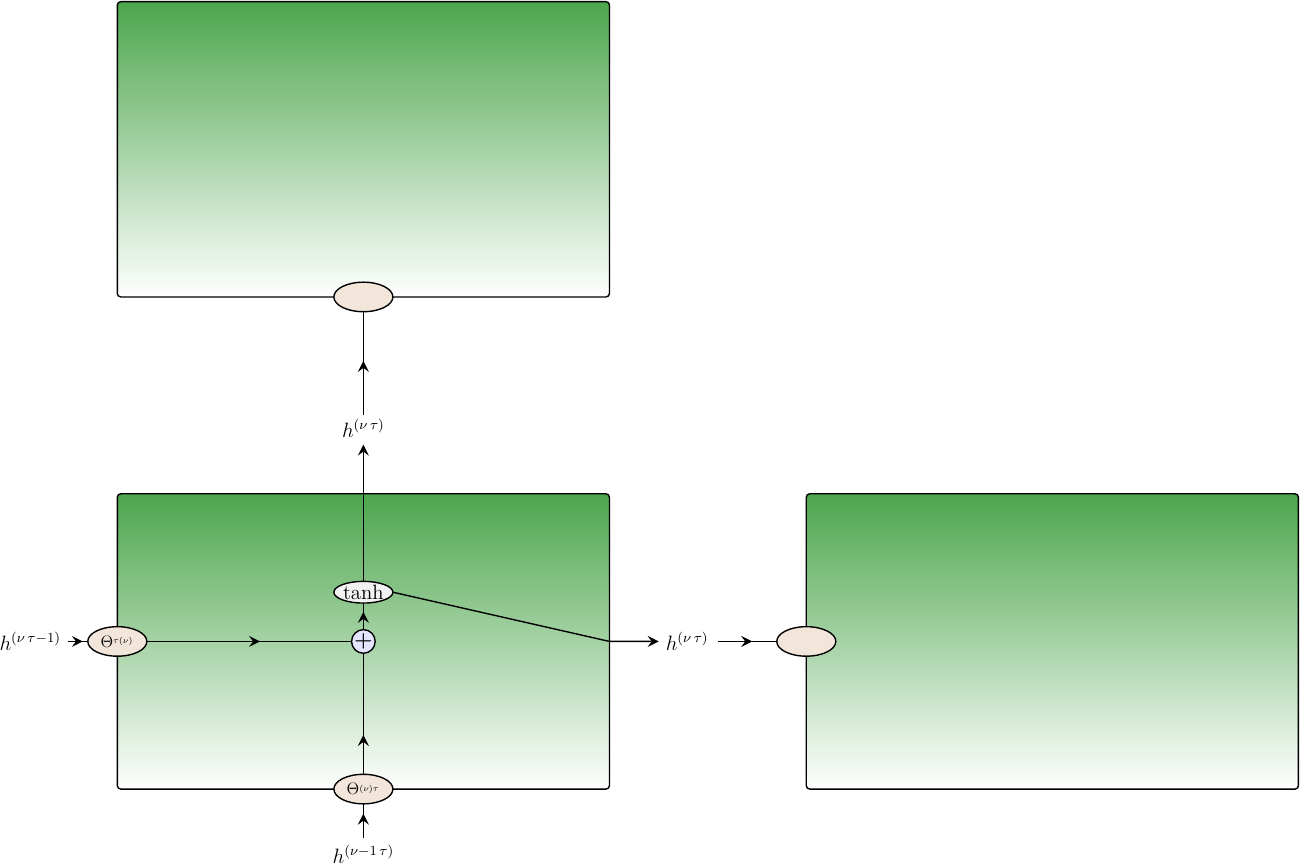}};
\end{tikzpicture}
\caption{\label{fig:RNN interaction}How the RNN hidden unit interact with each others}
\end{center}
\end{figure}

Lest us now mathematically express what is reprensented in figures \ref{fig:RNN hidden unit} and \ref{fig:RNN interaction}.

\subsection{Forward pass in a RNN}

 In a RNN, the update rules read for the first time slice (spatial layer at the extreme left of figure \ref{fig:RNN architecture})
 
\begin{align}
h^{(t)(\nu\tau)}_f&=\tanh\left(\sum_{f'=0}^{F_{{\nu-1}}-1}\Theta^{\nu(\nu)f}_{f'}
h^{(t)(\nu-1\tau)}_{f'}\right)\;,
\end{align}

and for the other ones

\begin{align}
h^{(t)(\nu\tau)}_f&=\tanh\left(\sum_{f'=0}^{F_{{\nu-1}}-1}\Theta^{\nu(\nu)f}_{f'}
h^{(t)(\nu-1\tau)}_{f'}+\sum_{f'=0}^{F_{{\nu}}-1}\Theta^{\tau(\nu)f}_{f'}
h^{(t)(\nu\tau-1)}_{f'}\right)\;.
\end{align}

\subsection{Backpropagation in a RNN}

The backpropagation philosophy will remain unchanged : find the error rate updates, from which one can deduce the weight updates. But as for the hidden layers, the $\delta$ now have both a spatial and a temporal component. We will thus have to compute 
\begin{align}
\delta^{(t)( \nu\tau)}_f&=\frac{\delta}{\delta h^{(t)( \nu+1\tau)}_f }J(\Theta)\;,
\end{align}
to deduce
\begin{align}
\Delta^{\Theta{\rm index}f}_{f'}&=\frac{\delta}{\delta \Delta^{\Theta{\rm index}f}_{f'} }J(\Theta)\;,
\end{align}
where the index can either be nothing (weights of the ouput layers), $\nu(\nu)$ (weights between two spatially connected layers) or  $\tau(\nu)$ (weights between two temporally connected layers). First, it is easy to compute (in the same way as in chapter 1 for FNN) for the MSE loss function
\begin{align}
\delta^{(t)(N-1\tau)}_f&= \frac{1}{T_{{\rm mb}}}\left(h_{f}^{(t)(N\tau)}-y_f^{(t)(\tau)}\right)\;,
\end{align}
and for the cross entropy loss function
\begin{align}
\delta^{(t)(N-1)}_{f}&= \frac{1}{T_{{\rm mb}}}\left(h_{f}^{(t)(N\tau)}-\delta^f_{y^{(t)(\tau)}}\right)\;.
\end{align}
Calling 
\begin{align}
\mathcal{T}_{f}^{(t)(\nu\tau)}&=1-\left(h_{f}^{(t)(\nu\tau)}\right)^2\;,
\end{align}
and
\begin{align}
\mathcal{H}^{(t')(\nu\tau)_a}_{ff'}&=\mathcal{T}^{(t')(\nu+1\tau)}_{f'}\Theta^{a(\nu+1)f'}_{f}\;,
\end{align}
we show in appendix \ref{sec:rnnappenderrorrate} that (if $\tau+1$ exists, otherwise the second term is absent)
\begin{align}
\delta^{(t)(\nu-1\tau)}_f&= 
\sum_{t'=0}^{T_{{\rm mb}}}J^{(tt')(\nu\tau)}_f\sum_{\epsilon=0}^{1}\sum_{f'=0}^{F_{\nu+1-\epsilon}-1}
\mathcal{H}^{(t')(\nu-\epsilon\tau+\epsilon)_{b_\epsilon}}_{ff'}\delta^{(t')(\nu-\epsilon\tau+\epsilon)}_{f'}\;.
\end{align}
where $b_0=\nu$ and $b_1=\tau$.

\subsection{Weight and coefficient updates in a RNN}

To complete the backpropagation algorithm, we need

\begin{align}
&\Delta^{\nu(\nu)f}_{f'}\;,&
&\Delta^{\tau(\nu)f}_{f'}\;,&
&\Delta^{f}_{f'}\;,&
&\Delta^{\beta(\nu \tau)}_{f}\;,&
&\Delta^{\gamma(\nu \tau)}_{f}\;.
\end{align}

We show in appendix \ref{sec:rnncoefficient} that

\begin{align}
\Delta^{\nu(\nu-)f}_{f'}&=\sum_{\tau=0}^{T-1}\sum_{t=0}^{T_{{\rm mb}}-1}
\mathcal{T}^{(t)(\nu\tau)}_{f}\delta^{(t)(\nu-1\tau)}_{f}h^{(t)(\nu-1\tau)}_{f'}\;,\\
\Delta^{\tau(\nu)f}_{f'}&=\sum_{\tau=1}^{T-1}\sum_{t=0}^{T_{{\rm mb}}-1}
\mathcal{T}^{(t)(\nu\tau)}_{f}\delta^{(t)(\nu-1\tau)}_{f}h^{(t)(\nu\tau-1)}_{f'}\;,\\
\Delta^{f}_{f'}&=\sum_{\tau=0}^{T-1}\sum_{t=0}^{T_{{\rm mb}}-1} h^{(t)(N-1\tau)}_{f'}\delta^{(t)(N-1\tau)}_{f}\;,\\
\Delta^{\beta(\nu \tau)}_{f}&=\sum_{t=0}^{T_{{\rm mb}}-1}\sum_{\epsilon=0}^{1}\sum_{f'=0}^{F_{\nu+1-\epsilon}-1}
\mathcal{H}^{(t')(\nu-\epsilon\tau+\epsilon)_{b_\epsilon}}_{ff'}\delta^{(t')(\nu-\epsilon\tau+\epsilon)}_{f'}\;,\\
\Delta^{\gamma(\nu \tau)}_{f}&=\sum_{t=0}^{T_{{\rm mb}}-1}\tilde{h}^{(t)(\nu\tau)}_{f}\sum_{\epsilon=0}^{1}\sum_{f'=0}^{F_{\nu+1-\epsilon}-1}
\mathcal{H}^{(t')(\nu-\epsilon\tau+\epsilon)_{b_\epsilon}}_{ff'}\delta^{(t')(\nu-\epsilon\tau+\epsilon)}_{f'}\;.
\end{align}

\section{LSTM specificities}

\subsection{LSTM structure}

\begin{figure}[H]
\begin{center}
\begin{tikzpicture}
\node[] at (0,0) {\includegraphics[scale=1.7]{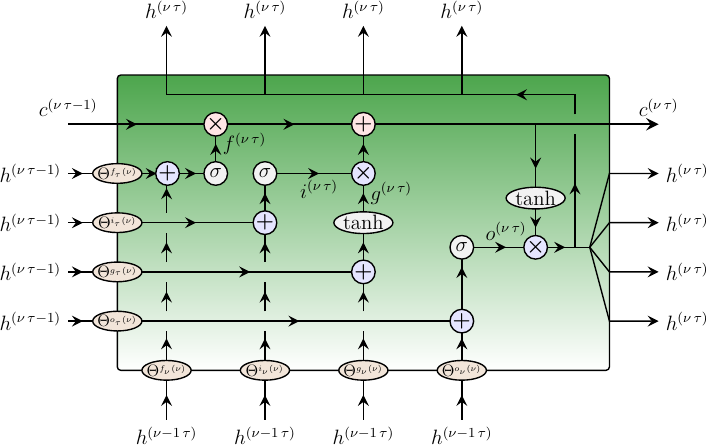}};
\end{tikzpicture}
\caption{\label{fig:Lstm1}LSTM hidden unit details}
\end{center}
\end{figure}

In a Long Short Term Memory Neural Network\cite{Gers:2000:LFC:1121912.1121915}, the state of a given unit is not directly determined by its left and bottom neighbours. Instead, a cell state is updated for each hidden unit, and the output of this unit is a probe of the cell state. This formulation might seem puzzling at first, but it is philosophically similar to the ResNet approach that we briefly encounter in the appendix of chapter \ref{sec:chapterFNN}: instead of trying to fit an input with a complicated function, we try to fit tiny variation of the input, hence allowing the gradient to flow in a smoother manner in the network. In the LSTM network, several gates are thus introduced : the input gate $i^{(t)(\nu\tau)}_f$ determines if we allow new information $g^{(t)(\nu\tau)}_f$ to enter into the cell state. The  output gate $o^{(t)(\nu\tau)}_f$ determines if we set or not the output hidden value to $0$, or really probes the current cell state. Finally, the forget state $f^{(t)(\nu\tau)}_f$ determines if we forget or not the past cell state. All theses concepts are illustrated on the figure \ref{fig:Lstm1}, which is the LSTM counterpart of the RNN structure of section \ref{sec:rnnstructure}. This diagram will be explained in details in the next section.

In a LSTM, the different hidden units interact in the following way

\begin{figure}[H]
\begin{center}
\begin{tikzpicture}
\node[] at (0,0) {\includegraphics[scale=0.8]{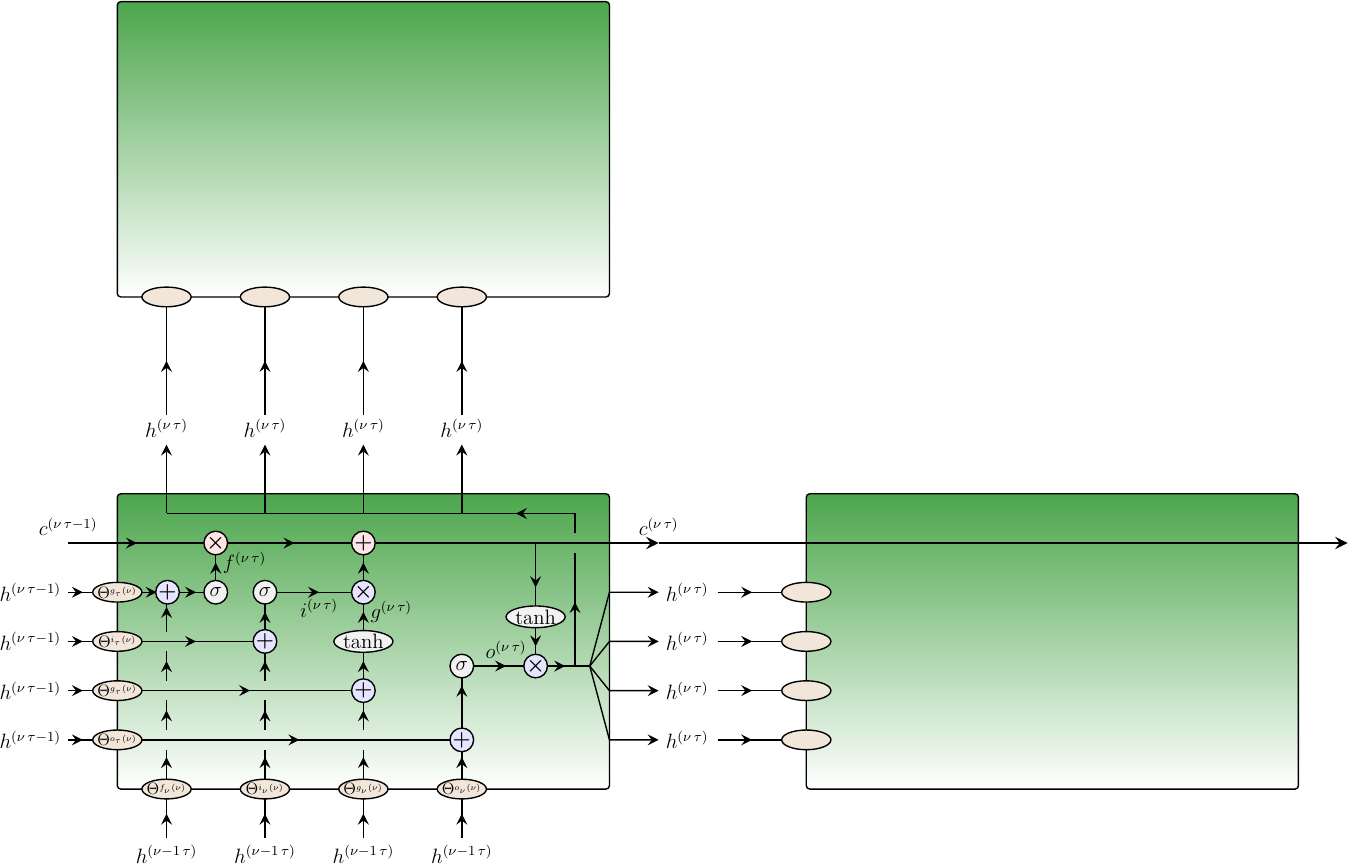}};
\end{tikzpicture}
\caption{\label{fig:Lstmall}How the LSTM hidden unit interact with each others}
\end{center}
\end{figure}

\subsection{Forward pass in LSTM}

Considering all the $\tau-1$ variable values to be $0$ when $\tau=0$, we get the following formula for the input, forget and output gates

\begin{align}
i^{(t)(\nu\tau)}_f&=\sigma\left(\sum_{f'=0}^{F_{{\nu-1}}-1}\Theta^{i_{_\nu}(\nu)f}_{f'}
h^{(t)(\nu-1\tau)}_{f'}+\sum_{f'=0}^{F{_{\nu}}-1}\Theta^{i_{_\tau}(\nu)f}_{f'}
h^{(t)(\nu\tau-1)}_{f'}\right)\;,\\
f^{(t)(\nu\tau)}_f&=\sigma\left(\sum_{f'=0}^{F{_{\nu-1}}-1}\Theta^{f_{_\nu}(\nu)f}_{f'}
h^{(t)(\nu-1\tau)}_{f'}+\sum_{f'=0}^{F_{{\nu}}-1}\Theta^{f_{_\tau}(\nu)f}_{f'}
h^{(t)(\nu\tau-1)}_{f'}\right)\;,\\
o^{(t)(\nu\tau)}_f&=\sigma\left(\sum_{f'=0}^{F_{{\nu-1}}-1}\Theta^{o_{_\nu}(\nu)f}_{f'}
h^{(t)(\nu-1\tau)}_{f'}+\sum_{f'=0}^{F_{{\nu}}-1}\Theta^{o_{_\tau}(\nu)f}_{f'}
h^{(t)(\nu\tau-1)}_{f'}\right)\;.
\end{align}
The sigmoid function is the reason why the $i,f,o$ functions are called gates: they take their values between $0$ and $1$, therefore either allowing or forbidding information to pass through the next step. The cell state update is then performed in the following way
\begin{align}
g^{(t)(\nu\tau)}_f&=\tanh\left(\sum_{f'=0}^{F_{{\nu-1}}-1}\Theta^{g_{_\nu}(\nu)f}_{f'}
h^{(t)(\nu-1\tau)}_{f'}+\sum_{f'=0}^{F_{{\nu}}-1}\Theta^{g_{_\tau}(\nu)f}_{f'}
h^{(t)(\nu\tau-1)}_{f'}\right)\;,\\
c^{(t)(\nu\tau)}_{f}&=
f^{(t)(\nu\tau)}_{f}c^{(t)(\nu\tau-1)}_{f}+i^{(t)(\nu\tau)}_{f}g^{(t)(\nu\tau)}_{f}\;,
\end{align}
and as announced, hidden state update is just a probe of the current cell state
\begin{align}
h^{(t)(\nu\tau)}_{f}&=o^{(t)(\nu\tau)}_{f}\tanh \left(c^{(t)(\nu\tau)}_{f}\right)\;.
\end{align}

These formula singularly complicates the feed forward and especially the backpropagation procedure. For completeness, we will us nevertheless carefully derive it. Let us mention in passing that recent studies tried to replace the tanh activation function of the hidden state $h^{(t)(\nu\tau)}_{f}$ and the cell update $g^{(t)(\nu\tau)}_f$ by Rectified Linear Units, and seems to report better results with a proper initialization of all the weight matrices, argued to be diagonal
\begin{align}
\Theta^f_{f'}(\text{init})&=\frac12\delta^f_{f'}\left(+\sqrt{\frac{6}{F_{\rm in}+F_{\rm out}}}\right)\;,
\end{align}
with the bracket term here to possibly (or not) include some randomness into the initialization

\subsection{Batch normalization}

In batchnorm The update rules for the gates are modified as expected

\begin{align}
i^{(t)(\nu\tau)}_f&=\sigma\left(\sum_{f'=0}^{F_{{\nu-1}}-1}\Theta^{i_\nu(\nu-)f}_{f'}
y^{(t)(\nu-1\tau)}_{f'}+\sum_{f'=0}^{F{_{\nu}}-1}\Theta^{i_\tau(-\nu)f}_{f'}
y^{(t)(\nu\tau-1)}_{f'}\right)\;,\\
f^{(t)(\nu\tau)}_f&=\sigma\left(\sum_{f'=0}^{F{_{\nu-1}}-1}\Theta^{f_\nu(\nu-)f}_{f'}
y^{(t)(\nu-1\tau)}_{f'}+\sum_{f'=0}^{F_{{\nu}}-1}\Theta^{f_\tau(-\nu)f}_{f'}
y^{(t)(\nu\tau-1)}_{f'}\right)\;,\\
o^{(t)(\nu\tau)}_f&=\sigma\left(\sum_{f'=0}^{F_{{\nu-1}}-1}\Theta^{o_\nu(\nu-)f}_{f'}
y^{(t)(\nu-1\tau)}_{f'}+\sum_{f'=0}^{F_{{\nu}}-1}\Theta^{o_\tau(-\nu)f}_{f'}
y^{(t)(\nu\tau-1)}_{f'}\right)\;,\\
g^{(t)(\nu\tau)}_f&=\tanh\left(\sum_{f'=0}^{F_{{\nu-1}}-1}\Theta^{g_\nu(\nu-)f}_{f'}
y^{(t)(\nu-1\tau)}_{f'}+\sum_{f'=0}^{F_{{\nu}}-1}\Theta^{g_\tau(-\nu)f}_{f'}
y^{(t)(\nu\tau-1)}_{f'}\right)\;,
\end{align}
where
\begin{align}
y^{(t)(\nu\tau)}_{f}&=\gamma^{(\nu\tau)}_{f}\tilde{h}^{(t)(\nu\tau)}_{f}+\beta^{(\nu\tau)}_{f}\;,
\end{align}
as well as
\begin{align}
\tilde{h}^{(t)(\nu\tau)}_{f}&=\frac{h^{(t)(\nu\tau)}_{f}-
\hat{h}^{(\nu\tau)}_{f}}{\sqrt{\left(\sigma^{(\nu\tau)}_{f}\right)^2+\epsilon}}
\end{align}
and
\begin{align}
\hat{h}^{(\nu\tau)}_{f}&=\frac{1}{T_{{\rm mb}}}\sum_{t=0}^{T_{{\rm mb}}-1}h^{(t)(\nu\tau)}_{f}\;,&
\left(\sigma^{(\nu\tau)}_{f}\right)^2&=\frac{1}{T_{{\rm mb}}}\sum_{t=0}^{T_{{\rm mb}}-1}\left(h^{(t)(\nu\tau)}_{f}
-\hat{h}^{(\nu\tau)}_{f}\right)^2\;.
\end{align}
It is important to compute a running sum for the mean and the variance, that will serve for the evaluation of the cross-validation and the test set (calling $e$ the number of iterations/epochs) 
\begin{align}
\mathbb{E}\left[h_{f}^{(t)(\nu\tau)}\right]_{e+1} &=
\frac{e\mathbb{E}\left[h_{f}^{(t)(\nu\tau)}\right]_{e}+\hat{h}_{f}^{(\nu\tau)}}{e+1}\;,\\
\mathbb{V}ar\left[h_{f}^{(t)(\nu\tau)}\right]_{e+1} &=
\frac{e\mathbb{V}ar\left[h_{f}^{(t)(\nu\tau)}\right]_{e}+\left(\hat{\sigma}_{f}^{(\nu\tau)}\right)^2}{e+1}
\end{align}
and what will be used at the end is $\mathbb{E}\left[h_{f}^{(t)(\nu\tau)}\right]$ and $\frac{T_{{\rm mb}}}{T_{{\rm mb}}-1}\mathbb{V}ar\left[h_{f}^{(t)(\nu\tau)}\right]$.

\subsection{Backpropagation in a LSTM} \label{sec:appendbackproplstm}

The backpropagation In a LSTM keeps the same structure as in a RNN, namely 

\begin{align}
\delta^{(t)(N-1\tau)}_f&= \frac{1}{T_{{\rm mb}}}\left(h_{f}^{(t)(N\tau)}-y_f^{(t)(\tau)}\right)\;,
\end{align}
and (shown in appendix \ref{sec:ARNNLSTMerror_rates})
\begin{align}
\delta^{(t)(\nu-1\tau)}_f&= 
\sum_{t'=0}^{T_{{\rm mb}}}J^{(tt')(\nu\tau)}_f\sum_{\epsilon=0}^{1}\sum_{f'=0}^{F_{\nu+1-\epsilon}-1}
\mathcal{H}^{(t')(\nu-\epsilon\tau+\epsilon)_{b_\epsilon}}_{ff'}\delta^{(t')(\nu-\epsilon\tau+\epsilon)}_{f'}\;.
\end{align}
What changes is the form of $\mathcal{H}$, now given by
\begin{align}
\mathcal{O}^{(t)(\nu\tau)}_{f}&=h^{(t)(\nu\tau)}_{f}
\left(1-o^{(t)(\nu\tau)}_{f}\right)\;,\notag\\
\mathcal{I}^{(t)(\nu\tau)}_{f}&=o^{(t)(\nu\tau)}_{f}\left(1-\tanh^2\left(c^{(t)(\nu\tau)}_{f}\right)
\right)g^{(t)(\nu\tau)}_{f} i^{(t)(\nu\tau)}_{f}\left(1-i^{(t)(\nu\tau)}_{f}\right)\;,\notag\\
\mathcal{F}^{(t)(\nu\tau)}_{f}&=o^{(t)(\nu\tau)}_{f}\left(1-\tanh^2\left(c^{(t)(\nu\tau)}_{f}\right)
\right)c^{(t)(\nu\tau-1)}_{f} f^{(t)(\nu\tau)}_{f}\left(1-f^{(t)(\nu\tau)}_{f}\right)\;,\notag\\
\mathcal{G}^{(t)(\nu\tau)}_{f}&=o^{(t)(\nu\tau)}_{f}\left(1-\tanh^2\left(c^{(t)(\nu\tau)}_{f}\right)
\right)i^{(t)(\nu\tau)}_{f}\left(1-\left(g^{(t)(\nu\tau)}_{f}\right)^2\right)\;,
\end{align}

and

\begin{align}
H^{(t)(\nu\tau)_a}_{ff'}&=\Theta^{o_a(\nu+1)f'}_{f}\mathcal{O}^{(t)(\nu+1\tau)}_{f'}
+\Theta^{f_a(\nu+1)f'}_{f}\mathcal{F}^{(t)(\nu+1\tau)}_{f'}\notag\\
&+\Theta^{g_a(\nu+1)f'}_{f}\mathcal{G}^{(t)(\nu+1\tau)}_{f'}
+\Theta^{i_a(\nu+1)f'}_{f}\mathcal{I}^{(t)(\nu+1\tau)}_{f'}\;.
\end{align}

\subsection{Weight and coefficient updates in a LSTM}

As for the RNN, (but with the $\mathcal{H}$ defined in section \ref{sec:appendbackproplstm}), we get for $\nu=1$
\begin{align}
\Delta^{\rho_\nu(\nu)f}_{f'}&=\sum_{\tau=0}^{T-1}\sum_{t=0}^{T_{{\rm mb}}-1}
\rho^{(\nu\tau)(t)}_{f}\delta^{(\nu\tau)(t)}_{f}h^{(\nu-1\tau)(t)}_{f'}\;,\\
\end{align}
and otherwise
\begin{align}
\Delta^{\rho_\nu(\nu)f}_{f'}&=\sum_{\tau=0}^{T-1}\sum_{t=0}^{T_{{\rm mb}}-1}
\rho^{(\nu\tau)(t)}_{f}\delta^{(\nu\tau)(t)}_{f}y^{(\nu-1\tau)(t)}_{f'}\;,\\
\rho^{(\nu\tau)(t)}_{f}\delta^{(\nu\tau)(t)}_{f}y^{(\nu-1\tau)(t)}_{f'}\;,\\
\Delta^{\rho_\tau(\nu)f}_{f'}&=\sum_{\tau=1}^{T-1}\sum_{t=0}^{T_{{\rm mb}}-1}
\rho^{(\nu\tau)(t)}_{f}\delta^{(\nu\tau)(t)}_{f}y^{(\nu\tau-1)(t)}_{f'}\;,\\
\Delta^{\beta(\nu \tau)}_{f}&=\sum_{t=0}^{T_{{\rm mb}}-1}\sum_{\epsilon=0}^{1}\sum_{f'=0}^{F_{\nu+1-\epsilon}-1}
\mathcal{H}^{(t)(\nu-\epsilon\tau+\epsilon)_{b_\epsilon}}_{ff'}\delta^{(t)(\nu-\epsilon\tau+\epsilon)}_{f'}\;,\\
\Delta^{\gamma(\nu \tau)}_{f}&=\sum_{t=0}^{T_{{\rm mb}}-1}\tilde{h}^{(t)(\nu\tau)}_{f}\sum_{\epsilon=0}^{1}\sum_{f'=0}^{F_{\nu+1-\epsilon}-1}
\mathcal{H}^{(t)(\nu-\epsilon\tau+\epsilon)_{b_\epsilon}}_{ff'}\delta^{(t)(\nu-\epsilon\tau+\epsilon)}_{f'}\;.
\end{align}

and

\begin{align}
\Delta^{f}_{f'}&=\sum_{\tau=0}^{T-1}\sum_{t=0}^{T_{{\rm mb}}-1} y^{(t)(N-1\tau)}_{f'}\delta^{(t)(N-1\tau)}_{f}\;.
\end{align}

\begin{subappendices}

\section{Backpropagation trough Batch Normalization}

For Backpropagation, we will need 

\begin{align}
\frac{\partial y^{(t')(\nu\tau)}_{f'}}{\partial h_{f}^{(t)(\nu\tau)}}&=
\gamma^{(\nu\tau)}_f\frac{\partial \tilde{h}_{f'}^{(t)(\nu\tau)}}{\partial h_{f}^{(t)(\nu\tau)}}\;.
\end{align}

Since

\begin{align}
\frac{\partial h^{(t')(\nu\tau)}_{f'}}{\partial h_{f}^{(t)(\nu\tau)}}&=\delta^{t'}_t\delta^{f'}_f\;,&
\frac{\partial \hat{h}_{f'}^{(\nu\tau)}}{\partial h_{f}^{(t)(\nu\tau)}}&=\frac{\delta^{f'}_f}{T_{{\rm mb}}}\:;
\end{align}

and

\begin{align}
\frac{\partial \left(\hat{\sigma}_{f'}^{(\nu\tau)}\right)^2}{\partial h_{f}^{(t)(\nu\tau)}}&=
\frac{2\delta^{f'}_f}{T_{{\rm mb}}}\left(h_{f}^{(t)(\nu\tau)}-\hat{h}_{f}^{(\nu\tau)}\right)\;,
\end{align}

we get

\begin{align}
\frac{\partial \tilde{h}_{f'}^{(t')(\nu\tau)}}{\partial h_{f}^{(t)(\nu\tau)}}&=
\frac{\delta^{f'}_f}{T_{{\rm mb}}}\left[\frac{T_{{\rm mb}}\delta^{t'}_t-1}
{\left(\left(\hat{\sigma}_{f}^{(\nu\tau)}\right)^2+\epsilon\right)^\frac12}-
\frac{\left(h_{f}^{(t')(\nu\tau)}-\hat{h}_{f}^{(\nu\tau)}\right)\left(h_{f}^{(t)(\nu\tau)}-\hat{h}_{f}^{(\nu\tau)}\right)}
{\left(\left(\hat{\sigma}_{f}^{(\nu\tau)}\right)^2+\epsilon\right)^\frac32}\right]\notag\\
&=\frac{\delta^{f'}_f}{\left(\left(\hat{\sigma}_{f}^{(\nu\tau)}\right)^2+\epsilon\right)^\frac12}
\left[\delta^{t'}_t-
\frac{1+\tilde{h}_{f}^{(t')(\nu\tau)}\tilde{h}_{f}^{(t)(\nu\tau)}}{T_{{\rm mb}}}\right]\;.
\end{align}

To ease the notation we will denote

\begin{align}
\tilde{\gamma}^{(\nu\tau)}_f&=
\frac{\gamma^{(\nu\tau)}_f}{\left(\left(\hat{\sigma}_{f}^{(\nu\tau)}\right)^2+\epsilon\right)^\frac12}\;.
\end{align}

so that

\begin{align}
\frac{\partial y_{f'}^{(t')(\nu\tau)}}{\partial h_{f}^{(t)(\nu\tau)}}&=
\tilde{\gamma}^{(\nu\tau)}_f \delta^{f'}_f\left[\delta^{t'}_t-
\frac{1+\tilde{h}_{f}^{(t')(\nu\tau)}\tilde{h}_{f}^{(t)(\nu\tau)}}{T_{{\rm mb}}}\right]\;.
\end{align}

This modifies the error rate backpropagation, as well as the formula for the weight update ($y$'s instead of $h$'s). In the following we will use the formula

\begin{align}
J^{(tt')(\nu\tau)}_{f}&=
\tilde{\gamma}^{(\nu\tau)}_f \left[\delta^{t'}_t-
\frac{1+\tilde{h}_{f}^{(t')(\nu\tau)}\tilde{h}_{f}^{(t)(\nu\tau)}}{T_{{\rm mb}}}\right]\;.
\end{align}

\section{RNN Backpropagation}

\subsection{RNN Error rate updates: details} \label{sec:rnnappenderrorrate}

Recalling the error rate definition

\begin{align}
\delta^{(t)( \nu\tau)}_f&=\frac{\delta}{\delta h^{(t)( \nu+1\tau)}_f }J(\Theta)\;,
\end{align}

we would like to compute it for all existing values of $\nu$ and $\tau$. As computed in chapter \ref{sec:chapterFNN}, one has for the maximum $\nu$ value

\begin{align}
\delta^{(t)(N-1\tau)}_f&= \frac{1}{T_{{\rm mb}}}\left(h_{f}^{(t)(N\tau)}-y_f^{(t)(\tau)}\right)\;.
\end{align}

Now since (taking Batch Normalization into account)

\begin{align}
h^{(t)(N\tau)}_{f}&=o\left(\sum_{f'=0}^{F_{N-1}-1}\Theta^f_{f'} y^{(t)(N-1\tau)}_{f}\right)\;,
\end{align}

and 

\begin{align}
h^{(t)(\nu\tau)}_f&=\tanh\left(\sum_{f'=0}^{F_{{\nu-1}}-1}\Theta^{\nu(\nu)f}_{f'}
y^{(t)(\nu-1\tau)}_{f'}+\sum_{f'=0}^{F_{{\nu}}-1}\Theta^{\tau(\nu)f}_{f'}
y^{(t)(\nu\tau-1)}_{f'}\right)\;,
\end{align}

we get for 

\begin{align}
\delta^{(t)(N-2\tau)}_f&= \sum_{t'=0}^{T_{{\rm mb}}}\left[\sum_{f'=0}^{F_{N}-1}
\frac{\delta h^{(t')(N\tau)}_{f'}}{\delta h^{(t)(N-1\tau)}_f }\delta^{(t')(N-1\tau)}_{f'}\right.\notag\\
&+\left.\sum_{f'=0}^{F_{N-1}-1}\frac{\delta h^{(t')(N-1\tau+1)}_{f'}}{\delta h^{(t)(N-1\tau)}_f }\delta^{(t')(N-2\tau+1)}_{f'}\right]\;.
\end{align}

Let us work out explicitly once (for a regression cost function and a trivial identity output function)

\begin{align}
\frac{\delta h^{(t')(N\tau)}_{f'}}{\delta h^{(t)(N-1\tau)}_f }&=\sum_{f''=0}^{F_{N-1}-1}\Theta^{f'}_{f''}\,
\frac{\delta y^{(t')(N-1\tau)}_{f''}}{\delta h^{(t)(N-1\tau)}_f } \notag\\
&=\Theta^{f'}_{f}\,J_f^{(tt')(N-1\tau)}\;.
\end{align}

as well as

\begin{align}
\frac{\delta h^{(t')(N-1\tau+1)}_{f'}}{\delta h^{(t)(N-1\tau)}_f }&=
\left[1-\left(h^{(t')(N-1\tau+1)}_{f'}\right)^2\right]\sum_{f''=0}^{F_{{N-1}}-1}\Theta^{\tau(N-1)f'}_{f''}
\frac{\delta y^{(t')(N-1\tau)}_{f''}}{\delta h^{(t)(N-1\tau)}_f }\notag\\
&=\mathcal{T}^{(t')(N-1\tau+1)}_{f'}\Theta^{\tau(N-1)f'}_{f}\,J_f^{(tt')(N-1\tau)}\;.
\end{align}

Thus

\begin{align}
\delta^{(t)(N-2\tau)}_f&= \sum_{t'=0}^{T_{{\rm mb}}}J_f^{(tt')(N-1\tau)}\left[\sum_{f'=0}^{F_{N}-1}
\Theta^{f'}_{f}\,\delta^{(t')(N-1\tau)}_{f'}\right.\notag\\
&\left.+\sum_{f'=0}^{F_{N-1}-1}\mathcal{T}^{(t')(N-1\tau+1)}_{f'}\Theta^{\tau(N-1)f'}_{f}\,\delta^{(t')(N-2\tau+1)}_{f'}\right]\;.
\end{align}

Here we adopted the convention that the $\delta^{(t')(N-2\tau+1)}$'s are $0$ if $\tau=T$. In a similar way, we derive for $\nu\leq N-1$

\begin{align}
\delta^{(t)(\nu-1\tau)}_f&= \sum_{t'=0}^{T_{{\rm mb}}}J^{(tt')(\nu\tau)}_f\left[\sum_{f'=0}^{F_{\nu+1}-1}
\mathcal{T}^{(t')(\nu+1\tau)}_{f'}\Theta^{\nu(\nu+1)f'}_{f}\,\delta^{(t')(\nu\tau)}_{f'}\right.\notag\\
&\left.+\sum_{f'=0}^{F_{\nu}-1}\mathcal{T}^{(t')(\nu\tau+1)}_{f'}\Theta^{\tau(\nu)f'}_{f}\,\delta^{(t')(\nu-1\tau+1)}_{f'}\right]\;.
\end{align}

Defining 

\begin{align}
\mathcal{T}^{(t')(N\tau)}_{f'}&=1\;,&
\Theta^{\nu(N)f'}_{f}&=\Theta^{f'}_{f}\;,
\end{align}

the previous $ \delta^{(t)(\nu-1\tau)}_f$ formula extends to the case $\nu  =N-1$. To unite the RNN and the LSTM formulas, let us finally define (with $a$ either $\tau$ or $\nu$

\begin{align}
\mathcal{H}^{(t')(\nu\tau)_a}_{ff'}&=\mathcal{T}^{(t')(\nu+1\tau)}_{f'}\Theta^{a(\nu+1)f'}_{f}\;,
\end{align}

thus (defining $b_0=\nu$ and $b_1=\tau$)

\begin{align}
\delta^{(t)(\nu-1\tau)}_f&= 
\sum_{t'=0}^{T_{{\rm mb}}}J^{(tt')(\nu\tau)}_f\sum_{\epsilon=0}^{1}\sum_{f'=0}^{F_{\nu+1-\epsilon}-1}
\mathcal{H}^{(t')(\nu-\epsilon\tau+\epsilon)_{b_\epsilon}}_{ff'}\delta^{(t')(\nu-\epsilon\tau+\epsilon)}_{f'}\;.
\end{align}

\subsection{RNN Weight and coefficient updates: details} \label{sec:rnncoefficient}

We want here to derive 

\begin{align}
\Delta^{\nu(\nu)f}_{f'}&=\frac{\partial}{\partial \Theta^{\nu(\nu)f}_{f'}} J(\Theta)&
\Delta^{\tau(\nu)f}_{f'}&=\frac{\partial}{\partial \Theta^{\tau(\nu)f}_{f'}} J(\Theta)\;.
\end{align}

We first expand

\begin{align}
\Delta^{\nu(\nu)f}_{f'}&=\sum_{\tau=0}^{T-1}\sum_{f''=0}^{F_\nu-1}\sum_{t=0}^{T_{{\rm mb}}-1}
\frac{\partial h^{(t)(\nu\tau)}_{f''}}{\partial \Theta^{\nu(\nu)f}_{f'}}
\delta^{(t)(\nu-1\tau)}_{f''}\;,\notag\\
\Delta^{\tau(\nu)f}_{f'}&=\sum_{\tau=0}^{T-1}\sum_{f''=0}^{F_\nu-1}\sum_{t=0}^{T_{{\rm mb}}-1}
\frac{\partial h^{(t)(\nu\tau)}_{f''}}{\partial \Theta^{\tau(\nu)f}_{f'}}
\delta^{(t)(\nu-1\tau)}_{f''}\;,
\end{align}

so that

\begin{align}
\Delta^{\nu(\nu)f}_{f'}&=\sum_{\tau=0}^{T-1}\sum_{t=0}^{T_{{\rm mb}}-1}
\mathcal{T}^{(t)(\nu\tau)}_{f}\delta^{(t)(\nu-1\tau)}_{f}h^{(t)(\nu-1\tau)}_{f'}\;,\\
\Delta^{\tau(\nu)f}_{f'}&=\sum_{\tau=1}^{T-1}\sum_{t=0}^{T_{{\rm mb}}-1}
\mathcal{T}^{(t)(\nu\tau)}_{f}\delta^{(t)(\nu-1\tau)}_{f}h^{(t)(\nu\tau-1)}_{f'}\;.
\end{align}

We also have to compute

\begin{align}
\Delta^{f}_{f'}&=\frac{\partial}{\partial \Theta^{f}_{f'}} J(\Theta)\;.
\end{align}

We first expand

\begin{align}
\Delta^{f}_{f'}&=\sum_{\tau=0}^{T-1}\sum_{f''=0}^{F_N-1}\sum_{t=0}^{T_{{\rm mb}}-1}
\frac{\partial h^{(t)(N\tau)}_{f''}}{\partial \Theta^{f}_{f'}}
\delta^{(t)(N-1\tau)}_{f''}\;
\end{align}

so that

\begin{align}
\Delta^{f}_{f'}&=\sum_{\tau=0}^{T-1}\sum_{t=0}^{T_{{\rm mb}}-1} h^{(t)(N-1\tau)}_{f'}\delta^{(t)(N-1\tau)}_{f}\;.
\end{align}

Finally, we need

\begin{align}
\Delta^{\beta(\nu \tau)}_{f}&=\frac{\partial}{\partial\beta^{(\nu \tau)}_{f}} J(\Theta)&
\Delta^{\gamma(\nu \tau)}_{f}&=\frac{\partial}{\partial \gamma^{(\nu \tau)}_{f}} J(\Theta)\;.
\end{align}

First

\begin{align}
\Delta^{\beta(\nu \tau)}_{f}&=\sum_{t=0}^{T_{{\rm mb}}-1}\left[\sum_{f'=0}^{F_{\nu+1}-1}
\frac{\partial h^{(t)(\nu+1\tau)}_{f'}}{\partial \beta^{(\nu \tau)}_{f}}\delta^{(t)(\nu\tau)}_{f'}+
\sum_{f'=0}^{F_{\nu}-1}
\frac{\partial h^{(t)(\nu\tau+1)}_{f'}}{\partial \beta^{(\nu \tau)}_{f}}\delta^{(t)(\nu-1\tau+1)}_{f'}\right]\;,\notag\\
\Delta^{\gamma(\nu \tau)}_{f}&=\sum_{t=0}^{T_{{\rm mb}}-1}\left[\sum_{f'=0}^{F_{\nu+1}-1}
\frac{\partial h^{(t)(\nu+1\tau)}_{f'}}{\partial \gamma^{(\nu \tau)}_{f}}\delta^{(t)(\nu\tau)}_{f'}+
\sum_{f'=0}^{F_{\nu}-1}
\frac{\partial h^{(t)(\nu\tau+1)}_{f'}}{\partial \gamma^{(\nu \tau)}_{f}}\delta^{(t)(\nu-1\tau+1)}_{f'}\right]\;.
\end{align}

So that

\begin{align}
\Delta^{\beta(\nu \tau)}_{f}&=\sum_{t=0}^{T_{{\rm mb}}-1}\left[\sum_{f'=0}^{F_{\nu+1}-1}
\mathcal{T}^{(t)(\nu+1\tau)}_{f'}\Theta^{\nu(\nu+1)f'}_{f}\delta^{(t)(\nu\tau)}_{f'}\right.\notag\\
&\left.+\sum_{f'=0}^{F_{\nu}-1}
\mathcal{T}^{(t)(\nu\tau+1)}_{f'}\Theta^{\tau(\nu)f'}_{f}\delta^{(t)(\nu-1\tau+1)}_{f'}\right]\;,\\
\Delta^{\gamma(\nu \tau)}_{f}&=\sum_{t=0}^{T_{{\rm mb}}-1}\left[\sum_{f'=0}^{F_{\nu+1}-1}
\mathcal{T}^{(t)(\nu+1\tau)}_{f'}\Theta^{\nu(\nu+1)f'}_{f}
\tilde{h}^{(t)(\nu\tau)}_{f}\delta^{(t)(\nu\tau)}_{f'}\right.\notag\\
&\left.+\sum_{f'=0}^{F_{\nu}-1}\mathcal{T}^{(t)(\nu\tau+1)}_{f'}
\Theta^{\tau(\nu)f'}_{f}\tilde{h}^{(t)(\nu\tau)}_{f}\delta^{(t)(\nu-1\tau+1)}_{f'}\right]\;,
\end{align}

which we can rewrite as

\begin{align}
\Delta^{\beta(\nu \tau)}_{f}&=\sum_{t=0}^{T_{{\rm mb}}-1}\sum_{\epsilon=0}^{1}\sum_{f'=0}^{F_{\nu+1-\epsilon}-1}
\mathcal{H}^{(t)(\nu-\epsilon\tau+\epsilon)_{b_\epsilon}}_{ff'}\delta^{(t)(\nu-\epsilon\tau+\epsilon)}_{f'}\;,\\
\Delta^{\gamma(\nu \tau)}_{f}&=\sum_{t=0}^{T_{{\rm mb}}-1}\tilde{h}^{(t)(\nu\tau)}_{f}\sum_{\epsilon=0}^{1}\sum_{f'=0}^{F_{\nu+1-\epsilon}-1}
\mathcal{H}^{(t)(\nu-\epsilon\tau+\epsilon)_{b_\epsilon}}_{ff'}\delta^{(t)(\nu-\epsilon\tau+\epsilon)}_{f'}\;.
\end{align}

\section{LSTM Backpropagation}

\subsection{LSTM Error rate updates: details} \label{sec:ARNNLSTMerror_rates}

As for the RNN 

\begin{align}
\delta^{(t)(N-1\tau)}_{f}&
=\frac{1}{T_{{\rm mb}}}\left(h^{(t)(N\tau)}_{f}-y^{(t)(\tau)}_{f}\right)\;.
\end{align}

Before  going any further, it will be useful to define 

\begin{align}
\mathcal{O}^{(t)(\nu\tau)}_{f}&=h^{(t)(\nu\tau)}_{f}
\left(1-o^{(t)(\nu\tau)}_{f}\right)\;,\notag\\
\mathcal{I}^{(t)(\nu\tau)}_{f}&=o^{(t)(\nu\tau)}_{f}\left(1-\tanh^2\left(c^{(t)(\nu\tau)}_{f}\right)
\right)g^{(t)(\nu\tau)}_{f} i^{(t)(\nu\tau)}_{f}\left(1-i^{(t)(\nu\tau)}_{f}\right)\;,\notag\\
\mathcal{F}^{(t)(\nu\tau)}_{f}&=o^{(t)(\nu\tau)}_{f}\left(1-\tanh^2\left(c^{(t)(\nu\tau)}_{f}\right)
\right)c^{(t)(\nu\tau-1)}_{f} f^{(t)(\nu\tau)}_{f}\left(1-f^{(t)(\nu\tau)}_{f}\right)\;,\notag\\
\mathcal{G}^{(t)(\nu\tau)}_{f}&=o^{(t)(\nu\tau)}_{f}\left(1-\tanh^2\left(c^{(t)(\nu\tau)}_{f}\right)
\right)i^{(t)(\nu\tau)}_{f}\left(1-\left(g^{(t)(\nu\tau)}_{f}\right)^2\right)\;,
\end{align}

and

\begin{align}
H^{(t)(\nu\tau)_a}_{ff'}&=\Theta^{o_a(\nu+1)f'}_{f}\mathcal{O}^{(t)(\nu+1\tau)}_{f'}
+\Theta^{f_a(\nu+1)f'}_{f}\mathcal{F}^{(t)(\nu+1\tau)}_{f'}\notag\\
&+\Theta^{g_a(\nu+1)f'}_{f}\mathcal{G}^{(t)(\nu+1\tau)}_{f'}
+\Theta^{i_a(\nu+1)f'}_{f}\mathcal{I}^{(t)(\nu+1\tau)}_{f'}\;.
\end{align}

As for RNN, we will start off by looking at 

\begin{align}
\delta^{(t)(N-2\tau)}_f&= \sum_{t'=0}^{T_{{\rm mb}}}\left[\sum_{f'=0}^{F_{N}-1}
\frac{\delta h^{(t')(N\tau)}_{f'}}{\delta h^{(t)(N-1\tau)}_f }\delta^{(t')(N-1\tau)}_{f'}\right.\notag\\
&+\left.\sum_{f'=0}^{F_{N-1}-1}\frac{\delta h^{(t')(N-1\tau+1)}_{f'}}{\delta h^{(t)(N-1\tau)}_f }\delta^{(t')(N-2\tau+1)}_{f'}\right]\;.
\end{align}

We will be able to get our hands on the second term with the general formula, so let us first look at

\begin{align}
\frac{\delta h^{(t')(N\tau)}_{f'}}{\delta h^{(t)(N-1\tau)}_f }&=\Theta^{f'}_{f}\,J_f^{(tt')(N-1\tau)}\;,
\end{align}

which is is similar to the RNN case. Let us put aside the second term of $\delta^{(t)(N-2\tau)}_f$, and look at the general case

\begin{align}
\delta^{(t)(\nu-1\tau)}_f&= \sum_{t'=0}^{T_{{\rm mb}}}\left[\sum_{f'=0}^{F_{\nu+1}-1}
\frac{\delta h^{(t')(\nu+1\tau)}_{f'}}{\delta h^{(t)(\nu\tau)}_f }\delta^{(t')(\nu\tau)}_{f'}
+\sum_{f'=0}^{F_{\nu}-1}\frac{\delta h^{(t')(\nu\tau+1)}_{f'}}{\delta h^{(t)(\nu\tau)}_f }\delta^{(t')(\nu-1\tau+1)}_{f'}\right]\;,
\end{align}
which involves to study in details
\begin{align}
\frac{\delta h^{(t')(\nu+1\tau)}_{f'}}{\delta h^{(t)(\nu\tau)}_f }&=
\frac{\delta o^{(t')(\nu+1\tau)}_{f'}}{\delta h^{(t)(\nu\tau)}_f }\tanh  c^{(t')(\nu+1\tau)}_{f'}\notag\\
&+\frac{\delta c^{(t')(\nu+1\tau)}_{f'}}{\delta h^{(t)(\nu\tau)}_f } o^{(t')(\nu+1\tau)}_{f'}
\left[1-\tanh^2  c^{(t')(\nu+1\tau)}_{f'}\right]\;.
\end{align}
Now
\begin{align}
\frac{\delta o^{(t')(\nu+1\tau)}_{f'}}{\delta h^{(t)(\nu\tau)}_f }&=o^{(t')(\nu+1\tau)}_{f'}
\left[1-o^{(t')(\nu+1\tau)}_{f'}\right]\sum_{f''=0}^{F_\nu-1}\Theta^{o_\nu(\nu+1)f'}_{f''}
\frac{\delta y^{(t')(\nu\tau)}_{f'}}{\delta h^{(t)(\nu\tau)}_f }\notag\\
&=o^{(t')(\nu+1\tau)}_{f'}
\left[1-o^{(t')(\nu+1\tau)}_{f'}\right]\Theta^{o_\nu(\nu+1)f'}_{f}
J^{(tt')(\nu\tau)}_f\;,
\end{align}
and
\begin{align}
\frac{\delta c^{(t')(\nu+1\tau)}_{f'}}{\delta h^{(t)(\nu\tau)}_f }&=
\frac{\delta i^{(t')(\nu+1\tau)}_{f'}}{\delta h^{(t)(\nu\tau)}_f }g^{(t')(\nu+1\tau)}_{f'}
+\frac{\delta g^{(t')(\nu+1\tau)}_{f'}}{\delta h^{(t)(\nu\tau)}_f }i^{(t')(\nu+1\tau)}_{f'}\notag\\
&+\frac{\delta f^{(t')(\nu+1\tau)}_{f'}}{\delta h^{(t)(\nu\tau)}_f }c^{(t')(\nu\tau)}_{f'}\;.
\end{align}
We continue our journey
\begin{align}
\frac{\delta i^{(t')(\nu+1\tau)}_{f'}}{\delta h^{(t)(\nu\tau)}_f }&=
 i^{(t')(\nu+1\tau)}_{f'}\left[1-  i^{(t')(\nu+1\tau)}_{f'}\right]\Theta^{i_\nu(\nu+1)f'}_{f}
J^{(tt')(\nu\tau)}_f\;,\notag\\
\frac{\delta f^{(t')(\nu+1\tau)}_{f'}}{\delta h^{(t)(\nu\tau)}_f }&=
 f^{(t')(\nu+1\tau)}_{f'}\left[1-  f^{(t')(\nu+1\tau)}_{f'}\right]\Theta^{f_\nu(\nu+1)f'}_{f}
J^{(tt')(\nu\tau)}_f\;,\notag\\
\frac{\delta g^{(t')(\nu+1\tau)}_{f'}}{\delta h^{(t)(\nu\tau)}_f }&=
\left[1-  \left(g^{(t')(\nu+1\tau)}_{f'}\right)^2\right]\Theta^{g_\nu(\nu+1)f'}_{f}
J^{(tt')(\nu\tau)}_f\;,
\end{align}
and our notations now come handy
\begin{align}
\frac{\delta h^{(t')(\nu+1\tau)}_{f'}}{\delta h^{(t)(\nu\tau)}_f }&=J^{(tt')(\nu\tau)}_fH^{(t)(\nu\tau)_\nu}_{ff'}\;.
\end{align}
This formula also allows us to compute the second term for $\delta^{(t)(N-2\tau)}_f$. In a totally similar manner
\begin{align}
\frac{\delta h^{(t')(\nu\tau+1)}_{f'}}{\delta h^{(t)(\nu\tau)}_f }&=J^{(tt')(\nu\tau)}_fH^{(t)(\nu-1\tau+1)_\tau}_{ff'}\;.
\end{align}
Going back to our general formula
\begin{align}
\delta^{(t)(\nu-1\tau)}_f&= \sum_{t'=0}^{T_{{\rm mb}}}J^{(tt')(\nu\tau)}_f\left[\sum_{f'=0}^{F_{\nu+1}-1}
H^{(t)(\nu\tau)_\nu}_{ff'}\delta^{(t')(\nu\tau)}_{f'}\right.\notag\\
&+\left.\sum_{f'=0}^{F_{\nu}-1}H^{(t)(\nu-1\tau+1)_\tau}_{ff'}\delta^{(t')(\nu-1\tau+1)}_{f'}\right]\;,
\end{align}
and as in the RNN case, we re-express it as (defining $b_0=\nu$ and $b_1=\tau$)
\begin{align}
\delta^{(t)(\nu-1\tau)}_f&= 
\sum_{t'=0}^{T_{{\rm mb}}}J^{(tt')(\nu\tau)}_f\sum_{\epsilon=0}^{1}\sum_{f'=0}^{F_{\nu+1-\epsilon}-1}
\mathcal{H}^{(t')(\nu-\epsilon\tau+\epsilon)_{b_\epsilon}}_{ff'}\delta^{(t')(\nu-\epsilon\tau+\epsilon)}_{f'}\;.
\end{align}
This formula is also valid for $\nu  =N-1$ if we define as for the RNN case
\begin{align}
\mathcal{H}^{(t')(N\tau)}_{f'}&=1\;,&
\Theta^{\nu(N)f'}_{f}&=\Theta^{f'}_{f}\;,
\end{align}

\subsection{LSTM Weight and coefficient updates: details}

We want to compute 

\begin{align}
\Delta^{\rho_{_\nu}(\nu)f}_{f'}&=\frac{\partial}{\partial \Theta^{\rho_{_\nu}(\nu)f}_{f'}} J(\Theta)&
\Delta^{\rho_{_\tau}(\nu)f}_{f'}&=\frac{\partial}{\partial \Theta^{\rho_{_\tau}(\nu)f}_{f'}} J(\Theta)\;,
\end{align}

with $\rho = (f,i,g,o)$. First we expand

\begin{align}
\Delta^{\rho_{_\nu}(\nu)f}_{f'}&=\sum_{\tau=0}^{T-1}\sum_{f''=0}^{F_\nu-1}\sum_{t=0}^{T_{{\rm mb}}-1}
\frac{\partial h^{(\nu\tau)(t)}_{f''}}{\partial \Theta^{\rho_{_\nu}(\nu)f}_{f'}}
\frac{\partial}{\partial h^{(\nu\tau)(t)}_{f''}} J(\Theta)\notag\\
&=\sum_{\tau=0}^{T-1}\sum_{f''=0}^{F_\nu-1}\sum_{t=0}^{T_{{\rm mb}}-1}
\frac{\partial h^{(\nu\tau)(t)}_{f''}}{\partial \Theta^{\rho_{_\nu}(\nu)f}_{f'}}
\delta^{(\nu\tau)(t)}_{f''}\;,
\end{align}

so that (with $\rho^{(\nu\tau)}=\left(\mathcal{F},\mathcal{I},\mathcal{G},\mathcal{O}\right)$) if $\nu=1$

\begin{align}
\Delta^{\rho_\nu(\nu-)f}_{f'}&=\sum_{\tau=0}^{T-1}\sum_{t=0}^{T_{{\rm mb}}-1}
\rho^{(\nu\tau)(t)}_{f}\delta^{(\nu\tau)(t)}_{f}h^{(\nu-1\tau)(t)}_{f'}\;,
\end{align}

and else

\begin{align}
\Delta^{\rho_\nu(\nu-)f}_{f'}&=\sum_{\tau=0}^{T-1}\sum_{t=0}^{T_{{\rm mb}}-1}
\rho^{(\nu\tau)(t)}_{f}\delta^{(\nu\tau)(t)}_{f}y^{(\nu-1\tau)(t)}_{f'}\;,\\
\Delta^{\rho_\tau(\nu)f}_{f'}&=\sum_{\tau=1}^{T-1}\sum_{t=0}^{T_{{\rm mb}}-1}
\rho^{(\nu\tau)(t)}_{f}\delta^{(\nu\tau)(t)}_{f}y^{(\nu\tau-1)(t)}_{f'}\;.
\end{align}

We will now need to compute

\begin{align}
\Delta^{\beta(\nu\tau)}_{f}&=\frac{\partial}{\partial \beta^{(\nu\tau)}_f} J(\Theta)&
\Delta^{\gamma(\nu\tau)}_{f}&=\frac{\partial}{\partial \gamma^{(\nu\tau)}_f} J(\Theta)\;.
\end{align}

For that we need to look at 

\begin{align}
\Delta^{\beta(\nu\tau)}_{f}&=\sum_{f'=0}^{F_{\nu+1}-1}\sum_{t'=0}^{T_{{\rm mb}}-1}
\frac{\partial h^{(\nu+1\tau)(t')}_{f'}}{\partial \beta^{(\nu\tau)}_f}\delta^{(\nu\tau)(t')}_{f'}
+\sum_{f'=0}^{F_{\nu}-1}\sum_{t'=0}^{T_{{\rm mb}}-1}\frac{\partial h^{(\nu\tau+1)(t')}_{f'}}
{\partial \beta^{(\nu\tau)}_f}\delta^{(\nu-1\tau+1)(t')}_{f'}\notag\\
&=\sum_{t=0}^{T_{{\rm mb}}-1}\left\{\sum_{f'=0}^{F_{\nu+1}-1}
H^{(t)(\nu\tau)}_{ff'}\delta^{(t)(\nu\tau)}_{f'}
+\sum_{f'=0}^{F_{\nu}-1}H^{(t)(\nu-1\tau+1)}_{ff'}\delta^{(t)(\nu\tau+1)}_{f'}\right\}\;. 
\end{align}

and

\begin{align}
\Delta^{\gamma(\nu\tau)}_{f}&=\sum_{t=0}^{T_{{\rm mb}}-1}\tilde{h}^{(t)(\nu\tau)}_{f}\left\{\sum_{f'=0}^{F_{\nu+1}-1}
H^{(t)(\nu\tau)}_{ff'}\delta^{(t)(\nu\tau)}_{f'}
+\sum_{f'=0}^{F_{\nu}-1}H^{(t)(\nu-1\tau+1)}_{ff'}\delta^{(t)(\nu-1\tau+1)}_{f'}\right\}\;,
\end{align}

which we can rewrite as

\begin{align}
\Delta^{\beta(\nu \tau)}_{f}&=\sum_{t=0}^{T_{{\rm mb}}-1}\sum_{\epsilon=0}^{1}\sum_{f'=0}^{F_{\nu+1-\epsilon}-1}
\mathcal{H}^{(t)(\nu-\epsilon\tau+\epsilon)_{b_\epsilon}}_{ff'}\delta^{(t)(\nu-\epsilon\tau+\epsilon)}_{f'}\;,\\
\Delta^{\gamma(\nu \tau)}_{f}&=\sum_{t=0}^{T_{{\rm mb}}-1}\tilde{h}^{(t)(\nu\tau)}_{f}\sum_{\epsilon=0}^{1}\sum_{f'=0}^{F_{\nu+1-\epsilon}-1}
\mathcal{H}^{(t)(\nu-\epsilon\tau+\epsilon)_{b_\epsilon}}_{ff'}\delta^{(t)(\nu-\epsilon\tau+\epsilon)}_{f'}\;.
\end{align}

Finally, as in the RNN case

\begin{align}
\Delta^{f}_{f'}&=\frac{\partial}{\partial \Theta^{f}_{f'}} J(\Theta)\;.
\end{align}

We first expand

\begin{align}
\Delta^{f}_{f'}&=\sum_{\tau=0}^{T-1}\sum_{f''=0}^{F_N-1}\sum_{t=0}^{T_{{\rm mb}}-1}
\frac{\partial h^{(t)(N\tau)}_{f''}}{\partial \Theta^{f}_{f'}}
\delta^{(t)(N-1\tau)}_{f''}\;
\end{align}

so that

\begin{align}
\Delta^{f}_{f'}&=\sum_{\tau=0}^{T-1}\sum_{t=0}^{T_{{\rm mb}}-1} h^{(t)(N-1\tau)}_{f'}\delta^{(t)(N-1\tau)}_{f}\;.
\end{align}

\newpage

\section{Peephole connexions}

\begin{figure}[H]
\begin{center}
\begin{tikzpicture}
\node[] at (0,0) {\includegraphics[scale=1.4]{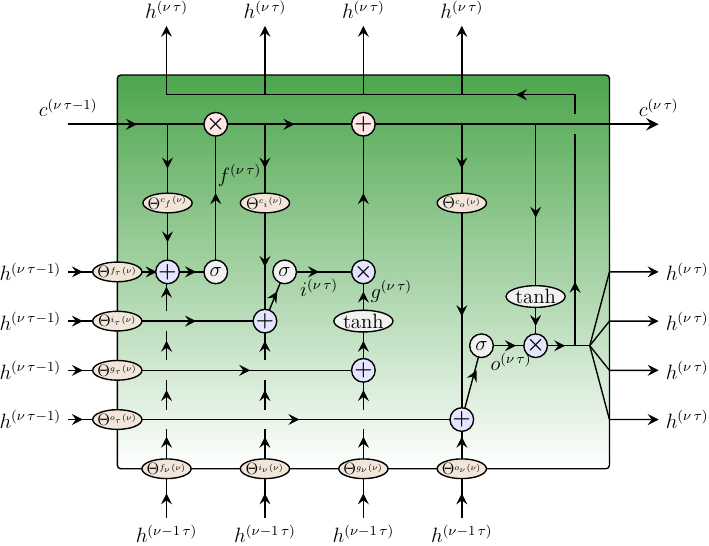}};
\end{tikzpicture}
\caption{\label{fig:peepholeLSTM}LSTM hidden unit with peephole}
\end{center}
\end{figure}

Some LSTM variants probe the cell state to update the gate themselves. This is illustrated in figure \ref{fig:peepholeLSTM}

Peepholes modify the gate updates in the following way

\begin{align}
i^{(\nu\tau)(t)}_f&=\sigma\left(\sum_{f'=0}^{F_{{\nu-1}}-1}\Theta^{i_{_\nu}(\nu)f}_{f'}
h^{(\nu-1\tau)(t)}_{f'}+\sum_{f'=0}^{F{_{\nu}}-1}\left[\Theta^{i_{_\tau}(\nu)f}_{f'}
h^{(\nu\tau-1)(t)}_{f'}+\Theta^{c_{_i}(\nu)f}_{f'}c^{(\nu\tau-1)(t)}_{f'}\right]\right)\;,\\
f^{(\nu\tau)(t)}_f&=\sigma\left(\sum_{f'=0}^{F{_{\nu-1}}-1}\Theta^{f_{_\nu}(\nu)f}_{f'}
h^{(\nu-1\tau)(t)}_{f'}+\sum_{f'=0}^{F_{{\nu}}-1}\left[\Theta^{f_{_\tau}(\nu)f}_{f'}
h^{(\nu\tau-1)(t)}_{f'}+\Theta^{c_{_f}(\nu)f}_{f'}c^{(\nu\tau-1)(t)}_{f'}\right]\right)\;,\\
o^{(\nu\tau)(t)}_f&=\sigma\left(\sum_{f'=0}^{F_{{\nu-1}}-1}\Theta^{o_{_\nu}(\nu)f}_{f'}
h^{(\nu-1\tau)(t)}_{f'}+\sum_{f'=0}^{F_{{\nu}}-1}\left[\Theta^{o_{_\tau}(\nu)f}_{f'}
h^{(\nu\tau-1)(t)}_{f'}+\Theta^{c_{_o}(\nu)f}_{f'}c^{(\nu\tau)(t)}_{f'}\right]\right)\;,
\end{align}
which also modifies the LSTM backpropagation algorithm in a non-trivial way. As it as been shown that different LSTM formulations lead to pretty similar results, we leave to the reader the derivation of the backpropagation update rules as an exercise.

\end{subappendices}

\chapter{Conclusion}

\yinipar{\fontsize{60pt}{72pt}\usefont{U}{Kramer}{xl}{n}W}e have come to the end of our journey. I hope this note lived up to its promises, and that the reader now understands better how a neural network is designed and how it works under the hood. To wrap it up, we have seen the architecture of the three most common neural networks, as well as the careful mathematical derivation of their training formulas.

\vspace{0.2cm}

Deep Learning seems to be a fast evolving field, and this material might be out of date in a near future, but the index approach adopted will still allow the reader -- as it as helped the writer -- to work out for herself what is behind the next state of the art architectures.

\vspace{0.2cm}

Until then, one should have enough material to encode from scratch its own FNN, CNN and RNN-LSTM, as the author did as an empirical proof of his formulas.

\printbibliography
\end{document}